%% file: main.tex
\PassOptionsToPackage{table}{xcolor}
\documentclass[letterpaper]{article} 
\usepackage{aaai25}  
\usepackage{times}  
\usepackage{helvet}  
\usepackage{courier}  
\usepackage[hyphens]{url}  
\usepackage{graphicx} 
\usepackage{natbib}  
\usepackage{caption} 
\usepackage{algorithm}
\usepackage{algorithmic}

\usepackage{newfloat}
\usepackage{listings}
\DeclareCaptionStyle{ruled}{labelfont=normalfont,labelsep=colon,strut=off} 
\floatstyle{ruled}
\newfloat{listing}{tb}{lst}{}
\floatname{listing}{Listing}
\usepackage{amsmath,amssymb,amsfonts,amsthm,mathtools,bm}
\usepackage{enumitem}
\usepackage{multirow}
\usepackage{bm}
\usepackage{array}
\usepackage{color}
\usepackage[table]{xcolor} 
\definecolor{lightgray}{rgb}{0.83, 0.83, 0.83}
\usepackage{multicol}
\usepackage{caption}
\usepackage{diagbox}
\usepackage{pifont}
\usepackage{pdfpages}
\usepackage{arydshln}
\usepackage{anyfontsize}

\definecolor{lightgray}{rgb}{0.83, 0.83, 0.83}
\definecolor{bubblegum}{rgb}{0.99, 0.76, 0.8}
\definecolor{cherryblossompink}{rgb}{1.0, 0.72, 0.77}
\definecolor{classicrose}{rgb}{0.98, 0.8, 0.91}
\definecolor{babypink}{rgb}{0.96, 0.76, 0.76}
\definecolor{blanchedalmond}{rgb}{1.0, 0.92, 0.8}
\definecolor{palepink}{rgb}{0.98, 0.85, 0.87}
\definecolor{aliceblue}{rgb}{0.94, 0.97, 1.0}
\definecolor{lightyellow}{rgb}{1.0, 1.0, 0.88}
\definecolor{almond}{rgb}{0.94, 0.87, 0.8}
\definecolor{amber}{rgb}{1.0, 0.75, 0.0}
\definecolor{antiquewhite}{rgb}{0.98, 0.92, 0.84}
\definecolor{deepchampagne}{rgb}{0.98, 0.84, 0.65}
\definecolor{brinkpink}{rgb}{0.98, 0.38, 0.5}
\definecolor{caribbeangreen}{rgb}{0.0, 0.8, 0.6}
\definecolor{carnelian}{rgb}{0.7, 0.11, 0.11}
\definecolor{beaver}{rgb}{0.62, 0.51, 0.44}
\definecolor{cinereous}{rgb}{0.6, 0.51, 0.48}
\definecolor{royalblue(web)}{rgb}{0.25, 0.41, 0.88}
\definecolor{royalblue(traditional)}{rgb}{0.0, 0.14, 0.4}

\newcolumntype{g}{>{\columncolor{lightgray}}c}
\newcolumntype{a}{>{\columncolor{blanchedalmond}}c}
\newcolumntype{s}{>{\columncolor{deepchampagne}}c}

\newcommand{\blue}[1]{\textcolor{black}{#1}}

\usepackage{multirow}
\usepackage{caption}
\usepackage{bm}

\input{macro}

\title{Towards Precise Prediction Uncertainty in GNNs:\\
Refining GNNs with Topology-grouping Strategy}
\author{
    Hyunjin Seo\textsuperscript{1,3},
    Kyusung Seo\textsuperscript{1}$^*$,
    Joonhyung Park\textsuperscript{1}$^*$,
    Eunho Yang\textsuperscript{1,2}$^\dag$
}
\affiliations{
    \textsuperscript{\rm 1}KAIST\\ \textsuperscript{\rm 2}AITRICS\\ \textsuperscript{\rm 3}Polymerize\\
    \{bella72, seo3650, deepjoon, eunhoy\}@kaist.ac.kr
}

\usepackage{bibentry}

\begin{document}

\maketitle

\def\thefootnote{$\dag$}\footnotetext{Corresponding Author.}
\def\thefootnote{$*$}\footnotetext{Equal Contribution.}

\begin{abstract}
Recent advancements in graph neural networks (GNNs) have highlighted the critical need of calibrating model predictions, with neighborhood prediction similarity recognized as a pivotal component. 
Existing studies suggest that nodes with analogous neighborhood prediction similarity often exhibit similar calibration characteristics. 
Building on this insight, recent approaches incorporate neighborhood similarity into node-wise temperature scaling techniques.
However, our analysis reveals that this assumption does not hold universally. 
Calibration errors can differ significantly even among nodes with comparable neighborhood similarity, depending on their confidence levels.
This necessitates a re-evaluation of existing GNN calibration methods, as a single, unified approach may lead to sub-optimal calibration.
In response, we introduce \textsc{Simi-Mailbox}, a novel approach that categorizes nodes by both neighborhood similarity and their own confidence, irrespective of proximity or connectivity.
Our method allows fine-grained calibration by employing \textit{group-specific} temperature scaling, with each temperature tailored to address the specific miscalibration level of affiliated nodes, rather than adhering to a uniform trend based on neighborhood similarity. 
Extensive experiments demonstrate the effectiveness of our \textsc{Simi-Mailbox} across diverse datasets on different GNN architectures, achieving up to 13.79\% error reduction compared to uncalibrated GNN predictions.
\end{abstract}

%

\section{Introduction}\label{1_introduction}

Graph neural networks (GNNs) have demonstrated remarkable performance in modeling graph data and addressing diverse graph-based tasks, such as node classification~\citep{kipf2016semi, hamilton2017inductive, xu2018representation, park2021graphens}, link prediction~\citep{zhang2018link, yun2021neo, ahn2021variational, zhu2021neural}, and graph classification~\citep{lee2018graph, sui2022causal, hou2022graphmae}.
Beyond achieving correct prediction, precisely quantifying prediction uncertainty is nontrivial for the reliable utilization of neural networks in downstream decision-making process.
Recognizing such need, numerous calibration studies have been actively proposed in vision and language domains~\citep{guo2017calibration, mukhoti2020calibrating, zhang2020mix, xing2019distance, jiang2021can, minderer2021revisiting}.

Recently, network calibration has also drawn attention in the field of GNNs~\citep{wang2021confident, hsu2022makes, hsu2022graph, shi2022select, wang2022gcl, liu2022calibration}, highlighting neighborhood prediction similarity as a crucial factor for calibration. 
Contemporary studies in GNN calibration, CaGCN~\citep{wang2021confident} and GATS~\citep{hsu2022makes}, suggest that nodes with similar neighborhood prediction similarity tend to exhibit analogous calibration characteristics.
Specifically, CaGCN asserts that nodes with disparate neighbors should ideally have lower confidence levels, as the local message propagation in GNNs makes accurately classifying such instances more challenging.
Conversely, GATS elucidates the correlation between neighborhood prediction similarity and calibration errors, indicating the highest errors for nodes with conflicting neighbors.
To account for these trends, they incorporate neighborhood similarity into node-wise temperature scaling, facilitating confidence propagation between adjacent nodes.

However, our analysis reveals that calibration cannot be effectively addressed by applying a single, unified trend.
Specifically, we observe that calibration errors vary significantly among nodes with comparable neighborhood similarity, depending on their individual confidence levels.
More critically, both over-confidence and under-confidence can occur in nodes with similar neighborhood similarity but differing confidence levels. This phenomenon has not been effectively captured in previous studies, as they do not fully account for both factors. Consequently, their assumptions may lead to sub-optimal calibration, as they are not universally applicable.

To address this, we introduce \textsc{Simi-Mailbox}, a novel post-hoc calibration method designed to overcome these limitations.
Our method categorizes nodes based on both neighborhood representational similarity and confidence, irrespective of proximity or connectivity.
This grouping strategy is grounded on our observation that nodes with comparable levels of neighborhood similarity and confidence exhibit similar calibration errors.
\textsc{Simi-Mailbox} then assigns \textit{group-specific} temperatures to adjust the predictions of nodes within each group.
This fine-grained approach ensures that each group-wise temperature is tailored to address the specific miscalibration of affiliated nodes, instead of relying on a uniform tendency.

In summary, our contributions are three-fold:
\begin{itemize}
    \item We elucidate the limitations inherent in current calibration methods, particularly concerning neighborhood prediction similarity - a recognized key component for GNN calibration.
    \item Given these limitations, we propose \textsc{Simi-Mailbox}, a novel calibration method that rectifies miscalibration by introducing group-specific temperatures. Each group-wise temperature is focused on adjusting the predictions of affiliated nodes, rather than scaling all nodes according to a unified trend.
    \item We validate the efficacy of \textsc{Simi-Mailbox} through comprehensive experiments, incorporating both quantitative and qualitative evaluations.
\end{itemize}

\section{Related Works}\label{2_related_works}

\paragraph{Uncertainty Quantification and Post-hoc Calibration.}
Network calibration, while sharing the root in uncertainty quantification with conformal prediction and bayesian methods, concentrates on aligning model predictions with empirical event frequencies, differing from providing intervals around prediction or modeling uncertainty under probability distribution. 
Conformal prediction aims to generate tight prediction sets that encompass the true outcome with a pre-specified coverage. Its foundational concept was presented in~\citep{vovk2005algorithmic}, and the study for providing good coverage has been consistently explored, evolving through \citep{romano2020classification, cauchois2021knowing, angelopoulos2020uncertainty}. 
Bayesian approaches, on the other hand, use probabilistic modeling to interpret the uncertainty via posterior distribution. 
Their representative techniques include ensembles~\citep{lakshminarayanan2017simple,wen2020batchensemble}, dropout~\citep{gal2016dropout} and Bayesian Neural Networks (BNNs) for applying Bayesian inference in neural networks~\citep{depeweg2018decomposition, maddox2019simple, dusenberry2020efficient}.

Distinct from the aforementioned approaches, calibration is focused on refining the trustworthiness of the model prediction. 
Their goal is focused on adjusting the model’s confidence to match the ground-truth probability. 
Among diverse calibration techniques, post-hoc calibration methods have found widespread adoption, owing to their computationally efficiency compared to traditional Bayesian approaches and model regularization-based methods~\citep{ma2021meta, jung2023scaling}. Moreover, they imposes no constraints during the pretraining phase of main models, thereby enhancing its versatility across diverse architectures.
Techniques such as Platt scaling~\citep{platt1999probabilistic}, Temperature scaling (TS)~\citep{guo2017calibration}, and Ensemble temperature scaling (ETS)~\citep{zhang2020mix} have been developed for this purpose, with TS being notably effective for its simplicity and effectiveness in multi-class calibration. 

\paragraph{Grouping-based Calibration.} Addressing miscalibrations in a group-wise manner has been studied in \citep{hebert2018multicalibration, perez2022beyond, yang2023beyond}. 
\citep{hebert2018multicalibration} introduced multicalibration strategy, aiming to achieve calibration within diverse, overlapping subgroups to enhance both fairness and accuracy in machine learning models. 
Meanwhile, \citep{perez2022beyond} presented the concept of grouping loss as a novel metric to assess the variance in true probabilities sharing the same confidence score, challenging existing calibration approaches. 
\citep{yang2023beyond} proposed a new semantic partitioning approach for neural network calibration and utilized learnable grouping function to refine calibration beyond traditional methods.
Nevertheless, these studies do not provide the \textbf{specific principles} for effective categorization, which highlights the distinction of our work from preceding ones. More unique aspects of our approach in comparison to prior works are discussed in the Appendix.

\paragraph{Uncertainty Quantification for GNNs.} 
Recent literature has increasingly focused on quantifying uncertainty in GNNs, with methods ranging from conformal prediction using local topologies~\citep{huang2024uncertainty, zargarbashi2023conformal} to Bayesian approaches~\citep{stadler2021graph, rong2019dropedge, hasanzadeh2020bayesian, elinas2020variational, pal2019bayesian, zhao2020uncertainty} that concentrate on the interdependent graph data and GNNs. The literature also highlights post-processing calibration strategies~\citep{wang2021confident, hsu2022makes, hsu2022graph, wang2022gcl, shi2022select, liu2022calibration}, with \citep{wang2021confident} pioneering in revealing unexpected underconfidence in GNN predictions. They introduced CaGCN, which employs GCN for node-specific calibration through adjacent predictions. Expanding this, GATS \citep{hsu2022makes} explored factors leading to GNN calibration errors and designed GAT-based node-wise calibration function considering these factors. They further introduced an edge-wise calibration error metric to capture the non-iid nature of graphs in \citep{hsu2022graph}. In a different approach, GCL~\citep{wang2022gcl} addressed the underconfidence of GNNs by integrating a minimal-entropy regularization with the cross-entropy loss, up-weighting the loss on highly confident nodes.

\section{Preliminaries}\label{3_preliminaries}

\paragraph{Problem Setup.}
We focus on calibrating the prediction uncertainty of GNNs for semi-supervised node classification in a post-hoc setting. In this context, uncertainty denotes the model's confidence level in its predictions, while calibration aims to align this uncertainty with the true accuracy, enhancing the model's reliability. Thus, our objective is to minimize the gap between the predicted probability and the actual accuracy of given data. During the post-hoc calibration phase, the validation set is used for training to enhance generalization to unseen data, avoiding the overfitting risk associated with reusing the original training set.

\input{obs_plot}

Let an undirected graph be denoted as $\mathcal{G}(\mathcal{V},\mathcal{E})$, where $\mathcal{V}$ and $\mathcal{E}$ indicate the sets of vertices and edges respectively.
The vertex set $\mathcal{V}$ is represented by a feature matrix $\bm X=[\mathbf x_1^\mathsf{T},...,\mathbf x_{|\mathcal{V}|}^\mathsf{T}]\in\mathbb{R}^{|\mathcal{V}|\times D}$ and the edge set $\mathcal{E}$ is denoted by an adjacency matrix $\bm A\in\mathbb{R}^{|\mathcal{V}|\times |\mathcal{V}|}$.
Given the node-wise predictions $\hat y=[\hat y_1,...,\hat y_{|\mathcal{V}|}]^\mathsf{T}$ and output confidence $\hat p = [\hat p_1,...,\hat p_{|\mathcal{V}|}]^\mathsf{T}\in\mathbb R^{|\mathcal{V}|}$ from a trained GNN, the GNN $f_\theta$ is \emph{well-calibrated} if ${\hat p}_i$ for each node $i$ accurately serves the ground-truth probability $p_{\text{true}}$, formulated as below:
\begin{equation}
    \mathbb{P}(\hat y_i=y_i|\hat p_i=p_{\text{true}})= p_{\text{true}},\quad\forall p_{\text{true}}\in[0,1].
\end{equation}
The expected calibration error (ECE)~\citep{naeini2015obtaining} has been recognized as the de facto metric to evaluate the calibration quality of network predictions. 
ECE groups nodes according to their confidences into $M$ equally partitioned confidence intervals $\{B_1,...,B_M\}$ and assesses the expected discrepancy between accuracy and average confidence within individual bins:
\begin{equation}
    \text{ECE}=\sum_{m=1}^M \frac{|B_m|}{|\mathcal{V}|} \Big|\text{acc}(B_m)-\text{conf}(B_m)\Big|,
\end{equation}
where $|B_m|$ refers to the number of nodes within the $m$-th interval.
Here, the accuracy and average confidence for the $m$-th bin are defined as $\text{acc}(B_m)=\frac{1}{|B_m|}\sum_{i\in B_m}\mathbf{1}[y_i=\hat y_i]$ and $\text{conf}(B_m)=\frac{1}{|B_m|}\sum_{i\in B_m}\hat p_i$, respectively. 

\paragraph{Neighborhood Similarity in Prior Studies.}
The concept of neighborhood similarity has been recognized as a primary element in the field of GNN calibration~\citep{wang2021confident, hsu2022makes,hsu2022graph, liu2022calibration}.
Among them, CaGCN~\citep{wang2021confident} advocates that given the challenges GNNs encounter in accurately classifying nodes with conflicting neighbors, the confidence levels in such cases should ideally remain still or decrease.
Conversely, confidence for nodes linked to agreeing nodes should elevate, addressing the prevalent underconfidence in GNNs.
Stemmed from this insight, they employ GCN~\citep{kipf2016semi} as a node-wise calibration function to propagate the confidence to neighboring counterparts.
In parallel, GATS~\citep{hsu2022makes} underscores the correlation between neighborhood prediction similarity and calibration error, demonstrating an increment in error with a decrement in similarity.
This relationship is incorporated into the normalized attention coefficients within their GAT~\citep{velivckovic2017graph}-founded node-level temperature function.

\section{In-depth Analysis on Neighborhood Similarity}\label{4_limitation}
In this section, we provide a comprehensive analysis of both uncalibrated and calibrated predictions from existing studies, CaGCN and GATS, using the CoraFull dataset~\citep{bojchevski2017deep}.
Leveraging GCN as the backbone architecture, we first partition nodes into 10 equal intervals $\{B^{(1)},...,B^{(10)}\}$ based on the proportion of neighbors sharing the same predicted labels, denoted as neighborhood prediction similarity $\bm s(i)$:
\begin{equation}
    \bm s(i)=\frac{\sum_{j\in\mathcal{N}_i}\bm 1[\hat y_i=\hat y_j]}{|\mathcal{N}_i|},
\end{equation}
where $\mathcal{N}_i$ represents the set of neighbors associated with node $i$. 
For each subgroup $B^{(l)}$, we calculate the calibration error as the discrepancy between their average confidence and the accuracy, \textit{i.e.}, $\text{acc}(B^{(l)})-\text{conf}(B^{(l)})$. 
These discrepancies are depicted as heatmap bars in the second row of Figure \ref{Fig: obs_plot}.

Furthermore, we also analyze predictions by considering both neighborhood similarity and confidence. 
We begin by grouping confidence into 10 equal intervals $\{B_1,...,B_{10}\}$, and then further categorize nodes within each confidence interval into 10 equal-width intervals based on $\bm s(i)$. 
The subgroup within the $l$-th similarity interval and $m$-th confidence interval is denoted as $B_m^{(l)}$.
For each subgroup $B_m^{(l)}$, the calibration error is computed as $\text{acc}(B_m)-\text{conf}(B_m^{(l)})$.
These discrepancies are illustrated as heatmap matrices in the last row of Figure \ref{Fig: obs_plot}, with $\hat p$ representing uncalibrated confidence and $\widetilde p$ representing calibrated confidence.

In Figure \ref{Fig: obs_plot}, the heatmap elements represent the differences between accuracy and average confidence. Deeper shades of \textbf{red} indicate that the calibrated confidence is lower than the accuracy (\textbf{under-confident}), while deeper shades of \textbf{green} indicate that confidence exceeds the accuracy (\textbf{over-confident}).
Our findings show that calibration errors can vary significantly among nodes with the same level of neighborhood similarity but different confidence.
Notably, both under- and over-confidence are observed in uncalibrated predictions within $\bm s(i)\in(0.1, 1.0]$ similarity intervals of the heatmap matrices.

Moreover, our analysis reveals that existing methods, which apply a unified policy to nodes with similar levels of neighborhood similarity, fail to achieve consistent calibration across diverse neighborhood similarity levels. 
While these methods may appear well-calibrated according to the heatmap bars in the second row, they demonstrate suboptimal results when their predictions are extended across confidence intervals.
Specifically, CaGCN exhibits severe under-confidence in $\widetilde p_i\in(0.9, 1.0]$ confidence interval, with a maximum discrepancy of approximately 16.34\% within the $\bm s(i)\in(0.1, 0.2]$ similarity range.
GATS, on the other hand, demonstrates suboptimal calibration in regions of low prediction similarity, particularly in the $\widetilde p_i\in(0.2,0.4]$ and $\widetilde p_i\in(0.6,0.8]$ ranges, where the average discrepancies are 7.45\% and 7.17\% in the $\bm s(i)\in(0, 0.4]$ intervals, respectively.
Hence, our observations suggest that a unified assumption to calibrating predictions based on neighborhood similarity cannot effectively achieve fine-grained calibration.
We also provide an algorithmic perspective on the limitations of previous work, along with additional investigation results on more benchmark datasets, in the Appendix.

\section{Proposed Method}\label{5_proposed_method}
Given the limitation of earlier studies, we introduce \textsc{Simi-Mailbox}, a post-hoc calibration method designed to rectify miscalibration in GNNs across varying levels of neighborhood similarity.
Building on our novel observation, \textsc{Simi-Mailbox} categorizes nodes based on both neighborhood similarity and confidence levels, ensuring that nodes within the same cluster exhibit similar calibration errors. 
Subsequently, our method employs group-specific temperature scaling to adjust the predictions of nodes in the designated cluster. 
These group-wise temperatures are tailored to correct the specific miscalibration associated with each group, instead of relying on a uniform tendency.
The temperatures are optimized by directly minimizing the discrepancy between average confidence and accuracy within each cluster.

\subsection{Intuition: Topology Grouping Matters}\label{5.1}
\vspace{-1mm}
\input{Table_obs}
For effective group-wise calibration, it is essential to categorize nodes in a manner that ensures they share a similar degree of miscalibration.
This allows each group's temperature to be precisely tailored to address specific miscalibration levels rather than applying a broad, generalized adjustment. 
To this end, we present a novel observation suggesting that nodes with similar neighborhood prediction similarity $s(i)$ and confidence $\hat p_i$ share similar magnitudes of calibration errors.
To substantiate this, we evaluate the variance of calibration errors under three different scenarios: (1) node-wise variance involving all nodes (specified as \textbf{Node-wise}), (2) variance within each confidence interval (specified as \textbf{Conf.}), and (3) variance within each neighborhood similarity sub-interval within each confidence interval (specified as \textbf{Neig. Sim.}).

To explore the third scenario, we assess the variability in calibration errors across neighborhood similarity intervals within each confidence interval. 
Let $B_m^{(l)}$ represent the set of nodes in $l$-th neighborhood similarity interval and $m$-th confidence interval.
The calibration error for each node $i$, defined as the absolute difference between its confidence and the accuracy associated with its confidence interval, is denoted as $D(i)$.
We first calculate the variance of calibration error within each $B_m^{(l)}$, denoted as $V(B_m^{(l)})$:
\begin{equation}
    \begin{split}
        &V(B_m^{(l)})=\frac{1}{|B_m^{(l)}|-1}\sum_{i\in B_m^{(l)}}(D(i)-\bar D_m^{(l)})^2,\\
        &D(i)=|\text{Acc}(B_{m})-\hat p_i|.
    \end{split}
\end{equation}

where $\bar D_m^{(l)}$ represents the mean calibration error for nodes in $B_m^{(l)}$. 
We then average these variances over the collection $B^{\text{sim}} =\{B_1^{(1)},B_1^{(2)}, ...,B_2^{(1)},B_2^{(2)},...\}$, which incorporates all $B_m^{(l)}$ spanning the entire confidence intervals: 
\begin{equation}
    V^{\text{sim}}=\frac{1}{|B^{\text{sim}}|}\sum_{B_m^{(l)}\in B^{\text{sim}}}V(B_m^{(l)}).
\end{equation}

Similarly, to assess the second scenario, we calculate the variability in calibration errors across confidence intervals $B^{\text{conf}}=\{B_1,B_2,...\}$ by computing the variance within each $B_m$:
\begin{equation}
    V(B_m)=\frac{1}{|B_m|-1}\sum_{i\in B_m}(D(i)-\bar D_m)^2,
\end{equation}
where $\bar D_m$ refers to the average calibration error for nodes within $B_m$. 
Following the approach used in GATS, we conceptualize node-wise calibration error as the calibration error of the confidence interval to which each node belongs. Consequently, the variance in calibration error related to individual nodes (the first scenario) is defined as the variance of all node-wise calibration errors.

As outlined in Table~\ref{Table_obs}, the variance within \textbf{Neig. Sim.} shows the lowest, particularly when compared to the variance across all nodes (\textbf{Node-wise}).
This demonstrates that nodes with comparable neighborhood predictions and confidence levels exhibit similar calibration error.

\subsection{\textsc{Simi-Mailbox}: A Topology-Grouping Strategy for Refining GNNs}\label{5.2}
\input{overall_framework}
Building on the observation discussed in previous section, \textsc{Simi-Mailbox} categorizes nodes by considering both neighborhood similarity and confidence levels.
We estimate the neighborhood similarity for each node $i$ by computing the average representational similarity with its neighbors, denoted as \textsc{mailbox} $\mathcal{M}^{simi}(i)$:
\begin{equation}
    \mathcal{M}^{simi}(i)=\frac{1}{|\mathcal{N}_i|}\sum_{j\in\mathcal{N}_i}\sigma(z_i^\mathsf{T}z_j),
\end{equation}
where $z_i$ represents the output logits for node $i$ from trained GNN, and $\sigma$ is a sigmoid function. 
Nodes with similar \textsc{mailbox} values and confidence levels are then grouped into $N$ distinct clusters. 
More precisely, \textsc{Simi-Mailbox} constructs a feature vector $F^{simi}_i = [\bar p_i, \;\bar{\mathcal{M}}^{simi}(i)]^\mathsf{T}$ for each node $i$, with the first dimension representing normalized confidence $\bar p_i$ and the second dimension representing a normalized \textsc{mailbox} value via min-max scaling. 
Subsequently, KMeans clustering is applied to $F^{simi}$ to construct $N$ similarity-based clusters $C=\{C_1,...,C_{N}\}$, ensuring the categorization adheres to both neighborhood similarity and confidence.

Once the categorization is completed, the original predictions for nodes within each cluster $C_n$ are scaled by a \textit{group-specific} temperature $\bm T_n$, a learnable parameter designed to rectify the miscalibration within the $n$-th cluster:
\begin{equation}
    \widetilde p_i = \max_k\sigma_{\text{sm}}\left(\frac{z_i}{\bm T_n}\right)_k\in\mathbb R,\quad i\in C_n.
\end{equation}
The group-wise temperature $\bm T\in\mathbb R^N$ is then optimized with a new loss $\mathcal{L}_{simi}$ with standard cross-entropy loss $\mathcal{L}_{\text{CE}}$:

\begin{equation}
    \begin{split}
        &\mathcal{L} = \mathcal{L}_{\text{CE}} + \lambda\mathcal{L}_{simi},\\
        &\mathcal{L}_{simi} = \sum_{n=1}^{N}||a_{val}^{(n)}-\frac{1}{|C_n|}\sum_{i\in C_n}\widetilde p_i||^2,
    \end{split}
\end{equation}
where $\lambda$ is a scaling factor for $\mathcal{L}_{simi}$. 
During calibration, $\mathcal L_{\text{CE}}$ encourages the reduction of entropy for correctly predicted classes while increasing it for the incorrectly predicted ones. 
In parallel, $\mathcal{L}_{simi}$ minimizes the discrepancy between the average scaled confidence of all nodes and the accuracy of validation nodes $a{val}^{(n)}$ within each cluster.
This approach directly adjusts the group-specific temperatures, with each $\bm T_n$ focused on minimizing the corresponding level of miscalibration.
The overall pipeline of \textsc{Simi-Mailbox} is illustrated in Figure~\ref{Fig: overall_framework}. 

\paragraph{On Accuracy Preservation.} The post-hoc group-wise temperatures in \textsc{Simi-Mailbox} ensures that the relative ordering of predictions remains unchanged. 
Let $f:\mathbb R^K\rightarrow\mathbb R^K$ as a calibration function and $z_i=[z_{i1},z_{i2},...,z_{iK}]^{\mathsf T}$ represents the logit vector for node $i$. We denote the group-specific temperature for the group to which node $i$ belongs as $\bm T_{g_i}$. Since the group-wise temperature $\bm T_{g_i}$ is uniformly applied to all elements of $z_i$, the order between elements in the calibrated logit $f_g(z_i)$ remains unchanged when subjected to the softmax operation $\sigma_{\text{sm}}$.
\begin{equation}
    \begin{split}
    &f_{g}(z_i)=[z_{i1}/\bm T_{g_i},z_{i2}/\bm T_{g_i},...,z_{iK}/\bm T_{g_i}],\\
    &\breve p_i=\sigma_{\text{sm}}\left(f_g(z_i)\right).
    \end{split}
\end{equation}
Thus, our method preserves the original classification accuracy, as the softmax function is order-preserving and scaling by $\bm T_{g_i}$ does not alter the relative ranking of logits.

\paragraph{Comparison with Prior Studies.} While our work shares the post-hoc temperature scaling framework with previous GNN calibration methods, \textsc{Simi-Mailbox} introduces group-specific temperatures \textit{independent of} node proximity or connectivity, thereby capturing high-level miscalibration patterns. Our method enables a more efficient optimization via $\mathcal L_{simi}$ due to its simplified number of parameters, compared to CaGCN and GATS requiring distinct temperatures for individual nodes. 
Moreover, the key distinction of our method lies in our discovery that nodes with similar neighborhood prediction similarity and confidence exhibit comparable calibration errors. 
This insight has not been explored in prior studies, as they do not fully consider the interplay between neighborhood similarity and confidence.

\section{Experiments}\label{6_experiments}
\input{Table1}
\input{Table2}
We validate the effectiveness of the proposed method under extensive experiments, leveraging two representative GNN architectures: GCN~\citep{kipf2016semi} and GAT~\citep{velivckovic2017graph}.
The performance of our \textsc{Simi-Mailbox} is evaluated across eight small- and medium-scale benchmark graphs adopted in \citep{hsu2022makes}: Cora, Citeseer, Pubmed~\citep{sen2008collective}, CoraFull~\citep{bojchevski2017deep}, Coauthor CS, Computers, and Photo~\citep{shchur2018pitfalls}. 
To further demonstrate the versatility, we extended our experiments to large-scale graphs, Arxiv~\citep{hu2020open} and Reddit~\citep{zeng2019graphsaint}. More experiments including comparison with recent baselines, evaluations on heterophilous graphs and other GNN backbones, and hyperparameter robustness are provided in the Appendix.

\paragraph{Baselines.} In alignment with precedent studies, we compare our method against classical calibration methods: temperature scaling (TS), vector scaling (VS)~\citep{guo2017calibration}, and ensemble temperature sclaing~(ETS)~\citep{zhang2020mix} and GNN-specialized calibration baselines: CaGCN~\citep{wang2021confident} and GATS~\citep{hsu2022makes}.
We provide an additional experiments to compare \textsc{Simi-Mailbox} and GPN~\citep{stadler2021graph} and GNNSafe~\citep{wu2023energy} for out-of-detection task in the Appendix.

\paragraph{Experimental Setup.} We undertake our experiments following the experimental protocols of GATS~\citep{hsu2022makes} in the scope of semi-supervised node classification. Details of the experiment configurations are provided in the Appendix. 
To assess the calibration performance, we use ECE as a principal metric~\citep{naeini2015obtaining}, following the common practice~\citep{wang2021confident, hsu2022makes}.
The optimal calibration models are chosen based on the lowest validation ECE on training set. 
Additional calibration metrics, including class-wise ECE~\citep{kull2019beyond, nixon2019measuring}, Kernel Density Estimation-based ECE~\citep{zhang2020mix}, Brier Score~\citep{brier1950verification}, and Negative Log-likelihood, are provided in the Appendix.

\paragraph{Results on Small- and Medium-scale Graphs.} 
Table~\ref{Table1} shows that \textsc{Simi-Mailbox} outperforms baselines in 15 of 16 settings. 
Notably, our method pioneers in achieving an error rate below $3\%$ on Cora and Citeseer datasets, with a significant lead on Cora using GCN, breaking into the $1\%$ error range. 
\textsc{Simi-Mailbox} also demonstrates marked improvements on Pubmed and CS datasets, first achieving ECE reductions to within the $[0.5, 0.8]$ range. 
Even on Computers and Photo datasets, where the original predictions are already well-calibrated, \textsc{Simi-Mailbox} further reduces calibration errors to below $1\%$ with GAT. 
Additionally, consistent improvement is observed on the CoraFull dataset, with our method achieving the first 2\% error range using GAT.


\input{qualitative_analysis}
\paragraph{Results on Large-scale Graphs.}
To further demonstrate the versatility of our method, we extended our experiments to large-scale graphs, following the evaluation protocol in \citep{hu2020open}. 
We employed GCN and GraphSAGE (SAGE)\citep{hamilton2017inductive}, which are representative architectures for large-scale benchmark datasets. 
As shown in Table\ref{Table2-1}, \textsc{Simi-Mailbox} outperforms all baselines to a considerable extent, achieving an error rate below 1\% in all examined settings. This superiority is particularly notable in the Reddit dataset with SAGE, where our method reduces miscalibration by 10.57\% compared to the uncalibrated baseline.
In addition to calibration performance, we also measured the total execution time for each run, as presented in Table~\ref{Table2-2}. Our method significantly improves time efficiency across all experiments, with a notable reduction in execution time on the Reddit dataset, decreasing by 61.86 and 182.10 seconds with GCN and SAGE. This gain is attributed to the simplified group-wise temperature approach, which allows for rapid optimization with only few parameters ($N$ clusters), in contrast to baselines that rely on complex GNNs for deriving node-wise temperature.

\paragraph{\textsc{Simi-Mailbox} on Self-training.}
In addition to improving calibration, calibrated predictions can be applied in self-training, utilizing pseudo-labels generated from unlabeled samples. As evidenced by~\citep{rizve2021defense}, poorly-calibrated models have a risk to choose pseudo-labeled samples with high confidence but incorrect classifications. Hence, confidence adjusted through calibration methods can lead to the selection of more accurate and high-confidence samples, improving classification accuracy. We broaden our evaluation of \textsc{Simi-Mailbox} to self-training scenarios, initially explored in CaGCN. 
Adhering to the same evaluation protocol in \citep{wang2021confident}, we validate the effectiveness of our method in generating qualified pseudo-labels over baselines. Detailed results of this experiment are provided in the Appendix. 


\paragraph{Effectiveness on Diverse Neighborhood Topology.}
To further validate the effectiveness of our method across different levels of neighborhood similarity, we present a qualitative comparison in Figure~\ref{Fig: qualitative_analysis}, utilizing a consistent dataset (CoraFull) and architecture (GCN) in preceding section. Similar to earlier analyses, the x-axis partitions nodes into intervals based on neighborhood prediction similarity, while the y-axis categorizes them by confidence intervals. Each cell in the heatmap represents the subtraction of average confidence of calibrated nodes from the accuracy. Deeper shades of \textbf{red} indicate that calibrated confidence is lower than accuracy (\textbf{under-confident}), while deeper shades of \textbf{green} signify that confidence exceeds accuracy (\textbf{over-confident}). Ideally, a perfectly calibrated model would produce a uniformly white heatmap, indicating perfect alignment between confidence and accuracy.
As illustrated, \textsc{Simi-Mailbox} significantly reduces the discrepancy between accuracy and average confidence across varying similarity levels compared to baseline methods. This improvement is particularly evident in the patterns identified in the previous analysis, where our method mitigates discrepancies in the $\bm s(i)\in(0.1,0.2]$ range within the $\hat p_i\in(0.9,1.0]$ interval for CaGCN, and addresses the prevalent under-confidence observed with GATS in the $\bm s(i)\in(0.0,0.4]$ range within the $\hat p_i\in (0.6,0.8]$ intervals. 

\section{Conclusion}\label{7_conclusion}
In this study, we presented a novel analysis that identifies the limitations of uniform design principles in existing GNN calibration methods, particularly based on neighborhood similarity. 
To address these limitations, we proposed \textsc{Simi-Mailbox}, a novel calibration method that employs group-specific temperatures to refine miscalibration in nodes categorized by both neighborhood similarity and confidence. 
Comprehensive experiments have demonstrated the effectiveness of \textsc{Simi-Mailbox}, supported by extensive empirical and technical analysis. 
As for future work, we are dedicated to developing a theoretical foundation for our method.

\section{Acknowledgment}
This work was supported by Institute for Information $\&$ communications Technology Planning $\&$ Evaluation(IITP) grant funded by the Korea government(MSIT) (RS-2019-II190075, Artificial Intelligence Graduate School Program(KAIST)). We sincerely appreciate Polymerize for their generous support of our manuscript. Special thanks also go to Changhun Kim and Taewon Kim for constructive comments on this manuscript.

\bibliography{aaai25}

\appendix
\newpage
\input{appendix}

\end{document}

%% file: macro.tex
\definecolor{customred}{RGB}{192, 0, 0}

\makeatletter
\newcommand*\rel@kern[1]{\kern#1\dimexpr\macc@kerna}
\newcommand*\widebar[1]{%
  \begingroup
  \def\mathaccent##1##2{%
    \rel@kern{0.8}%
    \overline{\rel@kern{-0.8}\macc@nucleus\rel@kern{0.2}}%
    \rel@kern{-0.2}%
  }%
  \macc@depth\@ne
  \let\math@bgroup\@empty \let\math@egroup\macc@set@skewchar
  \mathsurround\z@ \frozen@everymath{\mathgroup\macc@group\relax}%
  \macc@set@skewchar\relax
  \let\mathaccentV\macc@nested@a
  \macc@nested@a\relax111{#1}%
  \endgroup
}
\makeatother

%% file: obs_plot.tex
\begin{figure*}[t]
    \centering
    \includegraphics[width=.9\linewidth]{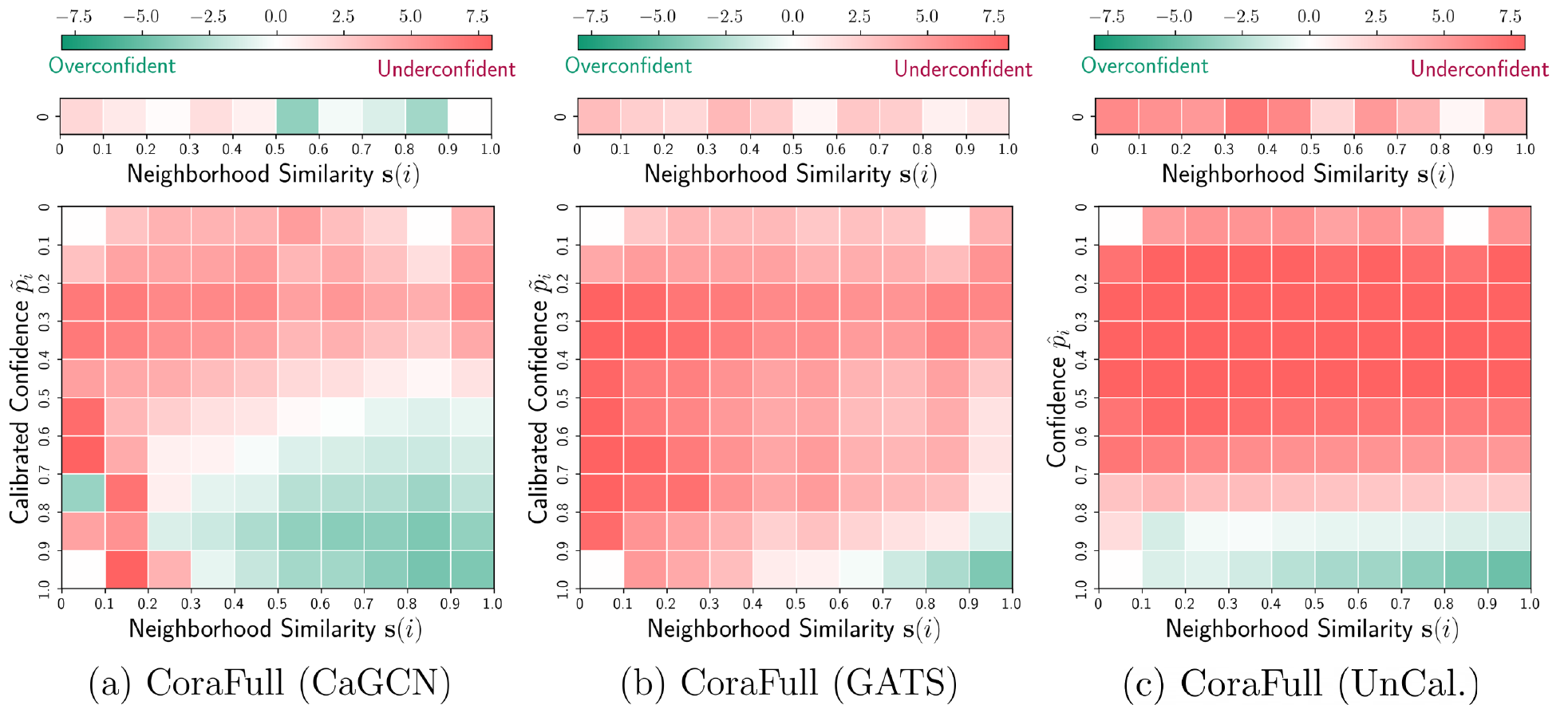}
    \vskip -5pt
    \caption{Analysis of uncalibrated and calibrated logits via prior works, CaGCN and GATS. The $x$-axis divides nodes into sub-intervals based on neighborhood similarity, while the $y$-axis represents corresponding confidence intervals. 
    Each cell in the heatmap represents the subtraction of the average confidence from the accuracy, with color intensity indicating the magnitude of this discrepancy.
    Contrary to the uniform assumptions in prior works on neighborhood similarity, the results demonstrate that calibration errors can significantly differ among nodes with comparable neighborhood similarity but different confidence levels. Moreover, prior approaches exhibit sub-optimal calibration across varying neighborhood similarity levels when predictions are extended across confidence intervals.}
    \label{Fig: obs_plot}
    \vskip -10pt
\end{figure*}

%% file: Table_obs.tex
{\renewcommand{\arraystretch}{1.2}
  \begin{table}[h]
  \centering
    \caption{Variance of calibration errors ($\mathbf{\times 100}$) involving neighborhood similarity sub-intervals (Neig. Sim.), confidence intervals (Conf), and total nodes (Node-wise).}
    \begin{center}
    \resizebox{\columnwidth}{!}{
    \Huge
    \begin{tabular}{ c | c | c | c | c | c | c | c | c | c }
    
    \hline\hline
     \multicolumn{2}{ c |}{\textbf{GNNs}} & {{Cora}} & {{Citeseer}} & {{Pubmed}} & {{Computers}} & {{Photo}} & {{CS}} & {{Physics}} & {{CoraFull}}\\
     \hline
     \multirow{3}{*}{GCN} & 
     {Node-wise}
     & {\fontsize{25}{26}\selectfont{6.139}} 
     & {\fontsize{25}{26}\selectfont{1.957}}
     & {\fontsize{25}{26}\selectfont{1.370} }
     & {\fontsize{25}{26}\selectfont{40.370}} 
     & {\fontsize{25}{26}\selectfont{7.200}}
     & {\fontsize{25}{26}\selectfont{31.470} }
     & {\fontsize{25}{26}\selectfont{3.312}}
     & {\fontsize{25}{26}\selectfont{34.560}}\\
     
     & {Conf.}
     & {\fontsize{25}{26}\selectfont{0.065} }
     & {\fontsize{25}{26}\selectfont{0.060}}
     & {\fontsize{25}{26}\selectfont{0.068} }
     & {\fontsize{25}{26}\selectfont\textbf{0.060} }
     & {\fontsize{25}{26}\selectfont{0.052} }
     &{ \fontsize{25}{26}\selectfont{0.058}}
     & {\fontsize{25}{26}\selectfont{0.041} }
     & {\fontsize{25}{26}\selectfont{0.064}}\\
     
     & {{Neig. Sim.}}
     & {\fontsize{25}{26}\selectfont\textbf{0.057} }
     & {\fontsize{25}{26}\selectfont\textbf{0.046}}
     & {\fontsize{25}{26}\selectfont\textbf{0.061} }
     & {\fontsize{25}{26}\selectfont{0.062} }
     & {\fontsize{25}{26}\selectfont\textbf{0.047} }
     & {\fontsize{25}{26}\selectfont\textbf{0.052}}
     & {\fontsize{25}{26}\selectfont\textbf{0.040} }
     & {\fontsize{25}{26}\selectfont\textbf{0.057}}\\

     \hline
     
     \multirow{3}{*}{GAT} & {Node-wise}
     & {\fontsize{25}{26}\selectfont{8.656} }
     & {\fontsize{25}{26}\selectfont{2.570}}
     & {\fontsize{25}{26}\selectfont{1.614} }
     & {\fontsize{25}{26}\selectfont{44.980} }
     & {\fontsize{25}{26}\selectfont{14.550}} 
     & {\fontsize{25}{26}\selectfont{42.660}}
     & {\fontsize{25}{26}\selectfont{3.346} }
     & {\fontsize{25}{26}\selectfont{58.67}}\\
     
     & {Conf.}
     & {\fontsize{25}{26}\selectfont{0.068} }
     & {\fontsize{25}{26}\selectfont{0.062}}
     & {\fontsize{25}{26}\selectfont{0.068} }
     & {\fontsize{25}{26}\selectfont{0.048} }
     & {\fontsize{25}{26}\selectfont{0.053} }
     & {\fontsize{25}{26}\selectfont{0.050}}
     & {\fontsize{25}{26}\selectfont\textbf{0.035} }
     & {\fontsize{25}{26}\selectfont{0.062}}\\
     
     & {Neig. Sim.}
     & {\fontsize{25}{26}\selectfont\textbf{0.056} }
     & {\fontsize{25}{26}\selectfont\textbf{0.044}}
     & {\fontsize{25}{26}\selectfont\textbf{0.055} }
     & {\fontsize{25}{26}\selectfont\textbf{0.045} }
     & {\fontsize{25}{26}\selectfont\textbf{0.047} }
     & {\fontsize{25}{26}\selectfont\textbf{0.044}}
     & {\fontsize{25}{26}\selectfont\textbf{0.035} }
     & {\fontsize{25}{26}\selectfont\textbf{0.055} }\\
    \hline\hline
    
    \end{tabular}}
    \label{Table_obs}
    \end{center}
    \vskip -8pt
\end{table}
}

%% file: overall_framework.tex
\begin{figure}[h]
    \centering
    \includegraphics[width=1.\linewidth]{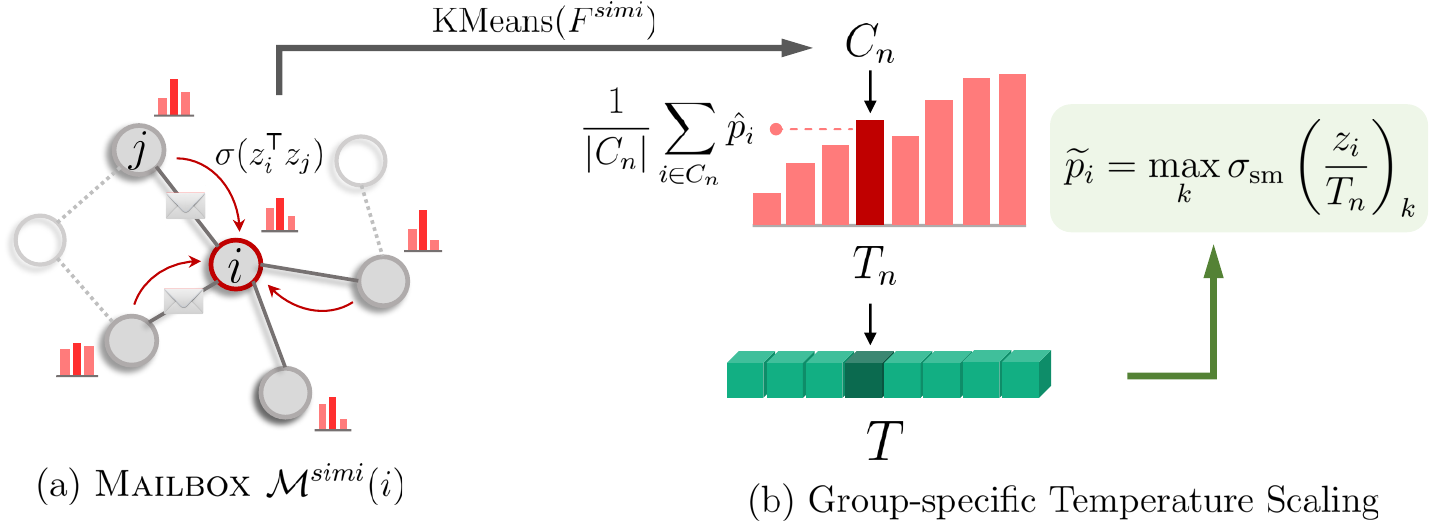}
    \caption{Overall framework of our \textsc{Simi-Mailbox}. 
    }
    \label{Fig: overall_framework}
    \vskip -8pt
\end{figure}

%% file: Table1.tex
{\renewcommand{\arraystretch}{1.02}
  \begin{table*}[t]
    \caption{ECE results (reported in percentage) for our proposed calibration method and baselines. A lower ECE indicates better calibration performance. The best and second best performances are represented by bold and underline texts.}
    \centering
    \resizebox{.99\textwidth}{!}{
    \huge{
    \begin{tabular}{ c | c | c | c | c | c | c | c | c }
     \hline\hline
     \multicolumn{2}{ c |}{\textbf{Methods}} & {{UnCal.}} & {{TS}} & {{VS}} & {{ETS}} & {{CaGCN}} & {{GATS}} & {{\textbf{Ours}}}\\
     \hline
     \hline
     \multirow{2}{*}{Cora} & GCN & 
     {12.43 \fontsize{18}{19}\selectfont$\pm$ 4.24} & 
     {3.87 \fontsize{18}{19}\selectfont$\pm$ 1.22} & 
     {4.30 \fontsize{18}{19}\selectfont$\pm$ 1.28} & 
     {3.78 \fontsize{18}{19}\selectfont$\pm$ 1.25} & 
     {5.22 \fontsize{18}{19}\selectfont$\pm$ 1.45} & 
     \underline{3.55 \fontsize{18}{19}\selectfont$\pm$ 1.28} & 
     \textbf{{1.97 \fontsize{18}{19}\selectfont$\pm$ 0.44}}\\
     & GAT & 
     {14.88 \fontsize{18}{19}\selectfont$\pm$ 4.30} &
     {3.42 \fontsize{18}{19}\selectfont$\pm$ 1.00} & 
     {3.45 \fontsize{18}{19}\selectfont$\pm$ 1.13} & 
     {3.32 \fontsize{18}{19}\selectfont$\pm$ 0.92} & 
     {3.81 \fontsize{18}{19}\selectfont$\pm$ 1.00} & 
     \underline{3.05 \fontsize{18}{19}\selectfont$\pm$ 0.78} & 
     \textbf{{2.08 \fontsize{18}{19}\selectfont$\pm$ 0.45}}\\
     \hline
     
     \multirow{2}{*}{Citeseer} & GCN & 
     {12.54 \fontsize{18}{19}\selectfont$\pm$ 8.58} & 
     {5.27 \fontsize{18}{19}\selectfont$\pm$ 1.70} & 
     {5.15 \fontsize{18}{19}\selectfont$\pm$ 1.46} & 
     {5.10 \fontsize{18}{19}\selectfont$\pm$ 1.76} & 
     {6.60 \fontsize{18}{19}\selectfont$\pm$ 1.76} & 
     \underline{4.49 \fontsize{18}{19}\selectfont$\pm$ 1.53} & 
     \textbf{{2.66 \fontsize{18}{19}\selectfont$\pm$ 0.53}}\\
     & GAT & 
     {16.65 \fontsize{18}{19}\selectfont$\pm$ 7.98} &
     {5.08 \fontsize{18}{19}\selectfont$\pm$ 1.48} & 
     {4.62 \fontsize{18}{19}\selectfont$\pm$ 1.58} & 
     {5.01 \fontsize{18}{19}\selectfont$\pm$ 1.46} & 
     {4.86 \fontsize{18}{19}\selectfont$\pm$ 1.68} & 
     \underline{4.01 \fontsize{18}{19}\selectfont$\pm$ 1.42} & 
     \textbf{{2.86 \fontsize{18}{19}\selectfont$\pm$ 0.56}}\\
     \hline
     
     \multirow{2}{*}{Pubmed} & GCN & 
     {7.30 \fontsize{18}{19}\selectfont$\pm$ 1.56} & 
     {1.27 \fontsize{18}{19}\selectfont$\pm$ 0.30} & 
     {1.46 \fontsize{18}{19}\selectfont$\pm$ 0.29} & 
     {1.26 \fontsize{18}{19}\selectfont$\pm$ 0.31} & 
     {1.05 \fontsize{18}{19}\selectfont$\pm$ 0.33} & 
     \underline{0.95 \fontsize{18}{19}\selectfont$\pm$ 0.32} & 
     \textbf{{0.75 \fontsize{18}{19}\selectfont$\pm$ 0.15}}\\
     & GAT & 
     {10.38 \fontsize{18}{19}\selectfont$\pm$ 1.89} &
     {1.15 \fontsize{18}{19}\selectfont$\pm$ 0.46} & 
     {1.05 \fontsize{18}{19}\selectfont$\pm$ 0.36} & 
     {1.13 \fontsize{18}{19}\selectfont$\pm$ 0.47} & 
     {0.99 \fontsize{18}{19}\selectfont$\pm$ 0.34} & 
     \underline{0.98 \fontsize{18}{19}\selectfont$\pm$ 0.36} & 
     \textbf{0.69 \fontsize{18}{19}\selectfont$\pm$ 0.16}\\
     \hline
     
     \multirow{2}{*}{Computers} & GCN & 
     {2.96 \fontsize{18}{19}\selectfont$\pm$ 0.76} & 
     {2.62 \fontsize{18}{19}\selectfont$\pm$ 0.55} & 
     {2.70 \fontsize{18}{19}\selectfont$\pm$ 0.61} & 
     {2.59 \fontsize{18}{19}\selectfont$\pm$ 0.72} & 
     \underline{1.70 \fontsize{18}{19}\selectfont$\pm$ 0.53} & 
     {2.15 \fontsize{18}{19}\selectfont$\pm$ 0.52} & 
     \textbf{{1.02 \fontsize{18}{19}\selectfont$\pm$ 0.26}}\\
     & GAT & 
     {1.58 \fontsize{18}{19}\selectfont$\pm$ 0.56} &
     {1.44 \fontsize{18}{19}\selectfont$\pm$ 0.35} & 
     {1.44 \fontsize{18}{19}\selectfont$\pm$ 0.40} & 
     {1.42 \fontsize{18}{19}\selectfont$\pm$ 0.43} & 
     {1.82 \fontsize{18}{19}\selectfont$\pm$ 0.63} & 
     \underline{1.36 \fontsize{18}{19}\selectfont$\pm$ 0.34} & 
     \textbf{{0.95 \fontsize{18}{19}\selectfont$\pm$ 0.37}}\\
     \hline

     \multirow{2}{*}{Photo} & GCN & 
     {2.11 \fontsize{18}{19}\selectfont$\pm$ 0.97} & 
     {1.68 \fontsize{18}{19}\selectfont$\pm$ 0.68} & 
     {1.75 \fontsize{18}{19}\selectfont$\pm$ 0.67} & 
     {1.63 \fontsize{18}{19}\selectfont$\pm$ 0.84} & 
     {1.98 \fontsize{18}{19}\selectfont$\pm$ 0.53} & 
     \underline{1.46 \fontsize{18}{19}\selectfont$\pm$ 0.51} & 
     \textbf{{1.01 \fontsize{18}{19}\selectfont$\pm$ 0.36}}\\
     & GAT & 
     {2.18 \fontsize{18}{19}\selectfont$\pm$ 1.54} &
     {1.56 \fontsize{18}{19}\selectfont$\pm$ 0.63} & 
     {1.65 \fontsize{18}{19}\selectfont$\pm$ 0.70} & 
     {1.57 \fontsize{18}{19}\selectfont$\pm$ 0.78} & 
     {2.04 \fontsize{18}{19}\selectfont$\pm$ 0.74} & 
     \underline{1.49 \fontsize{18}{19}\selectfont$\pm$ 0.65} & 
     \textbf{{0.97 \fontsize{18}{19}\selectfont$\pm$ 0.53}}\\
     \hline

     \multirow{2}{*}{CS} & GCN & 
     {1.72 \fontsize{18}{19}\selectfont$\pm$ 1.28} & 
     {1.01 \fontsize{18}{19}\selectfont$\pm$ 0.24} & 
     {0.94 \fontsize{18}{19}\selectfont$\pm$ 0.28} & 
     {0.97 \fontsize{18}{19}\selectfont$\pm$ 0.22} & 
     {2.32 \fontsize{18}{19}\selectfont$\pm$ 1.12} & 
     \underline{0.90 \fontsize{18}{19}\selectfont$\pm$ 0.29} & 
     \textbf{{0.58 \fontsize{18}{19}\selectfont$\pm$ 0.19}}\\
     & GAT & 
     {1.48 \fontsize{18}{19}\selectfont$\pm$ 0.79} &
     {1.07 \fontsize{18}{19}\selectfont$\pm$ 0.34} & 
     {1.01 \fontsize{18}{19}\selectfont$\pm$ 0.40} & 
     {1.03 \fontsize{18}{19}\selectfont$\pm$ 0.31} & 
     {2.27 \fontsize{18}{19}\selectfont$\pm$ 1.13} & 
     \underline{0.85 \fontsize{18}{19}\selectfont$\pm$ 0.23} & 
     \textbf{{0.72 \fontsize{18}{19}\selectfont$\pm$ 0.43}}\\
     \hline

     \multirow{2}{*}{Physics} & GCN & 
     {0.56 \fontsize{18}{19}\selectfont$\pm$ 0.33} & 
     {0.51 \fontsize{18}{19}\selectfont$\pm$ 0.19} & 
     {0.46 \fontsize{18}{19}\selectfont$\pm$ 0.15} & 
     {0.51 \fontsize{18}{19}\selectfont$\pm$ 0.19} & 
     {0.88 \fontsize{18}{19}\selectfont$\pm$ 0.47} & 
     \underline{0.45 \fontsize{18}{19}\selectfont$\pm$ 0.15} & 
     \textbf{{0.28 \fontsize{18}{19}\selectfont$\pm$ 0.11}}\\
     & GAT & 
     {0.55 \fontsize{18}{19}\selectfont$\pm$ 0.24} & 
     {0.56 \fontsize{18}{19}\selectfont$\pm$ 0.20} & 
     {0.56 \fontsize{18}{19}\selectfont$\pm$ 0.21} & 
     {0.55 \fontsize{18}{19}\selectfont$\pm$ 0.20} & 
     {1.06 \fontsize{18}{19}\selectfont$\pm$ 0.40} & 
     \textbf{{0.43 \fontsize{18}{19}\selectfont$\pm$ 0.16}} & 
     \underline{0.48 \fontsize{18}{19}\selectfont$\pm$ 0.22}\\
     \hline

     \multirow{2}{*}{CoraFull} & GCN & 
     {6.49 \fontsize{18}{19}\selectfont$\pm$ 1.28} &
     {5.55 \fontsize{18}{19}\selectfont$\pm$ 0.45} & 
     {5.79 \fontsize{18}{19}\selectfont$\pm$ 0.43} & 
     {5.49 \fontsize{18}{19}\selectfont$\pm$ 0.46} & 
     {5.92 \fontsize{18}{19}\selectfont$\pm$ 2.84} & 
     \underline{3.74 \fontsize{18}{19}\selectfont$\pm$ 0.63} & 
     \textbf{{3.46 \fontsize{18}{19}\selectfont$\pm$ 1.31}}\\
     & GAT & 
     {5.25 \fontsize{18}{19}\selectfont$\pm$ 1.32} &
     {4.41 \fontsize{18}{19}\selectfont$\pm$ 0.50} & 
     {4.42 \fontsize{18}{19}\selectfont$\pm$ 0.49} & 
     {4.36 \fontsize{18}{19}\selectfont$\pm$ 0.50} & 
     {6.80 \fontsize{18}{19}\selectfont$\pm$ 3.81} & 
     \underline{3.46 \fontsize{18}{19}\selectfont$\pm$ 0.46} & 
     \textbf{{2.64 \fontsize{18}{19}\selectfont$\pm$ 1.02}}\\
    \hline\hline
    \end{tabular}}}
    \label{Table1}
    \vskip -5pt
  \end{table*}
}

%% file: Table2.tex
{\renewcommand{\arraystretch}{1.1}
\begin{table*}[t]
    \begin{minipage}{.47\textwidth}
    \caption{ECE results (in percentage) for our method and baselines on large-scale datasets.}
    \centering
    \resizebox{\textwidth}{!}{
    \huge{
    \begin{tabular}{ c | c | c | c | c | c }
          \hline\hline
          \multicolumn{2}{ c |}{\textbf{Methods}} & {{UnCal.}} & {{CaGCN}} & {{GATS}} & {{\textbf{Ours}}}\\
          \hline\hline
          \multirow{2}{*}{Arxiv} & GCN &
          4.92 \fontsize{18}{19}\selectfont$\pm$ 0.36 & 
          1.97 \fontsize{18}{19}\selectfont$\pm$ 0.16 & 
          0.75 \fontsize{18}{19}\selectfont$\pm$ 0.06 & 
          \textbf{0.71 \fontsize{18}{19}\selectfont$\pm$ 0.13}\\
          & SAGE &
          3.00 \fontsize{18}{19}\selectfont$\pm$ 0.89 &
          1.84 \fontsize{18}{19}\selectfont$\pm$ 0.19 & 
          2.05 \fontsize{18}{19}\selectfont$\pm$ 0.28 & 
          \textbf{0.98 \fontsize{18}{19}\selectfont$\pm$ 0.23}\\
          \hline
          
          \multirow{2}{*}{Reddit} & GCN &
          8.55 \fontsize{18}{19}\selectfont$\pm$ 1.28 & 
          1.86 \fontsize{18}{19}\selectfont$\pm$ 0.19 & 
          2.56 \fontsize{18}{19}\selectfont$\pm$ 0.59 & 
          \textbf{0.35 \fontsize{18}{19}\selectfont$\pm$ 0.05}\\
          & SAGE &
          11.30 \fontsize{18}{19}\selectfont$\pm$ 1.99 &
          2.14 \fontsize{18}{19}\selectfont$\pm$ 0.35 & 
          4.66 \fontsize{18}{19}\selectfont$\pm$ 0.57 & 
          \textbf{0.73 \fontsize{18}{19}\selectfont$\pm$ 0.15}\\
          \hline\hline
    \end{tabular}
    \label{Table2-1}
    }
    }
    \end{minipage}%
    \hfill
    \begin{minipage}{.5\textwidth}
    \caption{Calibration duration (in seconds) for our method and baselines on large-scale datasets.}
    \centering
    \resizebox{\textwidth}{!}{
    \Huge{
    \begin{tabular}{ c | c | c | c | c }
          \hline\hline
          \multicolumn{2}{ c |}{\textbf{Methods}} & {{CaGCN}} & {{GATS}} & {{\textbf{Ours}}}\\
          \hline\hline
          \multirow{2}{*}{Arxiv} & GCN & 
          20.84 \fontsize{17}{18}\selectfont$\pm$ 2.69 &
          48.89 \fontsize{17}{18}\selectfont$\pm$ 11.39 &
          \textbf{7.10 \fontsize{17}{18}\selectfont$\pm$ 0.94} \textbf{\blue{(-41.79 sec)}}\\
          & SAGE & 
          23.02 \fontsize{17}{18}\selectfont$\pm$ 4.44 &
          61.67 \fontsize{17}{18}\selectfont$\pm$ 16.89 &
          \textbf{4.85 \fontsize{17}{18}\selectfont$\pm$ 0.65} \textbf{\blue{(-56.82 sec)}}\\
          \hline
          
          \multirow{2}{*}{Reddit} & GCN &
          55.98 \fontsize{17}{18}\selectfont$\pm$ 13.76 &
          72.90 \fontsize{17}{18}\selectfont$\pm$ 19.98 &
          \textbf{11.04 \fontsize{17}{18}\selectfont$\pm$ 0.30} \textbf{\blue{(-61.86 sec)}}\\
          & SAGE & 
          78.13 \fontsize{17}{18}\selectfont$\pm$ 27.35 &
          192.01 \fontsize{17}{18}\selectfont$\pm$ 177.57 &
          \textbf{9.91 \fontsize{17}{18}\selectfont$\pm$ 0.95} \textbf{\blue{(-182.1 sec)}}\\
          \hline\hline
    \end{tabular}
    }
    }
    \label{Table2-2}
    \end{minipage}
    \vskip -10pt
\end{table*}
}

%% file: qualitative_analysis.tex
\begin{figure*}[t!]
    \centering
    \includegraphics[width=.9\linewidth]{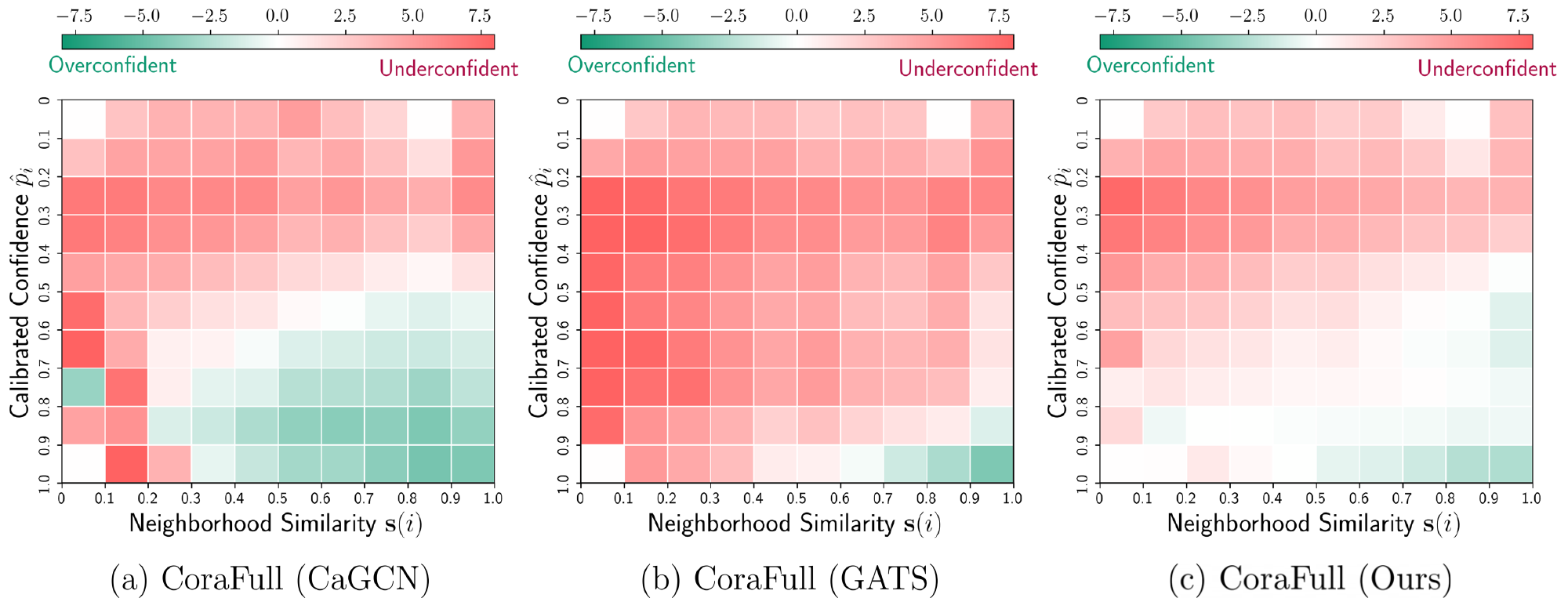}
    \caption{Qualitative analysis of our calibration results on CoraFull dataset, compared with CaGCN and GATS. 
    Each cell in the heatmap represents the subtraction of the average confidence of calibrated nodes from the accuracy, with color and intensity indicating the magnitude of this discrepancy.
    Throughout diverse neighborhood similarity levels, our method facilitates a better reduction in the gap between accuracy and confidence compared to baselines.
    }
    \label{Fig: qualitative_analysis}
    \vskip -10pt
\end{figure*}

%% file: appendix.tex
\appendix
\section*{Supplementary Materials}



%
\section{Detailed Experimental Setup}\label{A}

\subsection{Dataset Statistics}\label{A.1}
Table~\ref{Table_statistics} provides comprehensive statistics of the datasets used in our experiments, including the number of nodes, edges, classes, and features.
\input{Table_statistics}

\subsection{Implementation Details}\label{A.2}
We implement GNN models and the proposed method using PyTorch~\citep{NEURIPS2019_pytorch} and PyTorch Geometric~\citep{fey2019fast}. 
The experiments are conducted on RTX 2080ti (11GB) and RTX 3090ti GPU (24G). 
The experimental settings and evaluation protocols are largely consistent with those used in the GATS framework~\citep{hsu2022makes}.
We split the labeled and unlabeled data by 15\% and 85\%, respectively, and perform three-fold cross-validation on the former, 10\% for training and the remaining 5\% for validation. 
We conduct 75 runs in total for each experiment, considering 5 random splits, 5 random initializations, and three-fold cross-validation.
For optimization, we select the Adam optimizer with an initial learning rate of 0.01. The weight decay values are set to 5e-4 for Cora, Citeseeer, and Pubmed and 0 for the remaining datasets.
The GNN architectures are configured as follows: for GCN~\citep{kipf2016semi}, we employ two GNN layers with 64 hidden units; for GAT~\citep{velivckovic2017graph}, the number of attention heads is set to 8, with 8 hidden units per head.
We train GNNs in a maximum of 2000 epochs with early stopping based on a patience of 100 epochs.
A dropout rate of 0.5 is applied uniformly across all backbones.
To ensure fair comparisons, we refer to the implementation and setup details provided in the released code of GATS~\citep{hsu2022makes} for the baseline methods. 
In our method, the greedy KMeans++ algorithm (as implemented under the KMeans in the scikit-learn) is chosen for clustering. The number of clusters $N$ is selected from the range $[5, 30]$.
The regularization coefficient $\lambda$ is explored in the range of $[1, 50]$
During the evaluation, we set the number of bins for ECE and class-wise ECE measurements as $15$, following prior works. 

For experiments on large-scale datasets with GCN and SAGE~\citep{hamilton2017inductive}, we adhere to the same split ratio and evaluation protocol in~\citep{hu2020open}, and report ECE results based on the best validation performance, averaged over ten random seeds.
We employ three-layer and two-layer GNN on Arxiv and Reddit, respectivly, while fixing the hidden dimension as 256 across both GCN and SAGE.
Analogous to the regular-scale experiment, we select the Adam optimizer with an initial learning rate of 0.01 and weight decay as 0 uniformly across all large-scale settings.
Meanwhile, we construct hyperparameter search spaces for proposed \textsc{Simi-Mailbox} as follows. 

For the comparison between GPN~\citep{stadler2021graph} and \textsc{Simi-Mailbox}, we used the benchmark datasets and configurations from \citet{stadler2021graph}. In the out-of-distribution detection experiments, we followed the same experimental setup as GNNSafe~\citep{wu2023energy}.
The source code of \textsc{Simi-Mailbox} is provided via anonymous link: \url{https://anonymous.4open.science/r/Simi_Mailbox-04B4}


\input{Table_pce2}
\input{Table_gpn}
\input{sensitivity_all}
\input{Table_scaling}

\section{Further Discussions and Experiments}\label{B}
\subsection{Algorithmic Point of View on Previous GNN Calibration Studies}\label{B.1}
This subsection provides the limitation of previous GNN calibration approaches for addressing varying similarity levels in the algorithmic perspective. To begin with, the node-wise temperature $\bm T^{\text{CaGCN}}$ for $l$ layers in CaGCN~\citep{wang2021confident} is defined as below:
\begin{equation}     
\begin{split}
\bm T^{\text{CaGCN}} = &\sigma^+(\bm A\sigma_{\text{ReLU}}(...\bm A\sigma_{\text{ReLU}}(\bm A\bm ZW^{(1)})\\
&W^{(2)}...)W^{(l)})\in\mathbb R^{|\mathcal{V}|},
\end{split}
\end{equation}
where $\sigma^+$ and $\sigma_{\text{ReLU}}$ denote softplus and ReLU operation, while $\bm Z$ and $W$ represent logits from trained GNNs and trainable weights, respectively.

The core assumption behind CaGCN’s use of GCN as a temperature function is that the confidence of nodes connected to similar neighbors should increase, while the confidence of those with dissimilar neighbors should decrease. CaGCN argues that GCN can achieve such confidence adjustment by propagating predictions to their neighbors. However, our findings suggest that this assumption does not hold universally. Moreover, based on the formulation, the temperature function does not consistently produce higher temperatures for nodes with dissimilar neighbors or lower temperatures for those with similar neighbors. 
Consequently, this may lead to vague and inconsistent calibration outcomes.

Meanwhile, for an individual node $i$, the temperature function in GATS~\citep{hsu2022makes}, denoted as $\bm T^{\text{GATS}}_i$, is formulated as:
\begin{equation}
\bm T^{\text{GATS}}_i=\frac{1}{H}\sum_{h=1}^H\sigma^+(\omega\,\delta\hat c_i+\sum_{j\in\mathcal{N}_i}\alpha_{ij}\gamma_j\tau_j^h) + T_0\in\mathbb R.
\end{equation}
Here, $H$ represents the number of attention heads, and $T_0$is an initial bias term. The coefficient $\omega$ is a learnable parameter that scales the relative confidence $\delta\hat c_i$ of node $i$ against its neighborhood. The scaling factor $\gamma$ leverages the distance-to-training-nodes property, while $\tau_j$ refers to the logits $z_j$ of neighboring nodes, transformed by a linear layer and then sorted class-wise within each node's logits.

GATS introduces the attention coefficient $\alpha_{ij}$ to account for the observed increase in calibration error with decreasing representational similarity. While $\alpha_{ij}$ is designed to capture the affinity between nodes, the model's ability to accurately adjust for varying levels of similarity is limited. This limitation arises because the complex integration of factors, such as the initial bias term $T_0$ and $\omega\delta\hat c_i$, may obscure the influence of neighborhood similarity. As a result, the model might not adequately address the distinct calibration needs of nodes in different similarity contexts.

\subsection{Contribution of Loss Components in \textsc{Simi-Mailbox}}\label{B.2}

In this section, we provide a comprehensive explanation on distinct roles of our loss functions, $\mathcal L_{\text{CE}}$ and $\mathcal L_{simi}$. Specifically, $\mathcal L_{simi}$ is designed to directly minimize the discrepancy between per-bin average confidence and accuracy. In contrast, $\mathcal L_{\text{CE}}$ targets minimizing the prediction entropy. 

The important component of our method is the minimization of $\mathcal L_{simi}$, a direct objective that ensures each temperature focuses on reducing specific calibration errors within individual groups, thereby guiding the optimal adjustment of group-specific temperatures. Previous GNN calibration approaches~\citep{wang2021confident, hsu2022makes} employed GNN architectures to design node-wise calibration functions, requiring distinct temperatures for different nodes. Due to their complexity, it is challenging to be directly optimized via a loss function that targets minimizing the discrepancy between confidence and accuracy. In contrast, \textsc{Simi-Mailbox} requires a simplified, manageable number of parameters, which is merely the number of clusters $N$. The relationship between our group-specific temperatures and model calibration is more straightforward, facilitating the effective adjustment of these temperatures via minimizing calibration error from $\mathcal L_{simi}$. 

Nevertheless, for datasets like CoraFull and heterophilous graphs where the original GNNs’ accuracies are low, incorporating $\mathcal L_{\text{CE}}$ in calibration is crucial, since $\mathcal L_{\text{CE}}$ promotes reducing the entropy of the correctly predicted class while increasing it for the incorrectly predicted class.

\subsection{Discussion on GNNs and Substructure Patterns}\label{B.3}

Extensive literature on GNNs has explored how graph structural patterns influence their performance. One area of focus has been the impact of graph homophily on GNN effectiveness~\citep{mao2024demystifying, ma2021homophily, luan2022revisiting}. For instance, \citet{mao2024demystifying} examines performance of GNNs concerning structural disparities in both homophilous and heterophilous graphs.

While our investigation also leverages structural information to understand GNN prediction behavior, it is important to emphasize the distinct methodologies and objectives that set our \textsc{Simi-Mailbox} apart from the work of \citet{mao2024demystifying}. Whereas \citet{mao2024demystifying} explores the varying performance of GNNs across nodes with different structures within the same graph, our work uses graph structure to calibrate GNN predictions. Our goal is to enhance prediction reliability without compromising original performance. \citet{mao2024demystifying} introduces the concept of structural disparity, which mixes homophilous and heterophilous patterns based on the node homophily ratio—the proportion of neighbors with the same ground-truth label. In contrast, our approach focuses on neighborhood prediction similarity based on predicted labels, offering a different perspective on node similarity. 

Another line of research \citep{tang2020investigating} investigates the performance of GCNs across different node degree distributions. While both our work and \citet{tang2020investigating} explore the relationship between local topology and GNN prediction behavior, the specific goals and methodologies differ significantly.

\citet{tang2020investigating} employs an uncertainty scoring mechanism to generate pseudo-labels, with the goal of providing additional supervision for low-degree nodes to mitigate degree-related bias. In contrast, our work uses pseudo-labels in self-training experiment to demonstrate the utility of calibrated confidence. Additionally, \citet{tang2020investigating} quantifies the impact of node degree on GCN performance using influence functions from statistical fields. This approach differs conceptually from our metric, which measures neighborhood prediction similarity based on the proportion of neighbors sharing the same predicted labels.

Beyond node homophily and degree, we extend our discussion to studies that investigate GNNs through the lens of graph curvatures~\citep{nguyen2023revisiting, topping2021understanding, bober2023rewiring}. For instance, \citet{nguyen2023revisiting} proposes a comprehensive framework using Ollivier-Ricci curvature to explain both over-smoothing and over-squashing in GNNs.

While our work, like \citet{nguyen2023revisiting}, focuses on GNN behaviors within local geometries, their study investigates graph curvature from a differential geometry perspective to define the aforementioned issues. In contrast, we introduce neighborhood prediction similarity as a metric to evaluate the consistency of predictions among a node’s neighbors, rather than measuring the curvature or bending of space. Additionally, while Ollivier-Ricci curvature examines the inherent graph structure alone, our metric integrates model predictions, offering a distinct perspective. Moreover, \citet{nguyen2023revisiting} employs edge rewiring to address over-smoothing and over-squashing. In contrast, \textsc{Simi-Mailbox} utilizes neighborhood similarity to categorize nodes for precise group-specific temperature scaling.

\subsection{Discussion on Grouping-based Calibrations}\label{B.4}
We highlight the key differences between our work and previous categorization-aware calibration approaches in the image domain~\citep{hebert2018multicalibration, perez2022beyond, yang2023beyond}. While our approach shares the objective of addressing miscalibrations from a group-wise perspective, \textsc{Simi-Mailbox} is distinct in its design and focus on the unique challenges posed by graph structures. The methods and analyses in our work are deeply rooted in the properties of graph data, such as neighborhood affinity, which have no direct parallel in the grouping mechanisms discussed in \citep{hebert2018multicalibration, perez2022beyond, yang2023beyond}. We will elaborate on these distinctions one by one.

First, the primary difference between our work and \citet{hebert2018multicalibration} lies in the grouping mechanism. While \citet{hebert2018multicalibration} employs decision tree partitioning, our method uses a more sophisticated categorization based on neighborhood similarity and confidence levels. The approach in \citet{hebert2018multicalibration} is not designed to capture neighborhood similarity, a critical factor in GNN calibration. Moreover, their work does not provide principles for effective categorization, whereas we introduce criteria based on the novel observation of the correlation between neighborhood affinity and miscalibration levels. Additionally, the grouping scheme in \citet{hebert2018multicalibration} is used to quantify grouping loss rather than to develop a new calibration method based on the grouping algorithm.

Furthermore, there are key differentiators that distinguish our work from \citet{perez2022beyond}. While both approaches emphasize subgroup calibration, our method uses a different grouping mechanism. Specifically, we cluster nodes based on neighborhood prediction similarity and confidence levels within the graph. In contrast, \citet{perez2022beyond} proposes a more generalized framework for multicalibration without a specific focus on how ideal subgroups should be formed. Moreover, the core principle of group-wise calibration is to categorize instances based on similar levels of miscalibration. However, \citet{perez2022beyond} does not place particular emphasis on this aspect. Our method, on the other hand, introduces clear criteria—neighborhood similarity and confidence levels—stemming from our novel observation of the strong correlation between nodes with similar neighborhood affinity and confidence.

Lastly, our approach also differs from the method proposed by GC in \citet{yang2023beyond}, particularly in the grouping mechanism. Our method employs a sophisticated categorization based on neighborhood similarity and confidence levels, tailored specifically to the unique properties of GNNs. This careful binning, facilitated by KMeans clustering, leads to effective categorization. In contrast, \citet{yang2023beyond} proposes a learning-based grouping function that does not explicitly account for the inherent characteristics of GNNs. While it performs well in the vision domain, the learning-based grouping function—essentially a single linear layer—is insufficient to capture neighborhood affinity in GNNs.

\input{Table_hetero}
\input{table_hetero2}

\input{Table_st}
To further demonstrate the efficacy of our method over existing grouping-based strategies, we conducted additional experiments to compare \textsc{Simi-Mailbox} and GC~\citep{yang2023beyond}. Since GC requires the holdout set to train the calibration function, we adopted GC \textit{with} the holdout set (specified as \textbf{GC w/ HO}). 
We adopted two configurations, GC combined with TS or ETS, following the same combination as in the original paper.
On the comparison between our method and GC~\citep{yang2023beyond}, we randomly sample $10\%$ of test data and allocate them as the holdout set for GC, adjusting the original evaluation protocol in \citet{yang2023beyond}. Accordingly, we re-evaluate our method in the remaining $90\%$ of the test data. 
Note that unlike GC, \textbf{our method} is trained solely on the validation set, \textbf{without access to the holdout data}.

As a result, table~\ref{Table_pce2} further confirms the superiority of our method across 15 out of 16 settings, validating the efficacy of our proposed grouping strategy against GC, \textit{even} without the access to the holdout set. 
Specifically, our method is distinguished as the only approach to achieve error rates within the $2\%$ range for Cora and Citeseer datasets and $0.7\%$ for Pubmed dataset.
This indicate that the learning-based grouping function does not fully encompass the intrinsic properties of GNNs.

\input{Table_recent}
\subsection{Comparison with Recent GNN Calibration Methods}\label{B.5}
We further evaluated our method with recent GNN calibration approaches, SimCalib~\cite{simcalib}, DCGC~\cite{dcgc}, and GETS~\cite{gets}, to verify its effectiveness. Since the experimental settings in DCGC and GETS differ significantly from ours, we reproduced these methods under our setup to ensure a fair comparison. Additionally, we used GCN and GraphSAGE as backbone GNNs for DCGC, as it is designed for GNNs capable of processing edge weights as inputs. For SimCalib, we reported results from the original paper, as their experimental configurations align with ours.

As shown in Table~\ref{Table_recent}, our method demonstrates state-of-the-art performance in 13 of 16 settings, with performance improvements that reach 0.96\% compared to GETS, the strongest baseline. A similar trend is evident in Table 2, where \textsc{Simi-Mailbox} consistently outperforms DCGC in all cases, achieving a gap of up to 3.49\%. These results confirm the efficacy of our approach in comparison to recent methods.

\subsection{Comparison with Bayesian Calibration Approach}\label{B.6}
We provide additional experiments to compare \textsc{Simi-Mailbox} and GPN~\citep{stadler2021graph}, a representative Bayesian GNN uncertainty quantification approach, on the same benchmark datasets used in \citet{stadler2021graph}. On comparison with GPN, we adhere to the evaluation settings in \citet{stadler2021graph}.

As shown in Table~\ref{table_gpn}, our method demonstrates a significant enhancement in calibrating GPN, with the most substantial improvements observed within the CS, where it achieved a remarkable reduction in GPN's calibration error by 13.69\%. Interestingly, in Physics, our method successfully lowers the error rate to below 1\%. These results collectively demonstrate the effectiveness of our method across various graphs and architectures.

\subsection{Hyperparameter Sensitivity}\label{B.7}
We present the comprehensive sensitivity analysis on robustness of \textsc{Simi-Mailbox} with respect to its hyperparameters and the choice of scaling functions.
We compare the ECE results of the strongest baseline GATS (specified as dark brown) and ours across varying values of a scaling factor $\lambda$ (specified as green) and the number of bins $N$ (specified as pink) within the range of [5, 10, 15, 20, 25, 30]. The results on both GCN and GAT across whole benchmark datasets is depicted in Figure~\ref{fig: sensitivity_all}.

As demonstrated in the figure, \textsc{Simi-Mailbox} consistently outperforms the baseline across all hyperparameter configurations throughout diverse settings, with the number of bins $N$ demonstrating particularly stable performance trends. This robustness is attributed to our method's design, which accounts for the correlation between neighborhood similarity and the degree of miscalibration.

Meanwhile, the choice of min-max scaling is rooted in the intuition of potential disparity in distributions of neighborhood similarity and confidence. For instance, while neighborhood similarity can be evenly distributed between 0 and 1, confidence in high-accuracy datasets might be concentrated at higher values. In this situation, min-max scaling is an effective technique for normalizing data, especially when the values are concentrated in a specific range.

However, our method can accomplish prominent performance when equipped with different scaling functions. To verify this, we conducted additional experiment on our method with standard scaling (standard normalization) for constructing a feature vector, illustrated in Table~\ref{Table_scaling}. According to the table, \textsc{Simi-Mailbox} equipped with standard scaling consistently outperforms the strongest baseline in 15 of the 16 settings. Moreover, the performance gap between our method with standard scaling and the original \textsc{Simi-Mailbox} is marginal, suggesting that \textsc{Simi-Mailbox} is resilient to different choices of scaling method as well.

\subsection{\textsc{Simi-Mailbox} on Heterophilous Graphs}\label{B.8}
We conducted additional evaluations to further demonstrate the efficacy of \textsc{Simi-Mailbox} on heterophilous graphs, comparing it with uncalibrated GNNs (UnCal.), temperature scaling (TS)~\citep{guo2017calibration}, and GATS. The benchmark datasets for these experiments included Chameleon, Squirrel, Actor, Texas, Wisconsin, and Cornell from \citet{pei2020geom}, as well as Roman-empire, Amazon-ratings, Minesweeper, Tolokers, and Questions from \citet{luan2022revisiting}. For the datasets from \citet{pei2020geom}, we used 10 different train/validation/test splits provided in the official PyTorch Geometric Library~\citep{fey2019fast}. For the datasets from \citet{luan2022revisiting}, we adopted 10 splits from their official GitHub repository. For each split, we conducted 5 random initializations, resulting in 50 runs in total.

\input{table_sparse}
\input{Table_ood}

\input{table_mlp}

As shown in Table~\ref{Table_hetero}, our method outperforms the baselines in 14 out of 16 settings. Notably, on the Texas dataset, \textsc{Simi-Mailbox} reduces ECE by 2.65\% and 2.98\% for GCN and GAT, respectively, compared to uncalibrated results. In contrast, TS and GATS showed limited effectiveness in reducing calibration error on the Texas dataset; in fact, GATS increased ECE beyond the uncalibrated results.

Consistent with our prior findings, \textsc{Simi-Mailbox} significantly enhances calibration across various benchmarks and models, achieving a notable ECE reduction of 12.67\% for GAT on the Roman-empire dataset. Demonstrating superiority over baselines in 8 out of 10 cases, our method validates its efficacy and adaptability to low homophily levels. Interestingly, our method also improves calibration in GNNs that are already considered well-calibrated, as seen with the Questions dataset, where we achieved an error rate below 1\%. This underscores the versatility of our method, regardless of the original GNNs' miscalibration degree.

\subsection{\textsc{Simi-Mailbox} in Self-training}\label{B.9}
The efficacy of \textsc{Simi-Mailbox} was further evaluated in self-training scenarios, initially conducted in in CaGCN. For the fair comparisons, we adhered to the same datasets, split ratios, and evaluation protocols in \citet{wang2021confident}, also ensuring the uniform five random seeds across all experimental setups. 

As presented in Table~\ref{Table_st_gcn} and \ref{Table_st_gat}, \textsc{Simi-Mailbox} demonstrates superior performance over both uncalibrated GNNs and CaGCN across 17 out of 18 settings in total. This is especially evident in GAT on the Citeseer dataset, where our method achieves a performance increase of $4.18\%$ compared to uncalibrated GAT when label rate (L/C)=40. These results underscore the efficacy of our method in generating refined pseudo-labels through its sophisticated calibration process.

\subsection{\textsc{Simi-Mailbox} in Sparse Label Scenario}\label{B.10}
We performed additional experiments in sparse label scenario, where 20 nodes per class were randomly selected for training, as adopted in \citet{chen2024exploring, wang2021confident}. Following the common protocols in \citep{chen2024exploring, wang2021confident}, we utilized 500 random nodes for validation and 1000 nodes for testing, averaging results over 10 random splits. We benchmarked our method alongside uncalibrated GNNs (UnCal.) and the GATS, the most competitive baseline. 

According to the results in Table~\ref{table_sparse_gcn} and \ref{table_sparse_gat}, our method consistently achieves notable enhancements, especially on Cora and Citeseer datasets, with error reductions of 24.48\% and 19.84\% on average. Notably, our method shows stable calibration across diverse splits against uncalibrated GNNs, evidenced by the marked decrease in standard deviation. These results affirm the robustness of our \textsc{Simi-Mailbox} to label sparsity, highlighting its efficacy. 

\subsection{\textsc{Simi-Mailbox} in Out-of-distribution Detection}\label{B.11}
In this subsection, we explore the potential of \textsc{Simi-Mailbox} for out-of-distribution (OOD) detection, comparing its performance against specialized OOD methods, GNNSafe~\citep{wu2023energy} and GPN~\citep{stadler2021graph}. For this evaluation, we selected the Cora, Photo, and Physics datasets, using GCN as the backbone architecture, consistent with the setup in GNNSafe. In this experiment, we applied two perturbation strategies to generate OOD data: (1) structure manipulation, where a stochastic block model was used to randomly generate graph structures, and (2) feature manipulation, where random feature interpolation was employed to generate node features. 

We adhered to the same experimental configurations and evaluation protocols as GNNSafe, ensuring a fair comparison. The performance of each method was assessed using the Area Under the Receiver Operating Characteristic (AUROC) metric, with the results summarized in Table~\ref{Table_ood}.

Although OOD detection is not the primary focus of \textsc{Simi-Mailbox}, our method demonstrates commendable performance. Notably, \textsc{Simi-Mailbox} outperform GPN on the Photo and Physics datasets across all perturbation strategies. Furthermore, it is worth highlighting that \textsc{Simi-Mailbox} achieves performance levels comparable to the recent state-of-the-art approach, GNNSafe, particularly on the Photo dataset under structural perturbation, where the gap in AUROC is 0.19\%.

\subsection{\textsc{Simi-Mailbox} with Broader GNN Architectures}\label{B.12}

We extended our evaluations to include MLP, SAGE~\citep{hamilton2017inductive}, and GIN~\citep{xu2018powerful} to verify the versatility of \textsc{Simi-Mailbox}. We compared our method against uncalibrated GNNs (UnCal.) and GATS, using the same experimental configurations as in the main experiments on GCN and GAT, resulting in a total of 75 runs.

As shown in Tables~\ref{table_mlp}, \ref{table_sage}, and \ref{table_gin}, our method outperforms the baselines in 22 out of 24 settings. Notably, on the Citeseer dataset, \textsc{Simi-Mailbox} achieves significant ECE reductions of 8.75\%, 4.25\%, and 6.99\% for MLP, SAGE, and GIN, respectively. Additionally, in calibrating MLP, our method surpasses all baselines by an average margin of 3.37\%, highlighting the model-agnostic nature of \textsc{Simi-Mailbox}.

\subsection{Results from Different Evaluation Metrics}\label{B.13}
Here, we provide supplementary results evaluated with different calibration metrics, including class-wise ECE~\citep{kull2019beyond}, kernel density estimation-based ECE (KDE-ECE)~\citep{zhang2020mix}, negative log-likelihood (NLL), and Brier Score~\citep{brier1950verification}.
\begin{itemize}
    \item \textbf{KDE-ECE} utilizes a smoothing kernel function denoted as $K_h$ with a fixed bandwidth $h$ to estimate the accuracies $\hat\pi$ and marginal probabilities $\hat f$.
    The calibration error is then quantified through the integration of the absolute difference between the estimated accuracy and predicted confidence $\hat p$, formulated as follows:
    \begin{equation}
    \begin{gathered}
        \text{KDE-ECE}=\int |\hat\pi(\hat p)-\hat p|\hat f(\hat p)d\hat p,\\
        \hat\pi(\hat p)=\frac{\sum_{i\in\mathcal{V}}^N\mathbf{1}[y_i=\hat y_i]\prod_{k=1}^CK_h(\hat p-p_{i,k})}{\sum_{i\in\mathcal{V}}^N\prod_{k=1}^CK_h(\hat p-p_{i,k})},\\
        \hat f(\hat p)=\frac{h^{-1}}{N}\sum_{i\in\mathcal{V}}^N\prod_{k=1}^CK_h(\hat p-p_{i,k})
    \end{gathered}
    \end{equation}
    Following the precedent~\citep{hsu2022makes}, we implement the Triweight Kernel function $K_h(v)=(1/h)\frac{35}{32}(1-(v/h)^2)^3$~\citep{de1999use}, where the bandwidth is calculated as $h=1.06\sigma N^{-1/5}$~\citep{scott2015multivariate}, with $\sigma$ representing the standard deviation of the confidence here.
    
    \item \textbf{Class-wise ECE} extends the general concept of ECE to class-wise perspective. It measures the discrepancy between the ground-truth frequency and the average predicted probability within each confidence bin for each class $k$, defined as:
    \begin{equation}
    \begin{gathered}
    \text{ECE}(k)=\sum_{m=1}^M \frac{|B_m|}{N}|\text{freq}(B_{m,k})-\text{conf}(B_{m,k})|,\\
    \text{freq}(B_{m,k})=\frac{1}{|B_{m,k}|}\sum_{i\in B_{m,k}}\mathbf{1}[y_i=k]
    \end{gathered}
    \end{equation}
    The overall class-wise ECE is obtained by averaging $\text{ECE}(k)$ across all classes, i.e. $\text{Class-wise ECE}=\frac{1}{C}\sum_{k=1}^C\text{ECE}(k)$.
    
    \item \textbf{NLL} is frequently used for evaluating calibration to assess the overall miscalibration, computed by the average of logarithms of the predicted probability for each correct class, formulated as follows:
    \begin{equation}
        \text{NLL}=\frac{1}{N}\sum_{i\in\mathcal{V}}^N - y_{i}\log p_{i,y_i}
    \end{equation}
    
    \item \textbf{Brier Score} is widely employed metric to quantify the model calibration. It measures the accuracy of model prediction by comparing the predicted probabilities $\mathbf{p}_i$ with the ground-truth occurrences $\mathbf{o}_i$:
    \begin{equation}
        \text{Brier Score}=\frac{1}{N}\sum_{i\in\mathcal{V}}^N\sum_{k=1}^C(p_{i,k}-o_{i,k})
    \end{equation}

    Here, $\mathbf{o}_i$ represents a one-hot vector encoding the ground-truth class label.
\end{itemize}
We report the calibration results assessed by KDE-ECE, class-wise ECE, NLL, and Brier Score in Table~\ref{Table_kde},~\ref{Table_class_ece},~\ref{Table_nll}, and Table~\ref{Table_brier}, respectively.
The results demonstrate that our \textsc{Simi-Mailbox} generally outperforms state-of-the-art calibration methods on majority of the metrics, including all metrics, particularly when assessed via KDE-ECE, accomplishing state-of-the-art calibration performance across all settings.

\section{Additional Qualitative Results}\label{C}
In this section, we provide additional qualitative comparisons on the Citeseer (co-citation), Photo (Amazon), Physics (co-authored), and CoraFull (co-citation) datasets, as illustrated in Figures~\ref{Fig: citeseer_result_plot}, \ref{Fig: computers_result_plot}, \ref{Fig: physics_result_plot}, and \ref{Fig: corafull_result_plot}. As shown, our method generally achieves better reduction on the discrepancy between per-confidence accuracy and average confidence across varying neighborhood similarity sub-intervals, outperforming the baselines.

\section{Additional Observation Results}\label{D}
We present additional in-depth analysis on Citeseer, Photo, Physics, and CoraFull, as illustrated in Figure~\ref{Fig: citeseer_obs_plot}, \ref{Fig: computers_obs_plot}, \ref{Fig: physics_obs_plot}, \ref{Fig: corafull_obs_plot}. 
Taking everything into account, the results show that our findings are not limited to the single case discussed in the analysis section, challenging the uniform assumption of previous GNN calibration studies on neighborhood similarity.


\newpage
\input{Table_kde}
\input{Table_class_ece}
\input{Table_nll}
\input{Table_brier}

\newpage
\input{qualitative_analysis_appendix}
\input{obs_plot_appendix}


%% file: Table_statistics.tex
{\renewcommand{\arraystretch}{.92}
\begin{table}[h!]
    \captionsetup{width=.75\textwidth}
    \caption{Statistics of benchmark datasets.}
    \vskip 0.05in
    \centering
    \resizebox{.45\textwidth}{!}{
    \begin{tabular}{lccc}
    \hline
    \small
     \textbf{Dataset} & \textbf{\#Nodes} & \textbf{\#Edges} & \textbf{\#Classes}  \\
     \hline
     {Cora} & 2,708 & 10,556 & 7 \\
     {Citeseer} & 3,327 & 9,104 & 6 \\
     {Pubmed} & 19,717 & 88,648 & 3 \\
     {CoraFull} & 19,793 & 126,842 & 70 \\
     {Computers} & 13,752 & 491,722 & 10 \\
     {Photo} & 7,650 & 238,162 & 8 \\
     {CS} & 18,333 & 163,788 & 15 \\
     {Physics} & 34,493 & 495,924 & 5 \\
     {Arxiv} & 169,343 &  1,166,243 & 40 \\
     {Reddit} & 232,965 & 23,213,838 & 41 \\
     \hline
    \end{tabular}
    }
    \label{Table_statistics}
\vskip -10pt
\end{table}
}

%% file: Table_pce2.tex
{\renewcommand{\arraystretch}{1.02}
\begin{table*}[t]
\caption{ECE results (reported in percentage) for our proposed calibration method and GC \textit{with} the holdout set. A lower ECE indicates better calibration performance. Note that \textbf{Simi-Mailbox does not have an access} to the holdout data.}
\centering
\resizebox{\textwidth}{!}{
\Large
{
\begin{tabular}
{ c | c | c | c | c | c | c | c | c | c }
\hline\hline

\multicolumn{2}{ c |}{\textbf{Datasets}} & {Cora} & {Citeseer} & {Pubmed} & {Computers} & {Photo} & {CS} & {Physics} &{CoraFull} \\

\hline

\multirow{3}{*}{GCN} 
& GC+TS w/ HO &
{{3.59 \fontsize{11}{12}\selectfont$\pm$ 1.01}} &
{{4.16 \fontsize{11}{12}\selectfont$\pm$ 1.09}} &
{{1.27 \fontsize{11}{12}\selectfont$\pm$ 0.31}} &
{{3.17 \fontsize{11}{12}\selectfont$\pm$ 0.81}} &
{{2.09 \fontsize{11}{12}\selectfont$\pm$ 0.84}} &
{{0.99 \fontsize{11}{12}\selectfont$\pm$ 0.20}} &
{{0.49 \fontsize{11}{12}\selectfont$\pm$ 0.18}} &
{{5.57 \fontsize{11}{12}\selectfont$\pm$ 0.52}} \\

& GC+ETS w/ HO &
{{3.29 \fontsize{11}{12}\selectfont$\pm$ 0.94}} &
{{3.69 \fontsize{11}{12}\selectfont$\pm$ 1.02}} &
{{1.15 \fontsize{11}{12}\selectfont$\pm$ 0.40}} &
{{1.45 \fontsize{11}{12}\selectfont$\pm$ 0.45}} &
{{1.24 \fontsize{11}{12}\selectfont$\pm$ 0.45}} &
{{0.90 \fontsize{11}{12}\selectfont$\pm$ 0.24}} &
{{0.48 \fontsize{11}{12}\selectfont$\pm$ 0.20}} &
{{4.05 \fontsize{11}{12}\selectfont$\pm$ 0.47}} \\

& {\textbf{Ours}} &
\textbf{{2.06 \fontsize{11}{12}\selectfont$\pm$ 0.44}} &
\textbf{{2.76 \fontsize{11}{12}\selectfont$\pm$ 0.56}} &
\textbf{{0.77 \fontsize{11}{12}\selectfont$\pm$ 0.15}} &
\textbf{{1.06 \fontsize{11}{12}\selectfont$\pm$ 0.26}} &
\textbf{{1.04 \fontsize{11}{12}\selectfont$\pm$ 0.35}} &
\textbf{{0.60 \fontsize{11}{12}\selectfont$\pm$ 0.19}} &
\textbf{{0.29 \fontsize{11}{12}\selectfont$\pm$ 0.11}} &
\textbf{{3.47 \fontsize{11}{12}\selectfont$\pm$ 1.32}} \\
\hline

\multirow{3}{*}{GAT} 
& GC+TS w/ HO &
{{3.13 \fontsize{11}{12}\selectfont$\pm$ 0.97}} &
{{3.85 \fontsize{11}{12}\selectfont$\pm$ 1.16}} &
{{1.02 \fontsize{11}{12}\selectfont$\pm$ 0.41}} &
{{1.53 \fontsize{11}{12}\selectfont$\pm$ 0.48}} &
{{1.63 \fontsize{11}{12}\selectfont$\pm$ 0.79}} &
{{0.91 \fontsize{11}{12}\selectfont$\pm$ 0.25}} &
\textbf{{0.47 \fontsize{11}{12}\selectfont$\pm$ 0.17}} &
{{4.32 \fontsize{11}{12}\selectfont$\pm$ 0.50}} \\

& {GC+ETS w/ HO} &
{{3.15 \fontsize{11}{12}\selectfont$\pm$ 0.98}} &
{{3.60 \fontsize{11}{12}\selectfont$\pm$ 1.05}} &
{{1.07 \fontsize{11}{12}\selectfont$\pm$ 0.45}} &
{{1.26 \fontsize{11}{12}\selectfont$\pm$ 0.37}} &
{{1.34 \fontsize{11}{12}\selectfont$\pm$ 0.52}} &
{{0.84 \fontsize{11}{12}\selectfont$\pm$ 0.27}} &
{{0.51 \fontsize{11}{12}\selectfont$\pm$ 0.23}} &
{{3.55 \fontsize{11}{12}\selectfont$\pm$ 0.48}} \\

& {\textbf{Ours}} & 
\textbf{{2.15 \fontsize{11}{12}\selectfont$\pm$ 0.44}} &
\textbf{{2.97 \fontsize{11}{12}\selectfont$\pm$ 0.58}} &
\textbf{0.73 \fontsize{11}{12}\selectfont$\pm$ 0.17} &
\textbf{{0.98 \fontsize{11}{12}\selectfont$\pm$ 0.38}} &
\textbf{{1.00 \fontsize{11}{12}\selectfont$\pm$ 0.52}} &
\textbf{{0.75 \fontsize{11}{12}\selectfont$\pm$ 0.43}} &
{0.49 \fontsize{11}{12}\selectfont$\pm$ 0.21} &
\textbf{{2.66 \fontsize{11}{12}\selectfont$\pm$ 1.01}}\\
\hline\hline
\end{tabular}}
}
\label{Table_pce2}
\vskip -5pt
\end{table*}
}

%% file: Table_gpn.tex
{\renewcommand{\arraystretch}{1.2}
\begin{table*}[t]
\centering
\caption{\blue{ECE results (reported in percentage) for our proposed calibration method with GPN and uncalibrated GPN, averaged over 10 random splits.}}
\vskip -5pt
\resizebox{\textwidth}{!}{
\huge
\begin{tabular}{ c | c | c | c | c | c | c | c | c }
\hline
\hline
\textbf{Datasets} & CoraML & Cora & Citeseer & Pubmed & Computers & Photo & CS & Physics \\ \hline\hline
GPN & 9.93 \fontsize{18}{19}\selectfont$\pm$ 1.28 
& 5.93 \fontsize{18}{19}\selectfont$\pm$ 2.38 
& 5.76 \fontsize{18}{19}\selectfont$\pm$ 2.74 
& 4.82 \fontsize{18}{19}\selectfont$\pm$ 1.50 
& 9.94 \fontsize{18}{19}\selectfont$\pm$ 1.86 
& 12.16 \fontsize{18}{19}\selectfont$\pm$ 0.81 
& 18.48 \fontsize{18}{19}\selectfont$\pm$ 0.60 
& 10.69 \fontsize{18}{19}\selectfont$\pm$ 0.68 \\
\textbf{Ours} 
& \textbf{3.91 \fontsize{18}{19}\selectfont$\pm$ 0.57} 
& \textbf{3.76 \fontsize{18}{19}\selectfont$\pm$ 0.50} 
& \textbf{4.35 \fontsize{18}{19}\selectfont$\pm$ 0.49} 
& \textbf{1.07 \fontsize{18}{19}\selectfont$\pm$ 0.28} 
& \textbf{4.19 \fontsize{18}{19}\selectfont$\pm$ 0.46} 
& \textbf{2.40 \fontsize{18}{19}\selectfont$\pm$ 0.39} 
& \textbf{4.79 \fontsize{18}{19}\selectfont$\pm$ 1.14} 
& \textbf{0.69 \fontsize{18}{19}\selectfont$\pm$ 0.09} \\ \hline
\end{tabular}}
\label{table_gpn}
\vskip -10pt
\end{table*}
}

%% file: sensitivity_all.tex
\begin{figure*}[t]
    \centering
\includegraphics[width=.96\linewidth]{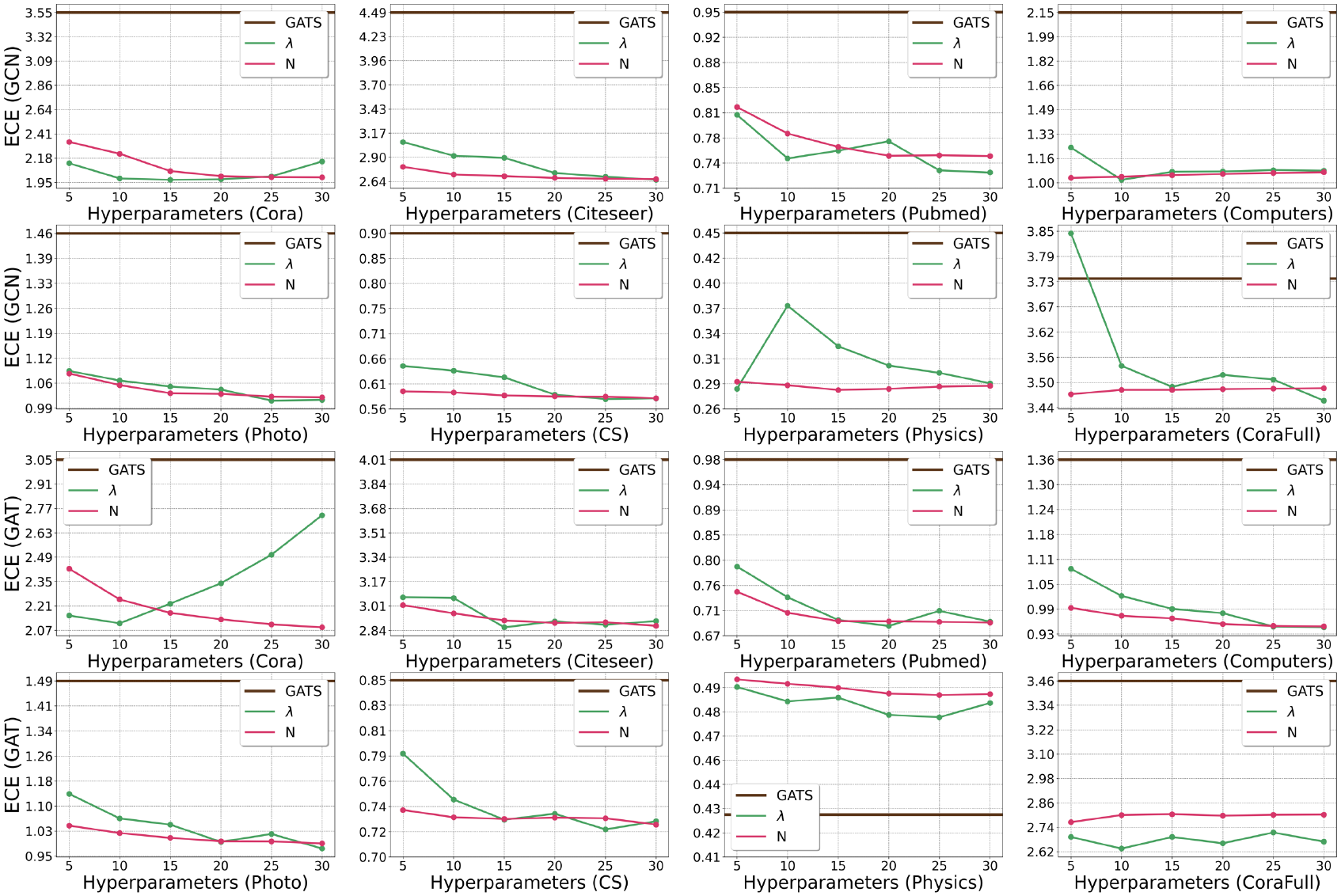} 
\caption{Hyperparameter sensitivity of scaling factor $\lambda$ and the number of bins $N$ across all benchmark datasets and GNN architectures.}
\label{fig: sensitivity_all}
\vskip -8pt
\end{figure*}

%% file: Table_scaling.tex
{\renewcommand{\arraystretch}{1.1}
\begin{table*}[t]
\caption{ECE results (reported in percentage) for our original calibration method with min-max scaling (specified as \textbf{Ours}) and standard scaling (specified as \textbf{Ours w/ standard scaling}), compared to GATS. A lower ECE indicates better calibration performance. The best and second best performances are represented by bold and underline texts, respectively.}
\centering
\resizebox{\textwidth}{!}{
\Huge
{
\begin{tabular}
{ c | c | c | c | c | c | c | c | c | c }
\hline\hline

\multicolumn{2}{ c |}{\textbf{Datasets}} & {Cora} & {Citeseer} & {Pubmed} & {Computers} & {Photo} & {CS} & {Physics} &{CoraFull} \\

\hline

\multirow{3}{*}{GCN} 
& GATS & 
{3.55 \fontsize{18}{19}\selectfont$\pm$ 1.28} &
{4.49 \fontsize{18}{19}\selectfont$\pm$ 1.53} &
{0.95 \fontsize{18}{19}\selectfont$\pm$ 0.32} & 
{2.15 \fontsize{18}{19}\selectfont$\pm$ 0.52} & 
{1.46 \fontsize{18}{19}\selectfont$\pm$ 0.51} & 
{0.90 \fontsize{18}{19}\selectfont$\pm$ 0.29} & 
{0.45 \fontsize{18}{19}\selectfont$\pm$ 0.15} &
{3.74 \fontsize{18}{19}\selectfont$\pm$ 0.63} \\

& {\textbf{Ours}} &
\textbf{1.97 \fontsize{18}{19}\selectfont$\pm$ 0.44} &
\textbf{2.66 \fontsize{18}{19}\selectfont$\pm$ 0.53} &
\textbf{0.75 \fontsize{18}{19}\selectfont$\pm$ 0.15} &
\textbf{1.02 \fontsize{18}{19}\selectfont$\pm$ 0.26} &
\textbf{1.01 \fontsize{18}{19}\selectfont$\pm$ 0.36} &
\textbf{0.58 \fontsize{18}{19}\selectfont$\pm$ 0.19} &
\textbf{0.28 \fontsize{18}{19}\selectfont$\pm$ 0.11} &
\textbf{3.46 \fontsize{18}{19}\selectfont$\pm$ 1.31} \\

& {\textbf{Ours w/ standard scaling}} &
\underline{2.02 \fontsize{18}{19}\selectfont$\pm$ 0.54} &
\underline{2.67 \fontsize{18}{19}\selectfont$\pm$ 0.52} &
\textbf{0.75 \fontsize{18}{19}\selectfont$\pm$ 0.14} &
\underline{1.06 \fontsize{18}{19}\selectfont$\pm$ 0.24} &
\textbf{1.01 \fontsize{18}{19}\selectfont$\pm$ 0.34} &
\underline{0.63 \fontsize{18}{19}\selectfont$\pm$ 0.18} &
\underline{0.30 \fontsize{18}{19}\selectfont$\pm$ 0.10} &
\underline{3.53 \fontsize{18}{19}\selectfont$\pm$ 1.30} \\
\hline

\multirow{3}{*}{GAT} 
& GATS & 
{3.05 \fontsize{18}{19}\selectfont$\pm$ 0.78} & 
{4.01 \fontsize{18}{19}\selectfont$\pm$ 1.42} & 
{0.98 \fontsize{18}{19}\selectfont$\pm$ 0.36} & 
{1.36 \fontsize{18}{19}\selectfont$\pm$ 0.34} & 
{1.49 \fontsize{18}{19}\selectfont$\pm$ 0.65} & 
{0.85 \fontsize{18}{19}\selectfont$\pm$ 0.23} & 
\textbf{0.43 \fontsize{18}{19}\selectfont$\pm$ 0.16} &
{3.46 \fontsize{18}{19}\selectfont$\pm$ 0.46} \\

& {\textbf{Ours}} & 
\textbf{2.08 \fontsize{18}{19}\selectfont$\pm$ 0.45} &
\textbf{2.86 \fontsize{18}{19}\selectfont$\pm$ 0.56} &
\textbf{0.69 \fontsize{18}{19}\selectfont$\pm$ 0.16} &
\underline{0.95 \fontsize{18}{19}\selectfont$\pm$ 0.37} &
\textbf{0.97 \fontsize{18}{19}\selectfont$\pm$ 0.53} &
\textbf{0.72 \fontsize{18}{19}\selectfont$\pm$ 0.43} &
\underline{0.48 \fontsize{18}{19}\selectfont$\pm$ 0.22} &
\textbf{2.64 \fontsize{18}{19}\selectfont$\pm$ 1.02} \\

& {\textbf{Ours w/ standard scaling}} &
\underline{2.23 \fontsize{18}{19}\selectfont$\pm$ 0.41} &
\underline{2.93 \fontsize{18}{19}\selectfont$\pm$ 0.58} &
\textbf{0.69 \fontsize{18}{19}\selectfont$\pm$ 0.19} &
\textbf{0.94 \fontsize{18}{19}\selectfont$\pm$ 0.37} &
\underline{0.98 \fontsize{18}{19}\selectfont$\pm$ 0.52} &
\underline{0.73 \fontsize{18}{19}\selectfont$\pm$ 0.43} &
{0.49 \fontsize{18}{19}\selectfont$\pm$ 0.21} &
\underline{2.82 \fontsize{18}{19}\selectfont$\pm$ 1.06} \\
\hline\hline
\end{tabular}}
}
\label{Table_scaling}
\vskip -10pt
\end{table*}
}

%% file: Table_hetero.tex
{\renewcommand{\arraystretch}{.92}
\begin{table*}[t]
\caption{ECE results (reported in percentage) on heterophilous datasets for our proposed calibration method and baselines, averaged over 50 runs. A lower ECE indicates better calibration performance.}
\centering
\resizebox{.89\textwidth}{!}{
\small
{
\begin{tabular}
{ c | c | c | c | c | c | c | c }
\hline\hline

\multicolumn{2}{ c |}{\textbf{Datasets}} & {Chameleon} & {Squirrel} & {Actor} & {Texas} & {Wisconsin} & {Cornell}  \\

\hline

\multirow{4}{*}{GCN} & UnCal. &
{9.39 \fontsize{8}{9}\selectfont$\pm$ 1.90} &
{7.23 \fontsize{8}{9}\selectfont$\pm$ 1.46} &
\textbf{2.69 \fontsize{8}{9}\selectfont$\pm$ 0.88} &
{18.15 \fontsize{8}{9}\selectfont$\pm$ 4.13} &
{17.76 \fontsize{8}{9}\selectfont$\pm$ 6.97} &
{19.17 \fontsize{8}{9}\selectfont$\pm$ 4.94} \\

& TS & 
{9.42 \fontsize{8}{9}\selectfont$\pm$ 1.94} &
{7.18 \fontsize{8}{9}\selectfont$\pm$ 1.37} &
{2.82 \fontsize{8}{9}\selectfont$\pm$ 0.95} &
{18.12 \fontsize{8}{9}\selectfont$\pm$ 4.95} &
{15.41 \fontsize{8}{9}\selectfont$\pm$ 5.08} &
{19.93 \fontsize{8}{9}\selectfont$\pm$ 5.31} \\

& GATS & 
{8.25 \fontsize{8}{9}\selectfont$\pm$ 2.15} &
{6.85 \fontsize{8}{9}\selectfont$\pm$ 1.09} &
{2.82 \fontsize{8}{9}\selectfont$\pm$ 1.01} &
{18.73 \fontsize{8}{9}\selectfont$\pm$ 4.34} &
{15.76 \fontsize{8}{9}\selectfont$\pm$ 5.33} &
{21.60 \fontsize{8}{9}\selectfont$\pm$ 5.28} \\

& \textbf{Ours} & 
\textbf{7.50 \fontsize{8}{9}\selectfont$\pm$ 1.40} &
\textbf{5.40 \fontsize{8}{9}\selectfont$\pm$ 1.04} &
{2.73 \fontsize{8}{9}\selectfont$\pm$ 0.89} &
\textbf{15.50 \fontsize{8}{9}\selectfont$\pm$ 4.44} &
\textbf{15.19 \fontsize{8}{9}\selectfont$\pm$ 3.58} &
\textbf{18.66 \fontsize{8}{9}\selectfont$\pm$ 5.24} \\
\hline

\multirow{4}{*}{GAT} & UnCal. &
{7.27 \fontsize{8}{9}\selectfont$\pm$ 1.43} &
{6.42 \fontsize{8}{9}\selectfont$\pm$ 1.30} &
{3.49 \fontsize{8}{9}\selectfont$\pm$ 1.11} &
{18.51 \fontsize{8}{9}\selectfont$\pm$ 4.47} &
{16.12 \fontsize{8}{9}\selectfont$\pm$ 4.36} &
\textbf{14.89 \fontsize{8}{9}\selectfont$\pm$ 6.35} \\

& TS &
{7.23 \fontsize{8}{9}\selectfont$\pm$ 1.44} &
{6.43 \fontsize{8}{9}\selectfont$\pm$ 1.31} &
{3.29 \fontsize{8}{9}\selectfont$\pm$ 1.15} &
{18.62 \fontsize{8}{9}\selectfont$\pm$ 3.99} &
{15.64 \fontsize{8}{9}\selectfont$\pm$ 3.25} &
{16.00 \fontsize{8}{9}\selectfont$\pm$ 6.72} \\

& GATS &
{7.79 \fontsize{8}{9}\selectfont$\pm$ 1.95} &
{6.66 \fontsize{8}{9}\selectfont$\pm$ 1.63} &
{3.41 \fontsize{8}{9}\selectfont$\pm$ 1.14} &
{18.91 \fontsize{8}{9}\selectfont$\pm$ 4.49} &
{15.16 \fontsize{8}{9}\selectfont$\pm$ 2.86} &
{18.08 \fontsize{8}{9}\selectfont$\pm$ 6.74} \\

& \textbf{Ours} &
\textbf{6.75 \fontsize{8}{9}\selectfont$\pm$ 1.73} &
\textbf{5.45 \fontsize{8}{9}\selectfont$\pm$ 1.30} &
\textbf{2.64 \fontsize{8}{9}\selectfont$\pm$ 0.99} &
\textbf{15.53 \fontsize{8}{9}\selectfont$\pm$ 3.48} &
\textbf{14.90 \fontsize{8}{9}\selectfont$\pm$ 3.29} &
{18.54 \fontsize{8}{9}\selectfont$\pm$ 6.35} \\
\hline\hline
\end{tabular}}
}
\label{Table_hetero}
\vskip -5pt
\end{table*}
}

%% file: table_hetero2.tex
{\renewcommand{\arraystretch}{.92}
\begin{table*}[t]
\caption{\blue{ECE results (reported in percentage) on broader range of heterophilous datasets for our proposed method and baselines, averaged over 50 runs. A lower ECE indicates better calibration performance.}}
\centering
\resizebox{.85\textwidth}{!}{
\small
{
\begin{tabular}{ c | c | c | c | c | c | c }
\hline\hline
\multicolumn{2}{ c |}{\textbf{Datasets}} & Roman-empire & Amazon-ratings & Minesweeper & Tolokers & Questions \\
\hline
\multirow{4}{*}{GCN} & UnCal. & 4.11 \fontsize{8}{9}\selectfont$\pm$ 0.38 & 3.04 \fontsize{8}{9}\selectfont$\pm$ 0.68 & 1.71 \fontsize{8}{9}\selectfont$\pm$ 0.88 & 3.90 \fontsize{8}{9}\selectfont$\pm$ 0.94 & 1.06 \fontsize{8}{9}\selectfont$\pm$ 0.66 \\

& TS & 4.07 \fontsize{8}{9}\selectfont$\pm$ 0.40 & 2.97 \fontsize{8}{9}\selectfont$\pm$ 0.64 & 1.65 \fontsize{8}{9}\selectfont$\pm$ 0.79 & 3.63 \fontsize{8}{9}\selectfont$\pm$ 0.75 & 1.17 \fontsize{8}{9}\selectfont$\pm$ 0.74 \\

& GATS & \textbf{2.48 \fontsize{8}{9}\selectfont$\pm$ 0.35} & \textbf{2.24 \fontsize{8}{9}\selectfont$\pm$ 0.56} & 1.85 \fontsize{8}{9}\selectfont$\pm$ 0.90 & 3.33 \fontsize{8}{9}\selectfont$\pm$ 0.59 & 1.09 \fontsize{8}{9}\selectfont$\pm$ 0.52 \\

& \textbf{Ours} & 
4.72 \fontsize{8}{9}\selectfont$\pm$ 0.54 & 
3.01 \fontsize{8}{9}\selectfont$\pm$ 0.72 & 
\textbf{1.36 \fontsize{8}{9}\selectfont$\pm$ 0.41} & 
\textbf{1.20 \fontsize{8}{9}\selectfont$\pm$ 0.32} & 
\textbf{0.32 \fontsize{8}{9}\selectfont$\pm$ 0.14} \\
\hline
\multirow{4}{*}{GAT} & UnCal. & 14.46 \fontsize{8}{9}\selectfont$\pm$ 0.82 & 3.04 \fontsize{8}{9}\selectfont$\pm$ 0.63 & 3.51 \fontsize{8}{9}\selectfont$\pm$ 1.08 & 4.73 \fontsize{8}{9}\selectfont$\pm$ 1.76 & 1.53 \fontsize{8}{9}\selectfont$\pm$ 0.38 \\
& TS & 3.13 \fontsize{8}{9}\selectfont$\pm$ 0.74 & 2.35 \fontsize{8}{9}\selectfont$\pm$ 0.60 & 2.82 \fontsize{8}{9}\selectfont$\pm$ 0.80 & 3.98 \fontsize{8}{9}\selectfont$\pm$ 0.90 & 0.27 \fontsize{8}{9}\selectfont$\pm$ 0.29 \\
& GATS & 3.88 \fontsize{8}{9}\selectfont$\pm$ 0.67 & 2.78 \fontsize{8}{9}\selectfont$\pm$ 0.60 & 2.97 \fontsize{8}{9}\selectfont$\pm$ 0.89 & 3.97 \fontsize{8}{9}\selectfont$\pm$ 1.01 & 0.51 \fontsize{8}{9}\selectfont$\pm$ 0.23 \\
& \textbf{Ours} & 
\textbf{1.79 \fontsize{8}{9}\selectfont$\pm$ 0.51} 
& \textbf{1.42 \fontsize{8}{9}\selectfont$\pm$ 0.36} 
& \textbf{1.30 \fontsize{8}{9}\selectfont$\pm$ 0.43} 
& \textbf{1.24 \fontsize{8}{9}\selectfont$\pm$ 0.44} 
& \textbf{0.13 \fontsize{8}{9}\selectfont$\pm$ 0.06} \\
\hline\hline
\end{tabular}}
}
\label{table_hetero2}
\vskip -10pt
\end{table*}
}

%% file: Table_st.tex
{\renewcommand{\arraystretch}{1.02}
\begin{table*}[t]
\begin{minipage}{.49\textwidth}
\caption{Average node classification accuracy of various label rate (L/C) for our proposed calibration method and baselines on GCN.}
\centering
\resizebox{0.9\textwidth}{!}{
\footnotesize
{
\begin{tabular}
{ c | c | c | c | c }
\hline\hline

\textbf{Datasets} & {L/C} & {UnCal.} & {CaGCN} & \textbf{Ours}  \\

\hline

\multirow{3}{*}{Cora} & 20 & 
{81.46 \fontsize{7}{8}\selectfont$\pm$ 0.29} &
{82.94 \fontsize{7}{8}\selectfont$\pm$ 0.19} &
\textbf{82.98 \fontsize{7}{8}\selectfont$\pm$ 0.37} \\

& 40 &
{83.70 \fontsize{7}{8}\selectfont$\pm$ 0.26} &
{84.12 \fontsize{7}{8}\selectfont$\pm$ 0.27} &
\textbf{84.58 \fontsize{7}{8}\selectfont$\pm$ 0.12} \\

& 60 &
{84.40 \fontsize{7}{8}\selectfont$\pm$ 0.24} &
{85.54 \fontsize{7}{8}\selectfont$\pm$ 0.19} &
\textbf{86.06 \fontsize{7}{8}\selectfont$\pm$ 0.16} \\
\hline

\multirow{3}{*}{Citeseer} & 20 & 
{71.64 \fontsize{7}{8}\selectfont$\pm$ 0.16} &
\textbf{74.90 \fontsize{7}{8}\selectfont$\pm$ 0.23} &
{74.44 \fontsize{7}{8}\selectfont$\pm$ 0.19} \\

& 40 &
{72.02 \fontsize{7}{8}\selectfont$\pm$ 0.26} &
{75.26 \fontsize{7}{8}\selectfont$\pm$ 0.36} &
\textbf{75.30 \fontsize{7}{8}\selectfont$\pm$ 0.32} \\

& 60 &
{73.32 \fontsize{7}{8}\selectfont$\pm$ 0.18} &
{76.12 \fontsize{7}{8}\selectfont$\pm$ 0.16} &
\textbf{76.16 \fontsize{7}{8}\selectfont$\pm$ 0.23} \\
\hline

\multirow{3}{*}{Pubmed} & 20 & 
{79.52 \fontsize{7}{8}\selectfont$\pm$ 0.26} &
{81.20 \fontsize{7}{8}\selectfont$\pm$ 0.33} &
\textbf{81.32 \fontsize{7}{8}\selectfont$\pm$ 0.48} \\

& 40 &
{80.42 \fontsize{7}{8}\selectfont$\pm$ 0.26} &
{82.78 \fontsize{7}{8}\selectfont$\pm$ 0.35} &
\textbf{82.82 \fontsize{7}{8}\selectfont$\pm$ 0.21} \\

& 60 &
{83.32 \fontsize{7}{8}\selectfont$\pm$ 0.15} &
{84.12 \fontsize{7}{8}\selectfont$\pm$ 0.28} &
\textbf{84.28 \fontsize{7}{8}\selectfont$\pm$ 0.40} \\

\hline\hline
\end{tabular}}
}
\label{Table_st_gcn}
\end{minipage}
\hfill
\hfill
\begin{minipage}{.49\textwidth}
\caption{Average node classification accuracy of various label rate (L/C) for our proposed calibration method and baselines on GAT.}
\centering
\resizebox{0.9\textwidth}{!}{
\footnotesize
{
\begin{tabular}
{ c | c | c | c | c }
\hline\hline

\textbf{Datasets} & {L/C} & {UnCal.} & {CaGCN} & \textbf{Ours}  \\

\hline

\multirow{3}{*}{Cora} & 20 & 
{81.78 \fontsize{7}{8}\selectfont$\pm$ 0.35} &
{81.98 \fontsize{7}{8}\selectfont$\pm$ 0.73} &
\textbf{84.14 \fontsize{7}{8}\selectfont$\pm$ 0.39} \\

& 40 &
{83.48 \fontsize{7}{8}\selectfont$\pm$ 0.36} &
{84.32 \fontsize{7}{8}\selectfont$\pm$ 1.08} &
\textbf{85.64 \fontsize{7}{8}\selectfont$\pm$ 0.45} \\

& 60 &
{84.72 \fontsize{7}{8}\selectfont$\pm$ 0.32} &
{85.20 \fontsize{7}{8}\selectfont$\pm$ 0.75} &
\textbf{86.48 \fontsize{7}{8}\selectfont$\pm$ 0.32} \\
\hline

\multirow{3}{*}{Citeseer} & 20 & 
{70.82 \fontsize{7}{8}\selectfont$\pm$ 0.34} &
{73.86 \fontsize{7}{8}\selectfont$\pm$ 0.66} &
\textbf{74.40 \fontsize{7}{8}\selectfont$\pm$ 0.44} \\

& 40 &
{71.64 \fontsize{7}{8}\selectfont$\pm$ 0.34} &
{75.28 \fontsize{7}{8}\selectfont$\pm$ 0.34} &
\textbf{75.82 \fontsize{7}{8}\selectfont$\pm$ 0.25} \\

& 60 &
{73.20 \fontsize{7}{8}\selectfont$\pm$ 0.21} &
{76.04 \fontsize{7}{8}\selectfont$\pm$ 0.37} &
\textbf{76.42 \fontsize{7}{8}\selectfont$\pm$ 0.13} \\
\hline

\multirow{3}{*}{Pubmed} & 20 & 
{79.38 \fontsize{7}{8}\selectfont$\pm$ 0.35} &
{80.14 \fontsize{7}{8}\selectfont$\pm$ 0.36} &
\textbf{80.50 \fontsize{7}{8}\selectfont$\pm$ 0.24} \\

& 40 &
{80.84 \fontsize{7}{8}\selectfont$\pm$ 0.36} &
{82.60 \fontsize{7}{8}\selectfont$\pm$ 0.81} &
\textbf{82.82 \fontsize{7}{8}\selectfont$\pm$ 0.12} \\

& 60 &
{83.42 \fontsize{7}{8}\selectfont$\pm$ 0.28} &
{83.36 \fontsize{7}{8}\selectfont$\pm$ 0.38} &
\textbf{83.78 \fontsize{7}{8}\selectfont$\pm$ 0.12} \\

\hline\hline
\end{tabular}}
}
\label{Table_st_gat}
\end{minipage}
\vskip -5pt
\end{table*}
}

%% file: Table_recent.tex
{\renewcommand{\arraystretch}{1}
\begin{table*}[t]
\caption{ECE results (reported in percentage) for our method, compared to recent baselines. A lower ECE indicates better calibration performance.}
\centering
\resizebox{\textwidth}{!}{
\huge
{
\begin{tabular}
{ c | c | c | c | c | c | c | c | c | c }
\hline\hline

\multicolumn{2}{ c |}{\textbf{Datasets}} & {Cora} & {Citeseer} & {Pubmed} & {Computers} & {Photo} & {CS} & {Physics} &{CoraFull} \\

\hline

\multirow{4}{*}{GCN} 
& SimCalib & 
{3.32 \fontsize{15}{16}\selectfont$\pm$ 0.99} &
{3.94 \fontsize{15}{16}\selectfont$\pm$ 1.12} &
{0.93 \fontsize{15}{16}\selectfont$\pm$ 0.32} & 
{1.37 \fontsize{15}{16}\selectfont$\pm$ 0.33} & 
{1.36 \fontsize{15}{16}\selectfont$\pm$ 0.59} & 
{0.81 \fontsize{15}{16}\selectfont$\pm$ 0.30} & 
{0.39 \fontsize{15}{16}\selectfont$\pm$ 0.14} &
{3.22 \fontsize{15}{16}\selectfont$\pm$ 0.74} \\

& DCGC &
{4.27 \fontsize{15}{16}\selectfont$\pm$ 1.69} &
{5.30 \fontsize{15}{16}\selectfont$\pm$ 2.79} &
{3.05 \fontsize{15}{16}\selectfont$\pm$ 1.50} &
{2.67 \fontsize{15}{16}\selectfont$\pm$ 1.24} &
{1.54 \fontsize{15}{16}\selectfont$\pm$ 0.55} &
{1.78 \fontsize{15}{16}\selectfont$\pm$ 0.59} &
{0.60 \fontsize{15}{16}\selectfont$\pm$ 0.30} &
{4.78 \fontsize{15}{16}\selectfont$\pm$ 0.96} \\

& GETS &
{2.87 \fontsize{15}{16}\selectfont$\pm$ 0.53} &
{3.62 \fontsize{15}{16}\selectfont$\pm$ 1.13} &
{0.98 \fontsize{15}{16}\selectfont$\pm$ 0.35} &
{1.84 \fontsize{15}{16}\selectfont$\pm$ 0.37} &
{1.32 \fontsize{15}{16}\selectfont$\pm$ 0.17} &
{1.08 \fontsize{15}{16}\selectfont$\pm$ 0.08} &
{0.76 \fontsize{15}{16}\selectfont$\pm$ 0.10} &
\textbf{2.71 \fontsize{15}{16}\selectfont$\pm$ 0.29} \\

& {\textbf{Ours}} &
\textbf{1.97 \fontsize{15}{16}\selectfont$\pm$ 0.44} &
\textbf{2.66 \fontsize{15}{16}\selectfont$\pm$ 0.53} &
\textbf{0.75 \fontsize{15}{16}\selectfont$\pm$ 0.15} &
\textbf{1.02 \fontsize{15}{16}\selectfont$\pm$ 0.26} &
\textbf{1.01 \fontsize{15}{16}\selectfont$\pm$ 0.36} &
\textbf{0.58 \fontsize{15}{16}\selectfont$\pm$ 0.19} &
\textbf{0.28 \fontsize{15}{16}\selectfont$\pm$ 0.11} &
{3.46 \fontsize{15}{16}\selectfont$\pm$ 1.31} \\
\hline

\multirow{3}{*}{GAT} 
& SimCalib &
{2.90 \fontsize{15}{16}\selectfont$\pm$ 0.87} &
{3.95 \fontsize{15}{16}\selectfont$\pm$ 1.30} &
{0.95 \fontsize{15}{16}\selectfont$\pm$ 0.35} &
{1.08 \fontsize{15}{16}\selectfont$\pm$ 0.33} &
{1.29 \fontsize{15}{16}\selectfont$\pm$ 0.55} &
{0.83 \fontsize{15}{16}\selectfont$\pm$ 0.32} &
{0.40 \fontsize{15}{16}\selectfont$\pm$ 0.13} &
{3.40 \fontsize{15}{16}\selectfont$\pm$ 0.91} \\

& GETS &
{2.60 \fontsize{15}{16}\selectfont$\pm$ 0.82} &
{3.48 \fontsize{15}{16}\selectfont$\pm$ 0.32} &
{0.81 \fontsize{15}{16}\selectfont$\pm$ 0.18} &
{1.04 \fontsize{15}{16}\selectfont$\pm$ 0.32} &
{1.13 \fontsize{15}{16}\selectfont$\pm$ 0.36} &
{0.83 \fontsize{15}{16}\selectfont$\pm$ 0.11} &
\textbf{0.34 \fontsize{15}{16}\selectfont$\pm$ 0.31} &
\textbf{2.02 \fontsize{15}{16}\selectfont$\pm$ 0.51} \\

& {\textbf{Ours}} & 
\textbf{2.08 \fontsize{15}{16}\selectfont$\pm$ 0.45} &
\textbf{2.86 \fontsize{15}{16}\selectfont$\pm$ 0.56} &
\textbf{0.69 \fontsize{15}{16}\selectfont$\pm$ 0.16} &
\textbf{0.95 \fontsize{15}{16}\selectfont$\pm$ 0.37} &
\textbf{0.97 \fontsize{15}{16}\selectfont$\pm$ 0.53} &
\textbf{0.72 \fontsize{15}{16}\selectfont$\pm$ 0.43} &
{0.48 \fontsize{15}{16}\selectfont$\pm$ 0.22} &
{2.64 \fontsize{15}{16}\selectfont$\pm$ 1.02} \\
\hline

\multirow{2}{*}{SAGE} 
& DCGC &
{4.15 \fontsize{15}{16}\selectfont$\pm$ 2.17} &
{6.32 \fontsize{15}{16}\selectfont$\pm$ 2.02} &
{1.05 \fontsize{15}{16}\selectfont$\pm$ 0.41} &
{1.96 \fontsize{15}{16}\selectfont$\pm$ 0.47} &
{1.95 \fontsize{15}{16}\selectfont$\pm$ 0.67} &
{1.80 \fontsize{15}{16}\selectfont$\pm$ 0.50} &
{OOM} &
{4.94 \fontsize{15}{16}\selectfont$\pm$ 1.55} \\

& {\textbf{Ours}} &
\textbf{2.08 \fontsize{15}{16}\selectfont$\pm$ 0.52} &
\textbf{2.83 \fontsize{15}{16}\selectfont$\pm$ 0.80} &
\textbf{0.91 \fontsize{15}{16}\selectfont$\pm$ 0.17} &
\textbf{1.33 \fontsize{15}{16}\selectfont$\pm$ 0.57} &
\textbf{0.82 \fontsize{15}{16}\selectfont$\pm$ 0.34} &
\textbf{0.87 \fontsize{15}{16}\selectfont$\pm$ 0.34} &
{0.79 \fontsize{15}{16}\selectfont$\pm$ 0.31} &
\textbf{4.60 \fontsize{15}{16}\selectfont$\pm$ 0.90} \\
\hline\hline
\end{tabular}}
}
\label{Table_recent}
\vskip -10pt
\end{table*}
}

%% file: table_sparse.tex
{\renewcommand{\arraystretch}{1.}
\begin{table*}[t]
\centering
\caption{\blue{ECE results (reported in percentage) for our proposed calibration method and baselines on GCN under sparse label scenario (label rate = 20).}}
\resizebox{.95\textwidth}{!}{
\begin{tabular}{ c | c | c | c | c | c | c | c | c }
\hline\hline
\textbf{GCN} & Cora & Citeseer & Pubmed & Computers & Photo & CS & Physics & CoraFull \\ \hline
UnCal. & 
22.15 \fontsize{8}{9}\selectfont$\pm$ 12.32 & 
22.14 \fontsize{8}{9}\selectfont$\pm$ 10.94 & 
7.91 \fontsize{8}{9}\selectfont$\pm$ 1.34 & 
11.00 \fontsize{8}{9}\selectfont$\pm$ 5.10 & 
5.89 \fontsize{8}{9}\selectfont$\pm$ 3.80 & 
28.93 \fontsize{8}{9}\selectfont$\pm$ 27.04 & 
19.95 \fontsize{8}{9}\selectfont$\pm$ 17.81 & 
13.27 \fontsize{8}{9}\selectfont$\pm$ 4.43 \\
GATS & 
4.01 \fontsize{8}{9}\selectfont$\pm$ 0.88 & 
5.81 \fontsize{8}{9}\selectfont$\pm$ 1.59 & 
4.16 \fontsize{8}{9}\selectfont$\pm$ 0.98 & 
7.62 \fontsize{8}{9}\selectfont$\pm$ 1.84 & 
3.45 \fontsize{8}{9}\selectfont$\pm$ 0.88 & 
2.66 \fontsize{8}{9}\selectfont$\pm$ 0.48 & 
2.87 \fontsize{8}{9}\selectfont$\pm$ 2.22 & 
9.29 \fontsize{8}{9}\selectfont$\pm$ 1.56 \\
\textbf{Ours} & 
\textbf{2.95 \fontsize{8}{9}\selectfont$\pm$ 0.99}
& \textbf{3.99 \fontsize{8}{9}\selectfont$\pm$ 1.06}
& \textbf{3.02 \fontsize{8}{9}\selectfont$\pm$ 0.97}
& \textbf{3.35 \fontsize{8}{9}\selectfont$\pm$ 0.83}
& \textbf{1.97 \fontsize{8}{9}\selectfont$\pm$ 0.37}
& \textbf{2.34 \fontsize{8}{9}\selectfont$\pm$ 0.65}
& \textbf{1.81 \fontsize{8}{9}\selectfont$\pm$ 0.45}
& \textbf{7.53 \fontsize{8}{9}\selectfont$\pm$ 2.77} \\ \hline\hline
\end{tabular}}
\label{table_sparse_gcn}
\vskip -5pt
\end{table*}}

{\renewcommand{\arraystretch}{.99}
\begin{table*}[t]
\centering
\caption{\blue{ECE results (reported in percentage) for our proposed calibration method and baselines on GAT under sparse label scenario (label rate = 20).}}
\resizebox{.95\textwidth}{!}{
\begin{tabular}{ c | c | c | c | c | c | c | c | c }
\hline\hline
\textbf{GAT} & Cora & Citeseer & Pubmed & Computers & Photo & CS & Physics & CoraFull \\ \hline
UnCal. & 
33.39 \fontsize{8}{9}\selectfont$\pm$ 18.78 & 
25.77 \fontsize{8}{9}\selectfont$\pm$ 11.88 & 
6.51 \fontsize{8}{9}\selectfont$\pm$ 3.91 & 
15.06 \fontsize{8}{9}\selectfont$\pm$ 19.85 & 
3.22 \fontsize{8}{9}\selectfont$\pm$ 1.44 & 
28.15 \fontsize{8}{9}\selectfont$\pm$ 33.97 & 
13.20 \fontsize{8}{9}\selectfont$\pm$ 14.27 & 
10.17 \fontsize{8}{9}\selectfont$\pm$ 4.61 \\
GATS & 
5.04 \fontsize{8}{9}\selectfont$\pm$ 4.19 & 
5.64 \fontsize{8}{9}\selectfont$\pm$ 1.26 & 
3.39 \fontsize{8}{9}\selectfont$\pm$ 0.69 & 
5.01 \fontsize{8}{9}\selectfont$\pm$ 1.43 & 
2.83 \fontsize{8}{9}\selectfont$\pm$ 1.47 & 
2.68 \fontsize{8}{9}\selectfont$\pm$ 0.69 & 
\textbf{2.33 \fontsize{8}{9}\selectfont$\pm$ 0.71} & 
8.04 $\pm$ 1.18 \\
\textbf{Ours} & 
\textbf{3.64 \fontsize{8}{9}\selectfont$\pm$ 2.17}
& \textbf{4.24 \fontsize{8}{9}\selectfont$\pm$ 0.91}
& \textbf{2.83 \fontsize{8}{9}\selectfont$\pm$ 0.80}
& \textbf{3.59 \fontsize{8}{9}\selectfont$\pm$ 1.05}
& \textbf{2.10 \fontsize{8}{9}\selectfont$\pm$ 0.43}
& \textbf{2.61 \fontsize{8}{9}\selectfont$\pm$ 0.78}
& 2.91 \fontsize{8}{9}\selectfont$\pm$ 1.23
& \textbf{7.90 \fontsize{8}{9}\selectfont$\pm$ 1.07} \\ \hline\hline
\end{tabular}}
\label{table_sparse_gat}
\vskip -10pt
\end{table*}}

%% file: Table_ood.tex
{\renewcommand{\arraystretch}{0.85}
  \begin{table}[t]
    \caption{Out-of-distribution detection performance evaluated by AUROC for our proposed calibration method and baselines. A lower AUROC indicates better detection performance.}
    \vskip -5pt
    \centering
    \resizebox{.46\textwidth}{!}{
    \small{
    \begin{tabular}{ c | c | c | c | c }
     \hline\hline
     \textbf{Datasets} & \textbf{Perturbation} & {GPN} & {GNNSafe} & {{\textbf{Ours}}}\\
     \hline
     \multirow{2}{*}{Cora} & Feature & 
     {85.88} & 
     {93.44} & 
     {83.92} \\
     & Structure &
     {77.47} & 
     {87.52} & 
     {68.91}\\
     \hline
     
     \multirow{2}{*}{Photo} & Feature & 
     {87.91} & 
     {98.55} & 
     {97.79} \\
     & Structure &
     {97.17} & 
     {99.58} & 
     {99.39}\\
     \hline

     \multirow{2}{*}{{Physics}} & {Feature} & 
     {72.56} & 
     {99.64} & 
     {98.63} \\
     & {Structure} &
     {34.67} & 
     {99.60} & 
     {97.94}\\
     
    \hline\hline
    \end{tabular}}}
    \label{Table_ood}
    \vskip -5pt
  \end{table}
}

%% file: table_mlp.tex
{\renewcommand{\arraystretch}{1.}
\begin{table*}[ht!]
\centering
\caption{ECE results (reported in percentage) for our proposed calibration method and baselines on MLP, averaged over 75 runs.}
\resizebox{.95\textwidth}{!}{
\begin{tabular}{ c | c | c | c | c | c | c | c | c }
\hline\hline
\textbf{MLP} & Cora & Citeseer & Pubmed & Computers & Photo & CS & Physics & CoraFull \\ \hline
UnCal. & 5.93 $\pm$ 1.77 & 11.79 $\pm$ 5.97 & 5.10 $\pm$ 2.59 & 7.02 $\pm$ 0.51 & 4.01 $\pm$ 1.15 & 1.67 $\pm$ 1.18 & 0.66 $\pm$ 0.50 & 7.02 $\pm$ 2.54 \\
GATS & 4.04 $\pm$ 1.14 & 4.31 $\pm$ 1.36 & 1.12 $\pm$ 0.32 & 5.56 $\pm$ 1.66 & 3.64 $\pm$ 0.78 & 1.18 $\pm$ 0.39 & 0.56 $\pm$ 0.17 & 3.03 $\pm$ 0.58 \\
\textbf{Ours} & \textbf{2.94 $\pm$ 0.79} 
& \textbf{3.04 $\pm$ 0.67} 
& \textbf{1.01 $\pm$ 0.16} 
& \textbf{3.28 $\pm$ 1.94} 
& \textbf{1.98 $\pm$ 0.62}
& \textbf{0.86 $\pm$ 0.35} 
& \textbf{0.38 $\pm$ 0.11} 
& \textbf{2.75 $\pm$ 0.99} \\ \hline\hline
\end{tabular}}
\label{table_mlp}
\vskip -5pt
\end{table*}}

{\renewcommand{\arraystretch}{1.}
\begin{table*}[ht!]
\centering
\caption{ECE results (reported in percentage) for our proposed calibration method and baselines on SAGE, averaged over 75 runs.}
\resizebox{.95\textwidth}{!}{
\begin{tabular}{ c | c | c | c | c | c | c | c | c }
\hline\hline
\textbf{SAGE} & Cora & Citeseer & Pubmed & Computers & Photo & CS & Physics & CoraFull \\ \hline
UnCal. & 8.15 $\pm$ 4.45 & 7.08 $\pm$ 4.87 & 1.47 $\pm$ 0.85 & 2.07 $\pm$ 0.36 & 1.47 $\pm$ 0.67 & 1.11 $\pm$ 0.45 & 0.91 $\pm$ 0.35 & 5.24 $\pm$ 0.69 \\
GATS & 3.61 $\pm$ 0.95 & 4.85 $\pm$ 1.60 & 1.10 $\pm$ 0.44 & 1.76 $\pm$ 0.33 & 1.24 $\pm$ 0.49 & 0.97 $\pm$ 0.26 & \textbf{0.51 $\pm$ 0.17} & \textbf{4.13 $\pm$ 0.50} \\
\textbf{Ours} & \textbf{2.08 $\pm$ 0.52} 
& \textbf{2.83 $\pm$ 0.80} 
& \textbf{0.91 $\pm$ 0.17} 
& \textbf{1.33 $\pm$ 0.57} 
& \textbf{0.82 $\pm$ 0.31} 
& \textbf{0.87 $\pm$ 0.34} 
& 0.79 $\pm$ 0.31 
& 4.60 $\pm$ 0.90 \\ \hline\hline
\end{tabular}}
\label{table_sage}
\vskip -5pt
\end{table*}}

{\renewcommand{\arraystretch}{1.}
\begin{table*}[ht!]
\centering
\caption{ECE results (reported in percentage) for our proposed calibration method and baselines on GIN, averaged over 75 runs.}
\resizebox{.95\textwidth}{!}{
\begin{tabular}{ c | c | c | c | c | c | c | c | c }
\hline\hline
\textbf{GIN} & Cora & Citeseer & Pubmed & Computers & Photo & CS & Physics & CoraFull \\ \hline
UnCal. & 11.82 $\pm$ 5.41 & 13.90 $\pm$ 5.02 & 7.16 $\pm$ 2.91 & 8.16 $\pm$ 3.54 & 6.18 $\pm$ 6.59 & 7.23 $\pm$ 2.06 & 4.79 $\pm$ 2.22 & 19.74 $\pm$ 1.09 \\
GATS & 8.34 $\pm$ 2.17 & 10.25 $\pm$ 2.67 & 4.47 $\pm$ 1.86 & 5.73 $\pm$ 2.36 & 3.85 $\pm$ 2.72 & 6.93 $\pm$ 1.59 & 3.55 $\pm$ 1.11 & 17.88 $\pm$ 1.60 \\
\textbf{Ours} & \textbf{6.09 $\pm$ 2.00}
& \textbf{6.91 $\pm$ 2.80}
& \textbf{1.70 $\pm$ 0.43}
& \textbf{1.50 $\pm$ 0.62}
& \textbf{1.55 $\pm$ 1.61}
& \textbf{3.18 $\pm$ 0.93}
& \textbf{1.18 $\pm$ 0.42}
& \textbf{8.23 $\pm$ 4.52} \\ \hline\hline
\end{tabular}}
\label{table_gin}
\vskip -10pt
\end{table*}}

%% file: Table_kde.tex
{\renewcommand{\arraystretch}{1.}
  \begin{table*}[t]
    \caption{KDE-ECE results (reported in percentage) for our proposed calibration method and baselines, averaged over 75 repetitions ($\pm$ STD). A lower value indicates better calibration performance.}
    \centering
    \resizebox{.95\textwidth}{!}{
    \huge{
    \begin{tabular}{ c | c | c | c | c | c | c | c | c }
     \hline\hline
     \multicolumn{2}{ c |}{\textbf{Methods}} & {{UnCal.}} & {{TS}} & {{VS}} & {{ETS}} & {{CaGCN}} & {{GATS}} & {{\textbf{Ours}}}\\
     \hline
     \hline
     \multirow{2}{*}{Cora} & GCN & 
     {12.76 \fontsize{15}{16}\selectfont$\pm$ 4.09} & 
     {3.13 \fontsize{15}{16}\selectfont$\pm$ 1.11} & 
     {3.52 \fontsize{15}{16}\selectfont$\pm$ 1.17} & 
     {3.16 \fontsize{15}{16}\selectfont$\pm$ 1.11} & 
     {4.01 \fontsize{15}{16}\selectfont$\pm$ 1.40} & 
     {2.95 \fontsize{15}{16}\selectfont$\pm$ 1.23} & 
     \textbf{{1.88 \fontsize{15}{16}\selectfont$\pm$ 0.26}}\\
     & GAT & 
     {15.02 \fontsize{15}{16}\selectfont$\pm$ 4.23} &
     {2.85 \fontsize{15}{16}\selectfont$\pm$ 0.80} & 
     {2.84 \fontsize{15}{16}\selectfont$\pm$ 0.85} & 
     {2.86 \fontsize{15}{16}\selectfont$\pm$ 0.76} & 
     {3.38 \fontsize{15}{16}\selectfont$\pm$ 1.29} & 
     {2.64 \fontsize{15}{16}\selectfont$\pm$ 0.77} & 
     \textbf{{1.98 \fontsize{15}{16}\selectfont$\pm$ 0.42}}\\
     \hline
     
     \multirow{2}{*}{Citeseer} & GCN & 
     {12.59 \fontsize{15}{16}\selectfont$\pm$ 8.53} & 
     {4.90 \fontsize{15}{16}\selectfont$\pm$ 1.67} & 
     {4.79 \fontsize{15}{16}\selectfont$\pm$ 1.47} & 
     {4.74 \fontsize{15}{16}\selectfont$\pm$ 1.72} & 
     {6.07 \fontsize{15}{16}\selectfont$\pm$ 1.76} & 
     {4.25 \fontsize{15}{16}\selectfont$\pm$ 1.49} & 
     \textbf{{2.40 \fontsize{15}{16}\selectfont$\pm$ 0.43}}\\
     & GAT & 
     {16.64 \fontsize{15}{16}\selectfont$\pm$ 7.96} &
     {4.74 \fontsize{15}{16}\selectfont$\pm$ 1.42} & 
     {4.29 \fontsize{15}{16}\selectfont$\pm$ 1.15} & 
     {4.67 \fontsize{15}{16}\selectfont$\pm$ 1.40} & 
     {4.57 \fontsize{15}{16}\selectfont$\pm$ 1.79} & 
     {3.84 \fontsize{15}{16}\selectfont$\pm$ 1.45} & 
     \textbf{{2.66 \fontsize{15}{16}\selectfont$\pm$ 0.50}}\\
     \hline
     
     \multirow{2}{*}{Pubmed} & GCN & 
     {7.44 \fontsize{15}{16}\selectfont$\pm$ 1.53} & 
     {1.33 \fontsize{15}{16}\selectfont$\pm$ 0.28} & 
     {1.58 \fontsize{15}{16}\selectfont$\pm$ 0.38} & 
     {1.38 \fontsize{15}{16}\selectfont$\pm$ 0.29} & 
     {1.25 \fontsize{15}{16}\selectfont$\pm$ 0.35} & 
     {1.06 \fontsize{15}{16}\selectfont$\pm$ 0.26} & 
     \textbf{{0.93 \fontsize{15}{16}\selectfont$\pm$ 0.12}}\\
     & GAT & 
     {10.38 \fontsize{15}{16}\selectfont$\pm$ 1.88} &
     {1.18 \fontsize{15}{16}\selectfont$\pm$ 0.35} & 
     {1.13 \fontsize{15}{16}\selectfont$\pm$ 0.31} & 
     {1.18 \fontsize{15}{16}\selectfont$\pm$ 0.35} & 
     {1.08 \fontsize{15}{16}\selectfont$\pm$ 0.29} & 
     {1.11 \fontsize{15}{16}\selectfont$\pm$ 0.34} & 
     \textbf{0.79 \fontsize{15}{16}\selectfont$\pm$ 0.12}\\
     \hline
     
     \multirow{2}{*}{Computers} & GCN & 
     {3.01 \fontsize{15}{16}\selectfont$\pm$ 0.91} & 
     {2.60 \fontsize{15}{16}\selectfont$\pm$ 0.65} & 
     {2.72 \fontsize{15}{16}\selectfont$\pm$ 0.74} & 
     {2.73 \fontsize{15}{16}\selectfont$\pm$ 0.76} & 
     {1.58 \fontsize{15}{16}\selectfont$\pm$ 0.45} & 
     {2.12 \fontsize{15}{16}\selectfont$\pm$ 0.60} & 
     \textbf{{1.27 \fontsize{15}{16}\selectfont$\pm$ 0.15}}\\
     & GAT & 
     {1.70 \fontsize{15}{16}\selectfont$\pm$ 0.63} &
     {1.56 \fontsize{15}{16}\selectfont$\pm$ 0.42} & 
     {1.59 \fontsize{15}{16}\selectfont$\pm$ 0.46} & 
     {1.63 \fontsize{15}{16}\selectfont$\pm$ 0.47} & 
     {1.64 \fontsize{15}{16}\selectfont$\pm$ 0.46} & 
     {1.52 \fontsize{15}{16}\selectfont$\pm$ 0.40} & 
     \textbf{{1.10 \fontsize{15}{16}\selectfont$\pm$ 0.20}}\\
     \hline

     \multirow{2}{*}{Photo} & GCN & 
     {2.45 \fontsize{15}{16}\selectfont$\pm$ 1.23} & 
     {1.81 \fontsize{15}{16}\selectfont$\pm$ 0.92} & 
     {1.95 \fontsize{15}{16}\selectfont$\pm$ 0.95} & 
     {1.88 \fontsize{15}{16}\selectfont$\pm$ 0.98} & 
     {1.64 \fontsize{15}{16}\selectfont$\pm$ 0.42} & 
     {1.65 \fontsize{15}{16}\selectfont$\pm$ 0.68} & 
     \textbf{{1.13 \fontsize{15}{16}\selectfont$\pm$ 0.18}}\\
     & GAT & 
     {2.42 \fontsize{15}{16}\selectfont$\pm$ 1.60} &
     {1.70 \fontsize{15}{16}\selectfont$\pm$ 0.73} & 
     {1.79 \fontsize{15}{16}\selectfont$\pm$ 0.78} & 
     {1.77 \fontsize{15}{16}\selectfont$\pm$ 0.80} & 
     {1.73 \fontsize{15}{16}\selectfont$\pm$ 0.62} & 
     {1.72 \fontsize{15}{16}\selectfont$\pm$ 0.70} & 
     \textbf{{1.19 \fontsize{15}{16}\selectfont$\pm$ 0.29}}\\
     \hline

     \multirow{2}{*}{CS} & GCN & 
     {2.19 \fontsize{15}{16}\selectfont$\pm$ 1.33} & 
     {1.12 \fontsize{15}{16}\selectfont$\pm$ 0.10} & 
     {1.12 \fontsize{15}{16}\selectfont$\pm$ 0.18} & 
     {1.12 \fontsize{15}{16}\selectfont$\pm$ 0.10} & 
     {1.94 \fontsize{15}{16}\selectfont$\pm$ 0.90} & 
     {1.08 \fontsize{15}{16}\selectfont$\pm$ 0.12} & 
     \textbf{{0.94 \fontsize{15}{16}\selectfont$\pm$ 0.11}}\\
     & GAT & 
     {1.77 \fontsize{15}{16}\selectfont$\pm$ 0.91} &
     {1.12 \fontsize{15}{16}\selectfont$\pm$ 0.23} & 
     {1.12 \fontsize{15}{16}\selectfont$\pm$ 0.25} & 
     {1.13 \fontsize{15}{16}\selectfont$\pm$ 0.23} & 
     {1.90 \fontsize{15}{16}\selectfont$\pm$ 0.97} & 
     {1.13 \fontsize{15}{16}\selectfont$\pm$ 0.20} & 
     \textbf{{0.95 \fontsize{15}{16}\selectfont$\pm$ 0.22}}\\
     \hline

     \multirow{2}{*}{Physics} & GCN & 
     {0.97 \fontsize{15}{16}\selectfont$\pm$ 0.31} & 
     {0.83 \fontsize{15}{16}\selectfont$\pm$ 0.09} & 
     {0.82 \fontsize{15}{16}\selectfont$\pm$ 0.07} & 
     {0.83 \fontsize{15}{16}\selectfont$\pm$ 0.09} & 
     {0.93 \fontsize{15}{16}\selectfont$\pm$ 0.19} & 
     {0.85 \fontsize{15}{16}\selectfont$\pm$ 0.09} & 
     \textbf{{0.70 \fontsize{15}{16}\selectfont$\pm$ 0.61}}\\
     & GAT & 
     {0.86 \fontsize{15}{16}\selectfont$\pm$ 0.15} & 
     {0.84 \fontsize{15}{16}\selectfont$\pm$ 0.10} & 
     {0.86 \fontsize{15}{16}\selectfont$\pm$ 0.09} & 
     {0.84 \fontsize{15}{16}\selectfont$\pm$ 0.10} & 
     {1.03 \fontsize{15}{16}\selectfont$\pm$ 0.21} & 
     {{0.82 \fontsize{15}{16}\selectfont$\pm$ 0.08}} & 
     \textbf{0.80 \fontsize{15}{16}\selectfont$\pm$ 0.11}\\
     \hline

     \multirow{2}{*}{CoraFull} & GCN & 
     {6.44 \fontsize{15}{16}\selectfont$\pm$ 1.33} &
     {5.46 \fontsize{15}{16}\selectfont$\pm$ 0.44} & 
     {5.68 \fontsize{15}{16}\selectfont$\pm$ 0.41} & 
     {5.42 \fontsize{15}{16}\selectfont$\pm$ 0.46} & 
     {5.74 \fontsize{15}{16}\selectfont$\pm$ 0.46} & 
     {3.70 \fontsize{15}{16}\selectfont$\pm$ 0.65} & 
     \textbf{{3.43 \fontsize{15}{16}\selectfont$\pm$ 1.27}}\\
     & GAT & 
     {5.26 \fontsize{15}{16}\selectfont$\pm$ 1.38} &
     {4.34 \fontsize{15}{16}\selectfont$\pm$ 0.48} & 
     {4.36 \fontsize{15}{16}\selectfont$\pm$ 0.46} & 
     {4.30 \fontsize{15}{16}\selectfont$\pm$ 0.48} & 
     {6.59 \fontsize{15}{16}\selectfont$\pm$ 3.62} & 
     {3.46 \fontsize{15}{16}\selectfont$\pm$ 0.45} & 
     \textbf{{2.64 \fontsize{15}{16}\selectfont$\pm$ 0.98}}\\
     \hline
    \hline\hline
    \end{tabular}}}
    \label{Table_kde}
\vskip -10pt
  \end{table*}
}

%% file: Table_class_ece.tex
{\renewcommand{\arraystretch}{1.}
  \begin{table*}[t]
    \caption{Class-wise ECE results (reported in percentage) for our proposed calibration method and baselines, averaged over 75 repetitions ($\pm$ STD). A lower value indicates better calibration performance.}
    \centering
    \resizebox{.95\textwidth}{!}{
    \huge{
    \begin{tabular}{ c | c | c | c | c | c | c | c | c }
     \hline\hline
     \multicolumn{2}{ c |}{\textbf{Methods}} & {{UnCal.}} & {{TS}} & {{VS}} & {{ETS}} & {{CaGCN}} & {{GATS}} & {{\textbf{Ours}}}\\
     \hline
     \hline
     \multirow{2}{*}{Cora} & GCN & 
     {4.14 \fontsize{15}{16}\selectfont$\pm$ 1.10} & 
     {2.03 \fontsize{15}{16}\selectfont$\pm$ 0.23} & 
     {2.09 \fontsize{15}{16}\selectfont$\pm$ 0.27} & 
     {2.03 \fontsize{15}{16}\selectfont$\pm$ 0.23} & 
     {2.21 \fontsize{15}{16}\selectfont$\pm$ 0.28} & 
     {1.99 \fontsize{15}{16}\selectfont$\pm$ 0.24} & 
     \textbf{{1.82 \fontsize{15}{16}\selectfont$\pm$ 0.19}}\\
     & GAT & 
     {4.78 \fontsize{15}{16}\selectfont$\pm$ 1.18} &
     {1.95 \fontsize{15}{16}\selectfont$\pm$ 0.23} & 
     {1.94 \fontsize{15}{16}\selectfont$\pm$ 0.25} & 
     {1.94 \fontsize{15}{16}\selectfont$\pm$ 0.23} & 
     {2.10 \fontsize{15}{16}\selectfont$\pm$ 0.29} & 
     {1.92 \fontsize{15}{16}\selectfont$\pm$ 0.24} & 
     \textbf{{1.80 \fontsize{15}{16}\selectfont$\pm$ 0.22}}\\
     \hline
     
     \multirow{2}{*}{Citeseer} & GCN & 
     {5.11 \fontsize{15}{16}\selectfont$\pm$ 2.77} & 
     {2.97 \fontsize{15}{16}\selectfont$\pm$ 0.65} & 
     {2.80 \fontsize{15}{16}\selectfont$\pm$ 0.43} & 
     {2.94 \fontsize{15}{16}\selectfont$\pm$ 0.69} & 
     {3.24 \fontsize{15}{16}\selectfont$\pm$ 0.78} & 
     {2.88 \fontsize{15}{16}\selectfont$\pm$ 0.78} &  
     \textbf{{2.53 \fontsize{15}{16}\selectfont$\pm$ 0.55}}\\
     & GAT & 
     {6.39 \fontsize{15}{16}\selectfont$\pm$ 2.52} &
     {3.03 \fontsize{15}{16}\selectfont$\pm$ 0.47} & 
     {2.85 \fontsize{15}{16}\selectfont$\pm$ 0.48} & 
     {3.02 \fontsize{15}{16}\selectfont$\pm$ 0.48} & 
     {3.07 \fontsize{15}{16}\selectfont$\pm$ 0.69} & 
     {2.96 \fontsize{15}{16}\selectfont$\pm$ 0.54} & 
     \textbf{{2.71 \fontsize{15}{16}\selectfont$\pm$ 0.44}}\\
     \hline
     
     \multirow{2}{*}{Pubmed} & GCN & 
     {5.04 \fontsize{15}{16}\selectfont$\pm$ 1.04} & 
     {1.39 \fontsize{15}{16}\selectfont$\pm$ 0.28} & 
     {1.54 \fontsize{15}{16}\selectfont$\pm$ 0.31} & 
     {1.40 \fontsize{15}{16}\selectfont$\pm$ 0.27} & 
     {1.33 \fontsize{15}{16}\selectfont$\pm$ 0.32} & 
     {1.26 \fontsize{15}{16}\selectfont$\pm$ 0.28} & 
     \textbf{{1.17 \fontsize{15}{16}\selectfont$\pm$ 0.23}}\\
     & GAT & 
     {7.19 \fontsize{15}{16}\selectfont$\pm$ 1.22} &
     {1.77 \fontsize{15}{16}\selectfont$\pm$ 0.40} & 
     {1.75 \fontsize{15}{16}\selectfont$\pm$ 0.30} & 
     {1.77 \fontsize{15}{16}\selectfont$\pm$ 0.40} & 
     {1.67 \fontsize{15}{16}\selectfont$\pm$ 0.39} & 
     {1.79 \fontsize{15}{16}\selectfont$\pm$ 0.36} & 
     \textbf{1.63 \fontsize{15}{16}\selectfont$\pm$ 0.32}\\
     \hline
     
     \multirow{2}{*}{Computers} & GCN & 
     {0.96 \fontsize{15}{16}\selectfont$\pm$ 0.16} & 
     {0.92 \fontsize{15}{16}\selectfont$\pm$ 0.11} & 
     {0.91 \fontsize{15}{16}\selectfont$\pm$ 0.13} & 
     {0.94 \fontsize{15}{16}\selectfont$\pm$ 0.13} & 
     {0.83 \fontsize{15}{16}\selectfont$\pm$ 0.10} & 
     {0.88 \fontsize{15}{16}\selectfont$\pm$ 0.08} & 
     \textbf{{0.81 \fontsize{15}{16}\selectfont$\pm$ 0.08}}\\
     & GAT & 
     {0.80 \fontsize{15}{16}\selectfont$\pm$ 0.13} &
     {0.78 \fontsize{15}{16}\selectfont$\pm$ 0.10} & 
     {0.76 \fontsize{15}{16}\selectfont$\pm$ 0.09} & 
     {0.80 \fontsize{15}{16}\selectfont$\pm$ 0.11} & 
     {0.80 \fontsize{15}{16}\selectfont$\pm$ 0.10} & 
     {0.78 \fontsize{15}{16}\selectfont$\pm$ 0.10} & 
     \textbf{{0.74 \fontsize{15}{16}\selectfont$\pm$ 0.09}}\\
     \hline

     \multirow{2}{*}{Photo} & GCN & 
     {0.86 \fontsize{15}{16}\selectfont$\pm$ 0.21} & 
     {0.78 \fontsize{15}{16}\selectfont$\pm$ 0.14} & 
     {0.81 \fontsize{15}{16}\selectfont$\pm$ 0.15} & 
     {0.78 \fontsize{15}{16}\selectfont$\pm$ 0.16} & 
     {0.79 \fontsize{15}{16}\selectfont$\pm$ 0.08} & 
     {0.76 \fontsize{15}{16}\selectfont$\pm$ 0.11} & 
     \textbf{{0.67 \fontsize{15}{16}\selectfont$\pm$ 0.05}}\\
     & GAT & 
     {0.96 \fontsize{15}{16}\selectfont$\pm$ 0.36} &
     {0.84 \fontsize{15}{16}\selectfont$\pm$ 0.17} & 
     {0.82 \fontsize{15}{16}\selectfont$\pm$ 0.16} & 
     {0.84 \fontsize{15}{16}\selectfont$\pm$ 0.19} & 
     {0.86 \fontsize{15}{16}\selectfont$\pm$ 0.13} & 
     {0.83 \fontsize{15}{16}\selectfont$\pm$ 0.18} & 
     \textbf{{0.74 \fontsize{15}{16}\selectfont$\pm$ 0.10}}\\
     \hline

     \multirow{2}{*}{CS} & GCN & 
     {0.40 \fontsize{15}{16}\selectfont$\pm$ 0.15} & 
     {0.30 \fontsize{15}{16}\selectfont$\pm$ 0.03} & 
     {0.32 \fontsize{15}{16}\selectfont$\pm$ 0.03} & 
     \textbf{0.29 \fontsize{15}{16}\selectfont$\pm$ 0.03} & 
     {0.42 \fontsize{15}{16}\selectfont$\pm$0.10} & 
     \textbf{0.29 \fontsize{15}{16}\selectfont$\pm$ 0.03} & 
     \textbf{{0.29 \fontsize{15}{16}\selectfont$\pm$ 0.03}}\\
     & GAT & 
     {0.39 \fontsize{15}{16}\selectfont$\pm$ 0.10} &
     {0.34 \fontsize{15}{16}\selectfont$\pm$ 0.03} & 
     {0.34 \fontsize{15}{16}\selectfont$\pm$ 0.03} & 
     {0.34 \fontsize{15}{16}\selectfont$\pm$ 0.03} & 
     {0.44 \fontsize{15}{16}\selectfont$\pm$ 0.10} & 
     {0.34 \fontsize{15}{16}\selectfont$\pm$ 0.04} & 
     \textbf{{0.33 \fontsize{15}{16}\selectfont$\pm$ 0.03}}\\
     \hline

     \multirow{2}{*}{Physics} & GCN & 
     {0.41 \fontsize{15}{16}\selectfont$\pm$ 0.33} & 
     {0.36 \fontsize{15}{16}\selectfont$\pm$ 0.06} & 
     {0.34 \fontsize{15}{16}\selectfont$\pm$ 0.04} & 
     {0.36 \fontsize{15}{16}\selectfont$\pm$ 0.06} & 
     {0.46 \fontsize{15}{16}\selectfont$\pm$ 0.14} & 
     {0.36 \fontsize{15}{16}\selectfont$\pm$ 0.05} & 
     \textbf{{0.33 \fontsize{15}{16}\selectfont$\pm$ 0.04}}\\
     & GAT & 
     {0.40 \fontsize{15}{16}\selectfont$\pm$ 0.08} & 
     {0.39 \fontsize{15}{16}\selectfont$\pm$ 0.07} & 
     \textbf{0.37 \fontsize{15}{16}\selectfont$\pm$ 0.05} & 
     {0.39 \fontsize{15}{16}\selectfont$\pm$ 0.07} & 
     {0.52 \fontsize{15}{16}\selectfont$\pm$ 0.13} & 
     {{0.38 \fontsize{15}{16}\selectfont$\pm$ 0.07}} & 
     {0.38 \fontsize{15}{16}\selectfont$\pm$ 0.06}\\
     \hline

     \multirow{2}{*}{CoraFull} & GCN & 
     {0.35 \fontsize{15}{16}\selectfont$\pm$ 0.04} &
     {0.33 \fontsize{15}{16}\selectfont$\pm$ 0.02} & 
     {0.34 \fontsize{15}{16}\selectfont$\pm$ 0.01} & 
     {0.33 \fontsize{15}{16}\selectfont$\pm$ 0.02} & 
     {0.34 \fontsize{15}{16}\selectfont$\pm$ 0.05} & 
     {0.33 \fontsize{15}{16}\selectfont$\pm$ 0.02} & 
     \textbf{{0.32 \fontsize{15}{16}\selectfont$\pm$ 0.01}}\\
     & GAT & 
     {0.34 \fontsize{15}{16}\selectfont$\pm$ 0.03} &
     {0.32 \fontsize{15}{16}\selectfont$\pm$ 0.01} & 
     {0.32 \fontsize{15}{16}\selectfont$\pm$ 0.01} & 
     {0.32 \fontsize{15}{16}\selectfont$\pm$ 0.01} & 
     {0.35 \fontsize{15}{16}\selectfont$\pm$ 0.07} & 
     \textbf{0.31 \fontsize{15}{16}\selectfont$\pm$ 0.01} & 
     \textbf{{0.31 \fontsize{15}{16}\selectfont$\pm$ 0.01}}\\
     \hline
    \hline\hline
    \end{tabular}}}
    \label{Table_class_ece}
  \end{table*}
}

%% file: Table_nll.tex
{\renewcommand{\arraystretch}{1.2}
  \begin{table*}[t!]
    \caption{NLL results (reported in percentage) for our proposed calibration method and baselines, averaged over 75 repetitions ($\pm$ STD). A lower value indicates better calibration performance. }
    \centering
    \resizebox{.98\textwidth}{!}{
    \huge{
    \begin{tabular}{ c | c | c | c | c | c | c | c | c }
     \hline\hline
     \multicolumn{2}{ c |}{\textbf{Methods}} & {{UnCal.}} & {{TS}} & {{VS}} & {{ETS}} & {{CaGCN}} & {{GATS}} & {{\textbf{Ours}}}\\
     \hline
     \hline
     \multirow{2}{*}{Cora} & GCN & 
     {0.6199 \fontsize{15}{16}\selectfont$\pm$ 0.0444} & 
     {0.5613 \fontsize{15}{16}\selectfont$\pm$ 0.0302} & 
     {0.5747 \fontsize{15}{16}\selectfont$\pm$ 0.0380} & 
     {0.5591 \fontsize{15}{16}\selectfont$\pm$ 0.0291} & 
     {0.6622 \fontsize{15}{16}\selectfont$\pm$ 0.0742} & 
     {0.5566 \fontsize{15}{16}\selectfont$\pm$ 0.0310} & 
     \textbf{{0.5429 \fontsize{15}{16}\selectfont$\pm$ 0.0249}}\\
     & GAT & 
     {0.6087 \fontsize{15}{16}\selectfont$\pm$ 0.0507} &
     {0.5162 \fontsize{15}{16}\selectfont$\pm$ 0.0238} & 
     {0.5228 \fontsize{15}{16}\selectfont$\pm$ 0.0332} & 
     {0.5151 \fontsize{15}{16}\selectfont$\pm$ 0.0232} & 
     {0.5420 \fontsize{15}{16}\selectfont$\pm$ 0.0360} & 
     {0.5124 \fontsize{15}{16}\selectfont$\pm$ 0.0209} & 
     \textbf{{0.5040 \fontsize{15}{16}\selectfont$\pm$ 0.0201}}\\
     \hline
     
     \multirow{2}{*}{Citeseer} & GCN & 
     {0.9265 \fontsize{15}{16}\selectfont$\pm$ 0.1038} & 
     {0.8800 \fontsize{15}{16}\selectfont$\pm$ 0.0428} & 
     {0.8734 \fontsize{15}{16}\selectfont$\pm$ 0.0234} & 
     {0.8770 \fontsize{15}{16}\selectfont$\pm$ 0.0386} & 
     {0.9204 \fontsize{15}{16}\selectfont$\pm$ 0.0578} & 
     {0.8702 \fontsize{15}{16}\selectfont$\pm$ 0.0404} & 
     \textbf{{0.8599 \fontsize{15}{16}\selectfont$\pm$ 0.0419}}\\
     & GAT & 
     {0.9602 \fontsize{15}{16}\selectfont$\pm$ 0.1025} &
     {0.8762 \fontsize{15}{16}\selectfont$\pm$ 0.0330} & 
     {0.8729 \fontsize{15}{16}\selectfont$\pm$ 0.0254} & 
     {0.8752 \fontsize{15}{16}\selectfont$\pm$ 0.0324} & 
     {0.8752 \fontsize{15}{16}\selectfont$\pm$ 0.0291} & 
     {0.8715 \fontsize{15}{16}\selectfont$\pm$ 0.0290} & 
     \textbf{{0.8611 \fontsize{15}{16}\selectfont$\pm$ 0.0283}}\\
     \hline
     
     \multirow{2}{*}{Pubmed} & GCN & 
     {0.3939 \fontsize{15}{16}\selectfont$\pm$0.0160} & 
     {0.3676 \fontsize{15}{16}\selectfont$\pm$ 0.0072} & 
     {0.3679 \fontsize{15}{16}\selectfont$\pm$ 0.0073} & 
     {0.3659 \fontsize{15}{16}\selectfont$\pm$ 0.0073} & 
     \textbf{0.3582 \fontsize{15}{16}\selectfont$\pm$ 0.0073} & 
     {0.3638 \fontsize{15}{16}\selectfont$\pm$ 0.0069} & 
     {0.3627 \fontsize{15}{16}\selectfont$\pm$ 0.0067}\\
     & GAT & 
     {0.4382 \fontsize{15}{16}\selectfont$\pm$ 0.0120} &
     {0.3871 \fontsize{15}{16}\selectfont$\pm$ 0.0078} & 
     {0.3864 \fontsize{15}{16}\selectfont$\pm$ 0.0070} & 
     {0.3870 \fontsize{15}{16}\selectfont$\pm$ 0.0078} & 
     {0.3845 \fontsize{15}{16}\selectfont$\pm$ 0.0072} & 
     {0.3866 \fontsize{15}{16}\selectfont$\pm$ 0.0077} & 
     \textbf{0.3844 \fontsize{15}{16}\selectfont$\pm$ 0.0075}\\
     \hline
     
     \multirow{2}{*}{Computers} & GCN & 
     {0.4297 \fontsize{15}{16}\selectfont$\pm$ 0.0119} & 
     {0.4295 \fontsize{15}{16}\selectfont$\pm$ 0.0116} & 
     {0.4291 \fontsize{15}{16}\selectfont$\pm$ 0.0113} & 
     {0.4130 \fontsize{15}{16}\selectfont$\pm$ 0.0146} & 
     {0.4333 \fontsize{15}{16}\selectfont$\pm$ 0.0356} & 
     {0.4243 \fontsize{15}{16}\selectfont$\pm$ 0.0134} & 
     \textbf{{0.4080 \fontsize{15}{16}\selectfont$\pm$ 0.0104}}\\
     & GAT & 
     {0.3739 \fontsize{15}{16}\selectfont$\pm$ 0.0145} &
     {0.3734 \fontsize{15}{16}\selectfont$\pm$ 0.0142} & 
     {0.3725 \fontsize{15}{16}\selectfont$\pm$ 0.0132} & 
     {0.3687 \fontsize{15}{16}\selectfont$\pm$ 0.0148} & 
     {0.3961 \fontsize{15}{16}\selectfont$\pm$ 0.0284} & 
     {0.3730 \fontsize{15}{16}\selectfont$\pm$ 0.0145} & 
     \textbf{{0.3670 \fontsize{15}{16}\selectfont$\pm$ 0.0137}}\\
     \hline

     \multirow{2}{*}{Photo} & GCN & 
     {0.2877 \fontsize{15}{16}\selectfont$\pm$ 0.0108} & 
     {0.2892 \fontsize{15}{16}\selectfont$\pm$ 0.0110} & 
     {0.2913 \fontsize{15}{16}\selectfont$\pm$ 0.0122} & 
     \textbf{0.2725 \fontsize{15}{16}\selectfont$\pm$ 0.0133} & 
     {0.3717 \fontsize{15}{16}\selectfont$\pm$ 0.0737} & 
     {0.2867 \fontsize{15}{16}\selectfont$\pm$ 0.0113} & 
     {0.2750 \fontsize{15}{16}\selectfont$\pm$ 0.0117}\\
     & GAT & 
     {0.2712 \fontsize{15}{16}\selectfont$\pm$ 0.0205} &
     {0.2703 \fontsize{15}{16}\selectfont$\pm$ 0.0166} & 
     {0.2692 \fontsize{15}{16}\selectfont$\pm$ 0.0161} & 
     {0.2657 \fontsize{15}{16}\selectfont$\pm$ 0.0185} & 
     {0.3228 \fontsize{15}{16}\selectfont$\pm$ 0.0563} & 
     {0.2704 \fontsize{15}{16}\selectfont$\pm$ 0.0172} & 
     \textbf{{0.2638 \fontsize{15}{16}\selectfont$\pm$ 0.0156}}\\
     \hline

     \multirow{2}{*}{CS} & GCN & 
     {0.2196 \fontsize{15}{16}\selectfont$\pm$ 0.0119} & 
     {0.2142 \fontsize{15}{16}\selectfont$\pm$ 0.0056} & 
     {0.2162 \fontsize{15}{16}\selectfont$\pm$ 0.0049} & 
     {0.2141 \fontsize{15}{16}\selectfont$\pm$ 0.0055} & 
     {0.2778 \fontsize{15}{16}\selectfont$\pm$ 0.0583} & 
     {0.2132 \fontsize{15}{16}\selectfont$\pm$ 0.0057} & 
     \textbf{{0.2127 \fontsize{15}{16}\selectfont$\pm$ 0.0054}}\\
     & GAT & 
     {0.2451 \fontsize{15}{16}\selectfont$\pm$ 0.0084} &
     {0.2432 \fontsize{15}{16}\selectfont$\pm$ 0.0057} & 
     {0.2425 \fontsize{15}{16}\selectfont$\pm$ 0.0053} & 
     {0.2428 \fontsize{15}{16}\selectfont$\pm$ 0.0057} & 
     {0.2786 \fontsize{15}{16}\selectfont$\pm$ 0.0350} & 
     {0.2422 \fontsize{15}{16}\selectfont$\pm$ 0.0054} & 
     \textbf{{0.2416 \fontsize{15}{16}\selectfont$\pm$ 0.0051}}\\
     \hline

     \multirow{2}{*}{Physics} & GCN & 
     {0.1199 \fontsize{15}{16}\selectfont$\pm$ 0.0043} & 
     {0.1190 \fontsize{15}{16}\selectfont$\pm$ 0.0035} & 
     {0.1190 \fontsize{15}{16}\selectfont$\pm$ 0.0033} & 
     {0.1190 \fontsize{15}{16}\selectfont$\pm$ 0.0035} & 
     {0.1289 \fontsize{15}{16}\selectfont$\pm$ 0.0114} & 
     {0.1188 \fontsize{15}{16}\selectfont$\pm$ 0.0033} & 
     \textbf{{0.1185 \fontsize{15}{16}\selectfont$\pm$ 0.0034}}\\
     & GAT & 
     {0.1288 \fontsize{15}{16}\selectfont$\pm$ 0.0045} & 
     {0.1287 \fontsize{15}{16}\selectfont$\pm$ 0.0043} & 
     \textbf{0.1283 \fontsize{15}{16}\selectfont$\pm$ 0.0041} & 
     {0.1287 \fontsize{15}{16}\selectfont$\pm$ 0.0043} & 
     {0.1334 \fontsize{15}{16}\selectfont$\pm$ 0.0055} & 
     {0.1286 \fontsize{15}{16}\selectfont$\pm$ 0.0042} & 
     {0.1285 \fontsize{15}{16}\selectfont$\pm$ 0.0042}\\
     \hline

     \multirow{2}{*}{CoraFull} & GCN & 
     {1.4310 \fontsize{15}{16}\selectfont$\pm$ 0.0221} &
     {1.4270 \fontsize{15}{16}\selectfont$\pm$ 0.0185} & 
     {1.4300 \fontsize{15}{16}\selectfont$\pm$ 0.0199} & 
     {1.4210 \fontsize{15}{16}\selectfont$\pm$ 0.0182} & 
     {1.4780 \fontsize{15}{16}\selectfont$\pm$ 0.1769} & 
     \textbf{1.4010 \fontsize{15}{16}\selectfont$\pm$ 0.0189} & 
     {1.4070 \fontsize{15}{16}\selectfont$\pm$ 0.0120}\\
     & GAT & 
     {1.3670 \fontsize{15}{16}\selectfont$\pm$ 0.0217} &
     {1.3620 \fontsize{15}{16}\selectfont$\pm$ 0.0176} & 
     {1.3630 \fontsize{15}{16}\selectfont$\pm$ 0.0177} & 
     {1.3610 \fontsize{15}{16}\selectfont$\pm$ 0.0175} & 
     {1.4570 \fontsize{15}{16}\selectfont$\pm$ 0.1953} & 
     {1.3550 \fontsize{15}{16}\selectfont$\pm$ 0.0170} & 
     \textbf{{1.3490 \fontsize{15}{16}\selectfont$\pm$ 0.0174}}\\
     \hline
    \hline\hline
    \end{tabular}}}
    \label{Table_nll}
  \end{table*}
}

%% file: Table_brier.tex
{\renewcommand{\arraystretch}{1.2}
  \begin{table*}[t!]
    \caption{Brier score results (reported in percentage) for our proposed calibration method and baselines, averaged over 75 repetitions ($\pm$ STD). A lower value indicates better calibration performance.}
    \centering
    \resizebox{.98\textwidth}{!}{
    \huge{
    \begin{tabular}{ c | c | c | c | c | c | c | c | c }
     \hline\hline
     \multicolumn{2}{ c |}{\textbf{Methods}} & {{UnCal.}} & {{TS}} & {{VS}} & {{ETS}} & {{CaGCN}} & {{GATS}} & {{\textbf{Ours}}}\\
     \hline
     \hline
     \multirow{2}{*}{Cora} & GCN & 
     {0.2828 \fontsize{15}{16}\selectfont$\pm$ 0.0189} & 
     {0.2555 \fontsize{15}{16}\selectfont$\pm$ 0.0092} & 
     {0.2564 \fontsize{15}{16}\selectfont$\pm$ 0.0096} & 
     {0.2555 \fontsize{15}{16}\selectfont$\pm$ 0.0091} & 
     {0.2607 \fontsize{15}{16}\selectfont$\pm$ 0.0097} & 
     {0.2552 \fontsize{15}{16}\selectfont$\pm$ 0.0100} & 
     \textbf{{0.2541 \fontsize{15}{16}\selectfont$\pm$ 0.0086}}\\
     & GAT & 
     {0.2766 \fontsize{15}{16}\selectfont$\pm$ 0.0222} &
     {0.2416 \fontsize{15}{16}\selectfont$\pm$ 0.0084} & 
     {0.2419 \fontsize{15}{16}\selectfont$\pm$ 0.0105} & 
     {0.2416 \fontsize{15}{16}\selectfont$\pm$ 0.0083} & 
     {0.2462 \fontsize{15}{16}\selectfont$\pm$ 0.0086} & 
     {0.2412 \fontsize{15}{16}\selectfont$\pm$ 0.0080} & 
     \textbf{{0.2402 \fontsize{15}{16}\selectfont$\pm$ 0.0080}}\\
     \hline
     
     \multirow{2}{*}{Citeseer} & GCN & 
     {0.4377 \fontsize{15}{16}\selectfont$\pm$ 0.0494} & 
     {0.4094 \fontsize{15}{16}\selectfont$\pm$ 0.0097} & 
     {0.4104 \fontsize{15}{16}\selectfont$\pm$ 0.0099} & 
     {0.4092 \fontsize{15}{16}\selectfont$\pm$ 0.0099} & 
     {0.4157 \fontsize{15}{16}\selectfont$\pm$ 0.0129} & 
     {0.4082 \fontsize{15}{16}\selectfont$\pm$ 0.0097} & 
     \textbf{{0.4044 \fontsize{15}{16}\selectfont$\pm$ 0.0084}}\\
     & GAT & 
     {0.4517 \fontsize{15}{16}\selectfont$\pm$ 0.0508} &
     {0.4099 \fontsize{15}{16}\selectfont$\pm$ 0.0090} & 
     {0.4108 \fontsize{15}{16}\selectfont$\pm$ 0.0102} & 
     {0.4098 \fontsize{15}{16}\selectfont$\pm$ 0.0090} & 
     {0.4107 \fontsize{15}{16}\selectfont$\pm$ 0.0094} & 
     {0.4097 \fontsize{15}{16}\selectfont$\pm$ 0.0087} & 
     \textbf{{0.4063 \fontsize{15}{16}\selectfont$\pm$ 0.0078}}\\
     \hline
     
     \multirow{2}{*}{Pubmed} & GCN & 
     {0.2135 \fontsize{15}{16}\selectfont$\pm$ 0.0078} & 
     {0.2020 \fontsize{15}{16}\selectfont$\pm$ 0.0039} & 
     {0.2024 \fontsize{15}{16}\selectfont$\pm$ 0.0040} & 
     {0.2020 \fontsize{15}{16}\selectfont$\pm$ 0.0038} & 
     \textbf{0.2002 \fontsize{15}{16}\selectfont$\pm$ 0.0039} & 
     {0.2017 \fontsize{15}{16}\selectfont$\pm$ 0.0039} & 
     {0.2014 \fontsize{15}{16}\selectfont$\pm$ 0.0038}\\
     & GAT & 
     {0.2377 \fontsize{15}{16}\selectfont$\pm$ 0.0103} &
     {0.2181 \fontsize{15}{16}\selectfont$\pm$ 0.0042} & 
     {0.2178 \fontsize{15}{16}\selectfont$\pm$ 0.0040} & 
     {0.2181 \fontsize{15}{16}\selectfont$\pm$ 0.0042} & 
     {0.2172 \fontsize{15}{16}\selectfont$\pm$ 0.0042} & 
     {0.2180 \fontsize{15}{16}\selectfont$\pm$ 0.0042} & 
     \textbf{0.2168 \fontsize{15}{16}\selectfont$\pm$ 0.0042}\\
     \hline
     
     \multirow{2}{*}{Computers} & GCN & 
     {0.1856 \fontsize{15}{16}\selectfont$\pm$ 0.0083} & 
     {0.1850 \fontsize{15}{16}\selectfont$\pm$ 0.0073} & 
     {0.1842 \fontsize{15}{16}\selectfont$\pm$ 0.0069} & 
     {0.1850 \fontsize{15}{16}\selectfont$\pm$ 0.0073} & 
     \textbf{1.1812 \fontsize{15}{16}\selectfont$\pm$ 0.0074} & 
     {0.1841 \fontsize{15}{16}\selectfont$\pm$ 0.0070} & 
     {0.1814 \fontsize{15}{16}\selectfont$\pm$ 0.0068}\\
     & GAT & 
     {0.1709 \fontsize{15}{16}\selectfont$\pm$ 0.0083} &
     {0.1707 \fontsize{15}{16}\selectfont$\pm$ 0.0080} & 
     {0.1692 \fontsize{15}{16}\selectfont$\pm$ 0.0065} & 
     {0.1707 \fontsize{15}{16}\selectfont$\pm$ 0.0080} & 
     {0.1712 \fontsize{15}{16}\selectfont$\pm$ 0.0078} & 
     {0.1708 \fontsize{15}{16}\selectfont$\pm$ 0.0079} & 
     \textbf{{0.1691 \fontsize{15}{16}\selectfont$\pm$ 0.0077}}\\
     \hline

     \multirow{2}{*}{Photo} & GCN & 
     {0.1166 \fontsize{15}{16}\selectfont$\pm$ 0.0062} & 
     {0.1156 \fontsize{15}{16}\selectfont$\pm$ 0.0067} & 
     {0.1157 \fontsize{15}{16}\selectfont$\pm$ 0.0060} & 
     {0.1156 \fontsize{15}{16}\selectfont$\pm$ 0.0057} & 
     {0.1161 \fontsize{15}{16}\selectfont$\pm$ 0.0049} & 
     {0.1151 \fontsize{15}{16}\selectfont$\pm$ 0.0054} & 
     \textbf{{0.1141 \fontsize{15}{16}\selectfont$\pm$ 0.0050}}\\
     & GAT & 
     {0.1167 \fontsize{15}{16}\selectfont$\pm$ 0.0100} &
     {0.1155 \fontsize{15}{16}\selectfont$\pm$ 0.0079} & 
     {0.1143 \fontsize{15}{16}\selectfont$\pm$ 0.0068} & 
     {0.1156 \fontsize{15}{16}\selectfont$\pm$ 0.0080} & 
     {0.1166 \fontsize{15}{16}\selectfont$\pm$ 0.0072} & 
     {0.1156 \fontsize{15}{16}\selectfont$\pm$ 0.0079} & 
     \textbf{{0.1140 \fontsize{15}{16}\selectfont$\pm$ 0.0072}}\\
     \hline

     \multirow{2}{*}{CS} & GCN & 
     {0.1032 \fontsize{15}{16}\selectfont$\pm$ 0.0040} & 
     {0.1028 \fontsize{15}{16}\selectfont$\pm$ 0.0023} & 
     {0.1020 \fontsize{15}{16}\selectfont$\pm$ 0.0020} & 
     {0.1018 \fontsize{15}{16}\selectfont$\pm$ 0.0023} & 
     {0.1065 \fontsize{15}{16}\selectfont$\pm$ 0.0043} & 
     {0.1016 \fontsize{15}{16}\selectfont$\pm$ 0.0024} & 
     \textbf{{0.1014 \fontsize{15}{16}\selectfont$\pm$ 0.0023}}\\
     & GAT & 
     {0.1133 \fontsize{15}{16}\selectfont$\pm$ 0.0034} &
     {0.1126 \fontsize{15}{16}\selectfont$\pm$ 0.0025} & 
     \textbf{0.1122 \fontsize{15}{16}\selectfont$\pm$ 0.0023} & 
     {0.1126 \fontsize{15}{16}\selectfont$\pm$ 0.0025} & 
     {0.1152 \fontsize{15}{16}\selectfont$\pm$ 0.0037} & 
     {0.1126 \fontsize{15}{16}\selectfont$\pm$ 0.0024} & 
     {0.1123 \fontsize{15}{16}\selectfont$\pm$ 0.0025}\\
     \hline

     \multirow{2}{*}{Physics} & GCN & 
     {0.0614 \fontsize{15}{16}\selectfont$\pm$ 0.0020} & 
     {0.0614 \fontsize{15}{16}\selectfont$\pm$ 0.0019} & 
     {0.0614 \fontsize{15}{16}\selectfont$\pm$ 0.0018} & 
     {0.0614 \fontsize{15}{16}\selectfont$\pm$ 0.0019} & 
     {0.0625 \fontsize{15}{16}\selectfont$\pm$ 0.0022} & 
     {0.0613 \fontsize{15}{16}\selectfont$\pm$ 0.0019} & 
     \textbf{{0.0612 \fontsize{15}{16}\selectfont$\pm$ 0.0019}}\\
     & GAT & 
     {0.0657 \fontsize{15}{16}\selectfont$\pm$ 0.0018} & 
     {0.0657 \fontsize{15}{16}\selectfont$\pm$ 0.0018} & 
     \textbf{0.0656 \fontsize{15}{16}\selectfont$\pm$ 0.0018} & 
     {0.0657 \fontsize{15}{16}\selectfont$\pm$ 0.0018} & 
     {0.0665 \fontsize{15}{16}\selectfont$\pm$ 0.0018} & 
     \textbf{{0.0656 \fontsize{15}{16}\selectfont$\pm$ 0.0018}} & 
     {0.0657 \fontsize{15}{16}\selectfont$\pm$ 0.0018}\\
     \hline

     \multirow{2}{*}{CoraFull} & GCN & 
     {0.5231 \fontsize{15}{16}\selectfont$\pm$ 0.0074} &
     {0.5208 \fontsize{15}{16}\selectfont$\pm$ 0.0052} & 
     {0.5201 \fontsize{15}{16}\selectfont$\pm$ 0.0050} & 
     {0.5207 \fontsize{15}{16}\selectfont$\pm$ 0.0052} & 
     {0.5221 \fontsize{15}{16}\selectfont$\pm$ 0.0138} & 
     \textbf{0.5159 \fontsize{15}{16}\selectfont$\pm$ 0.0054} & 
     {0.5176 \fontsize{15}{16}\selectfont$\pm$ 0.0054}\\
     & GAT & 
     {0.5117 \fontsize{15}{16}\selectfont$\pm$ 0.0072} &
     {0.5099 \fontsize{15}{16}\selectfont$\pm$ 0.0057} & 
     \textbf{0.5080 \fontsize{15}{16}\selectfont$\pm$ 0.0057} & 
     {0.5098 \fontsize{15}{16}\selectfont$\pm$ 0.0057} & 
     {0.5178 \fontsize{15}{16}\selectfont$\pm$ 0.0162} & 
     {0.5089 \fontsize{15}{16}\selectfont$\pm$ 0.0057} & 
     \textbf{{0.5080 \fontsize{15}{16}\selectfont$\pm$ 0.0057}}\\
     \hline
    \hline\hline
    \end{tabular}}}
    \label{Table_brier}
    \vskip -10pt
  \end{table*}
}

%% file: qualitative_analysis_appendix.tex
\begin{figure*}[t!]
    \centering
    \includegraphics[width=1\linewidth]{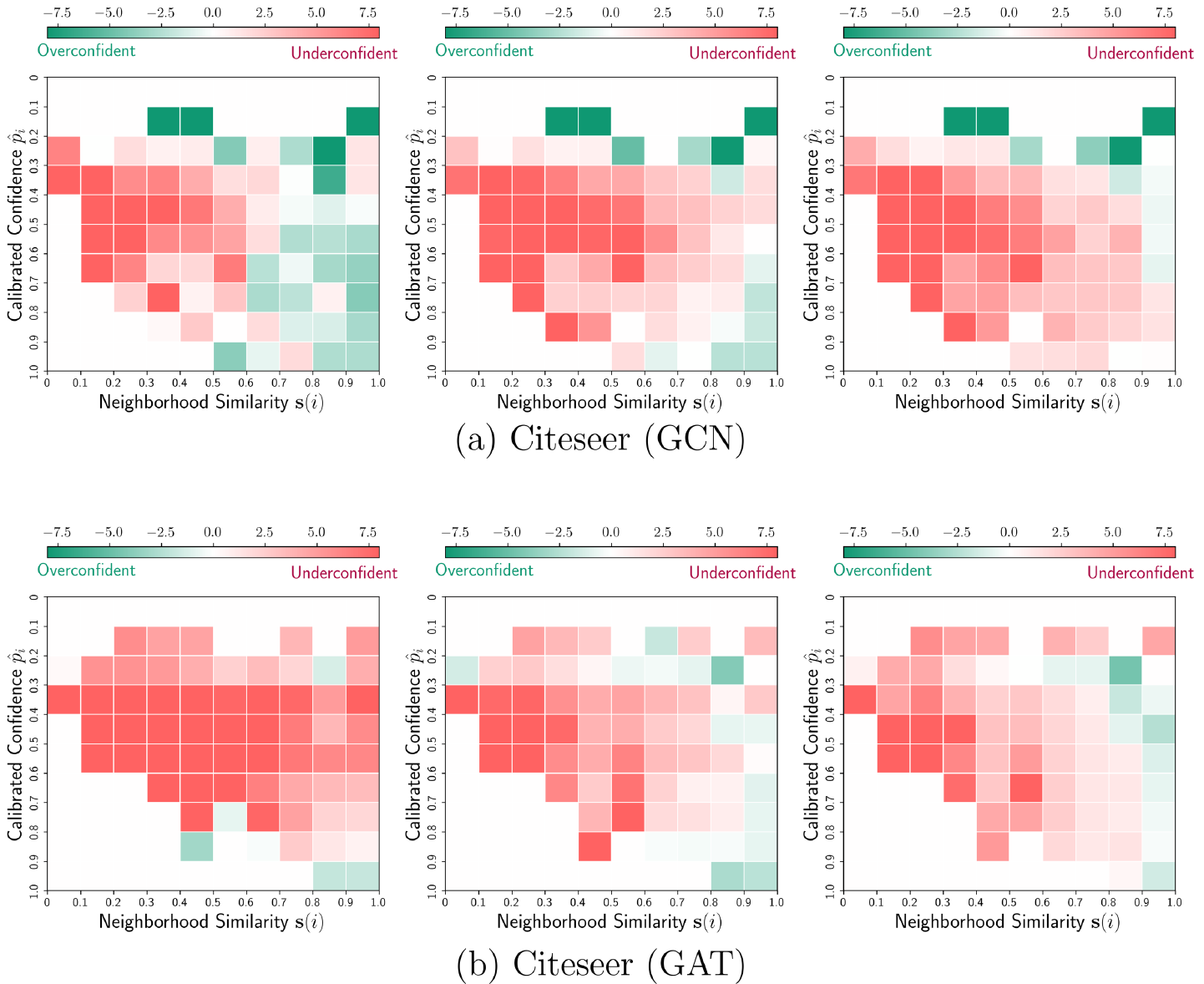}
    \caption{Qualitative analysis of our calibration results  (right) on the Citeseer dataset, compared with CaGCN (left) and GATS (center). Each cell in the heatmap represents the subtraction of the average confidence of calibrated nodes from the accuracy, with color and intensity indicating the magnitude of this discrepancy.
    }
    \label{Fig: citeseer_result_plot}
    \vskip -5pt
\end{figure*}
\begin{figure*}[h!]
    \centering
    \includegraphics[width=1\linewidth]{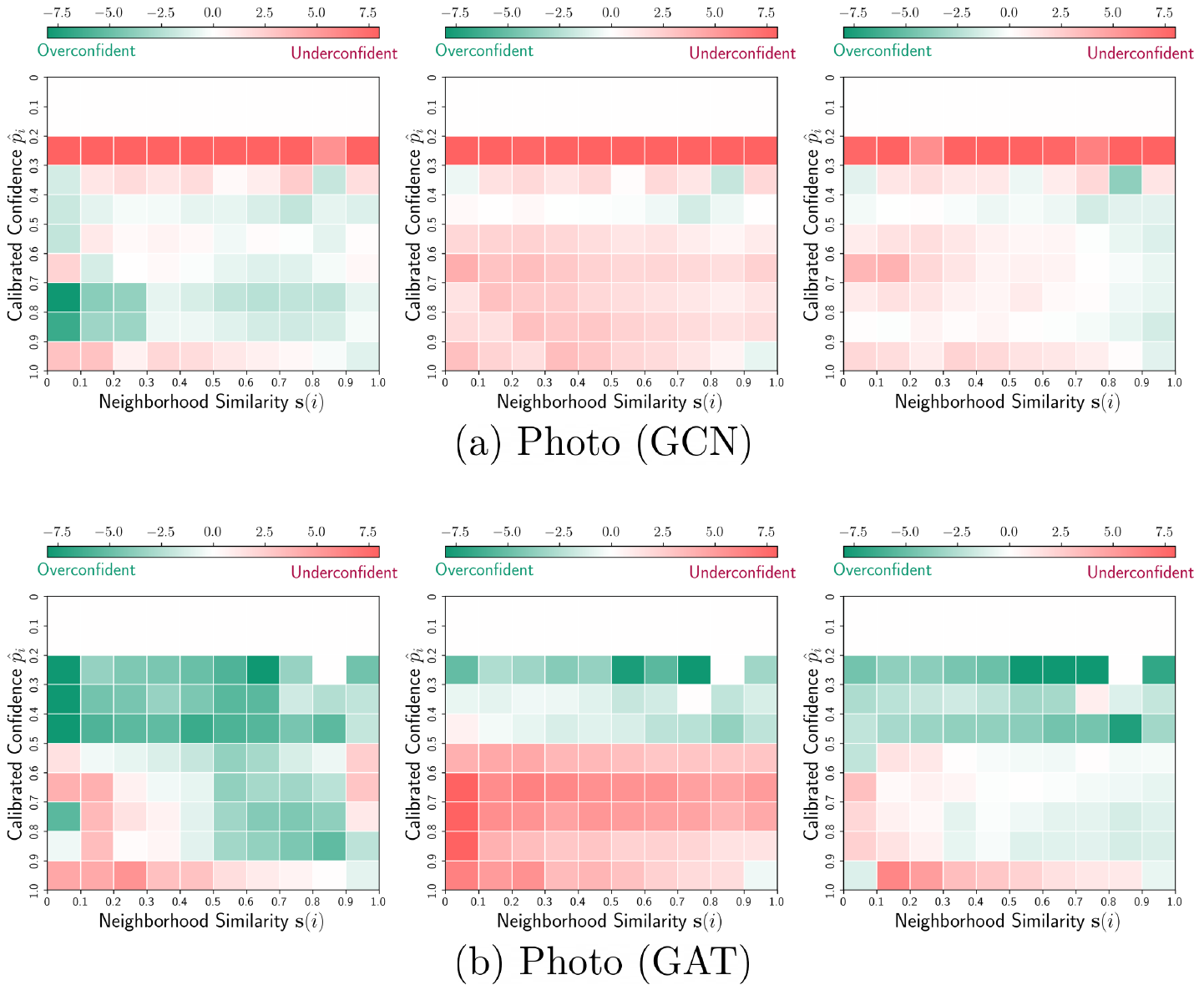}
    \caption{Qualitative analysis of our calibration results  (right) on the Photo dataset, compared with CaGCN (left) and GATS (center). Each cell in the heatmap represents the subtraction of the average confidence of calibrated nodes from the accuracy, with color and intensity indicating the magnitude of this discrepancy.
    }
    \label{Fig: computers_result_plot}
    \vskip -5pt
\end{figure*}
\begin{figure*}[h!]
    \centering
    \includegraphics[width=1\linewidth]{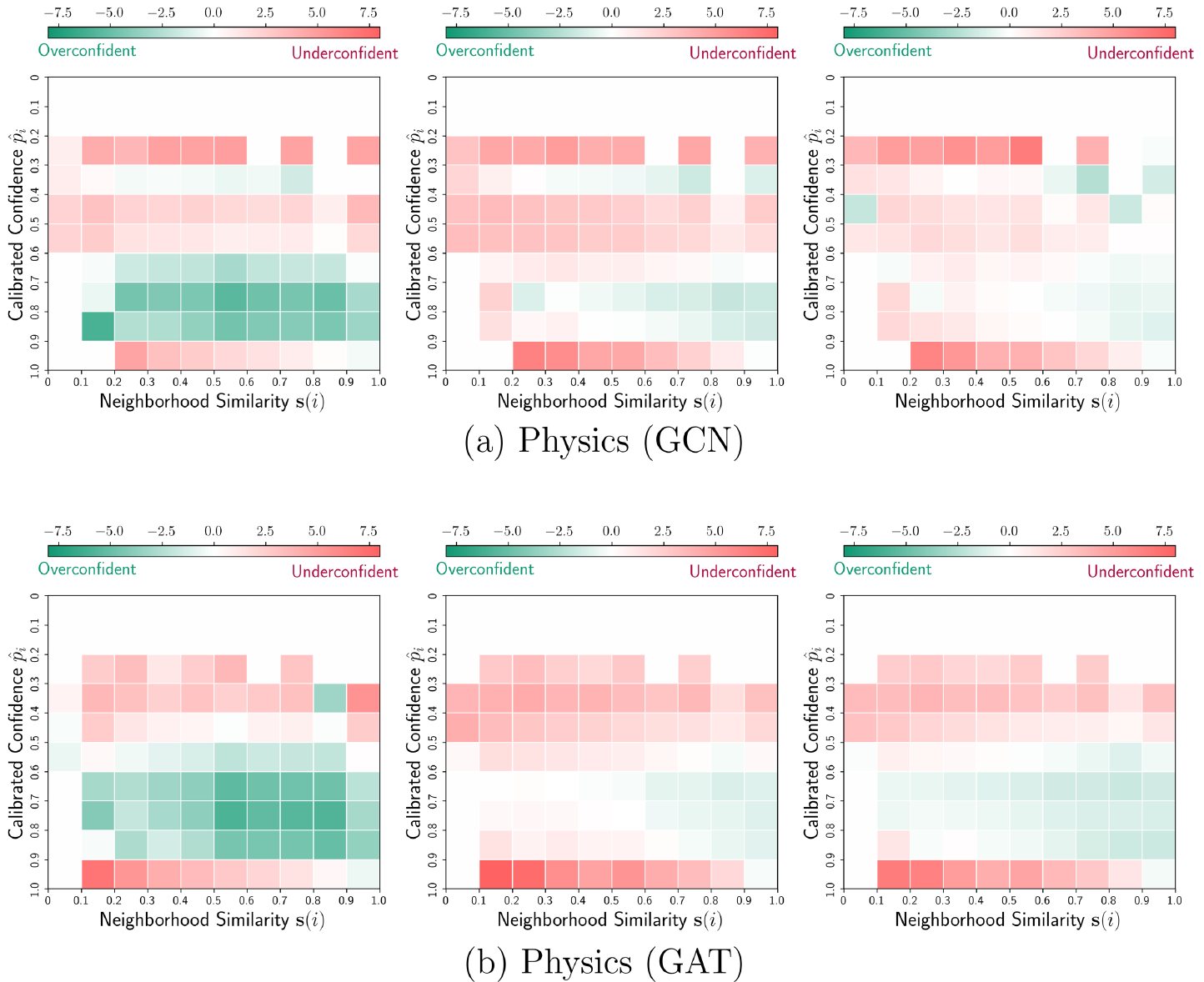}
    \caption{Qualitative analysis of our calibration results  (right) on the Physics dataset, compared with CaGCN (left) and GATS (center). Each cell in the heatmap represents the subtraction of the average confidence of calibrated nodes from the accuracy, with color and intensity indicating the magnitude of this discrepancy.
    }
    \label{Fig: physics_result_plot}
    \vskip -5pt
\end{figure*}
\begin{figure*}[h!]
    \centering
    \includegraphics[width=1\linewidth]{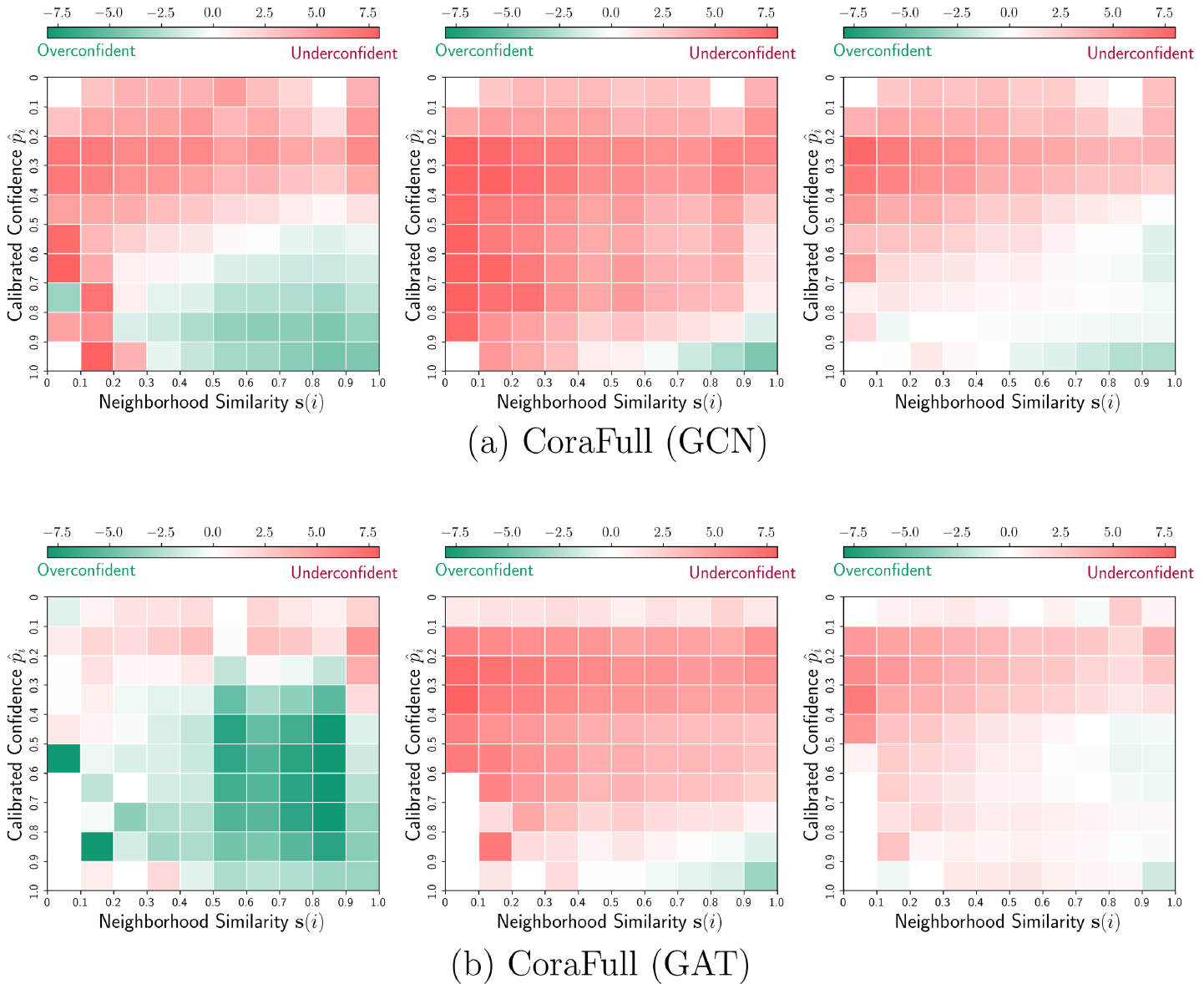}
    \caption{Qualitative analysis of our calibration results  (right) on the CoraFull dataset, compared with CaGCN (left) and GATS (center). Each cell in the heatmap represents the subtraction of the average confidence of calibrated nodes from the accuracy, with color and intensity indicating the magnitude of this discrepancy.
    }
    \label{Fig: corafull_result_plot}
    \vskip -5pt
\end{figure*}

%% file: obs_plot_appendix.tex
\begin{figure*}[h]
    \centering
    \includegraphics[width=1\linewidth]{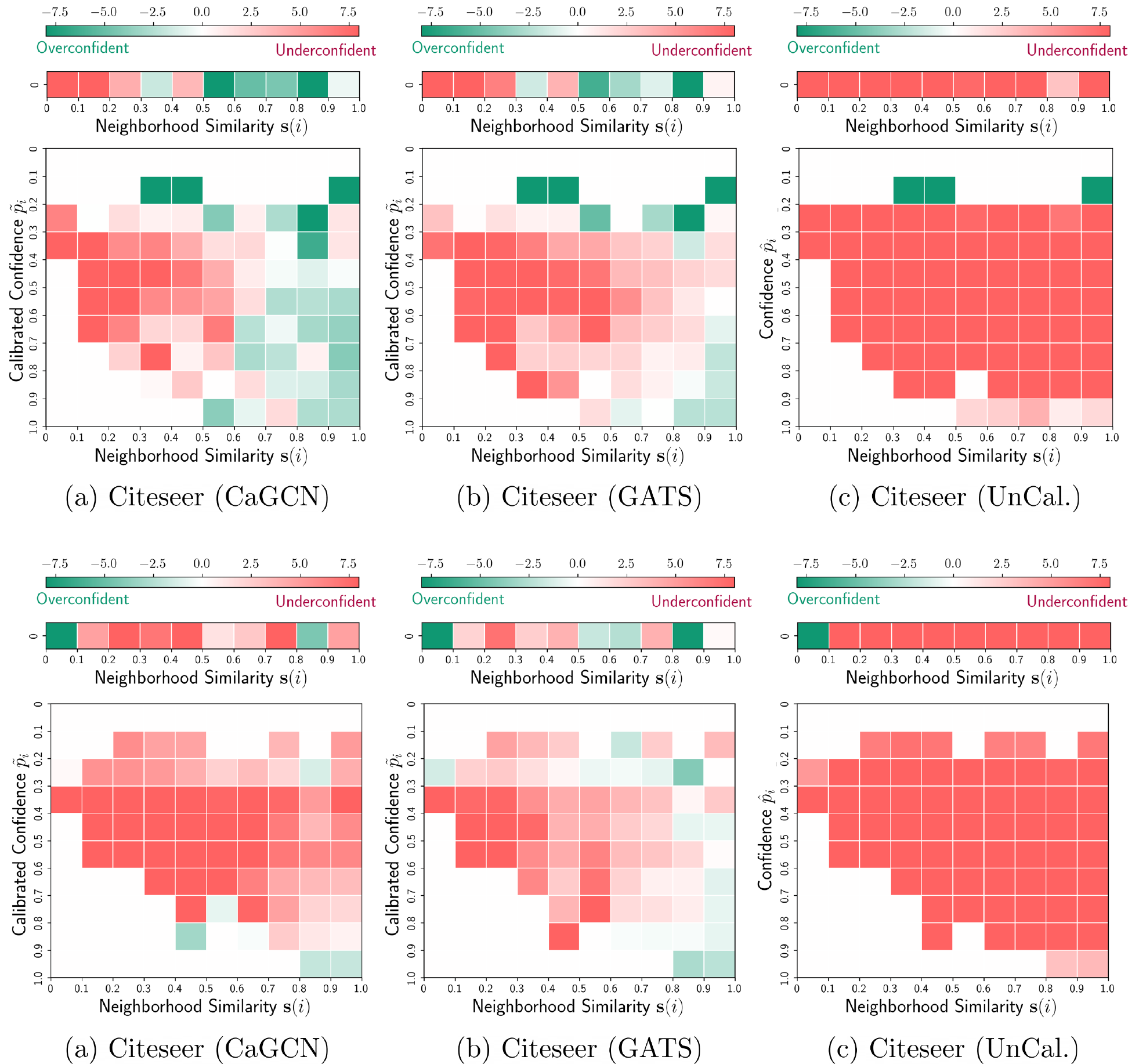}
    \caption{Investigation results comparing uncalibrated and calibrated logits on the Citeseer dataset using CaGCN and GATS, with GCN in the upper plot and GAT in the lower plot. 
    }
    \label{Fig: citeseer_obs_plot}
    \vskip -10pt
\end{figure*}

\begin{figure*}[h]
    \centering
    \includegraphics[width=1\linewidth]{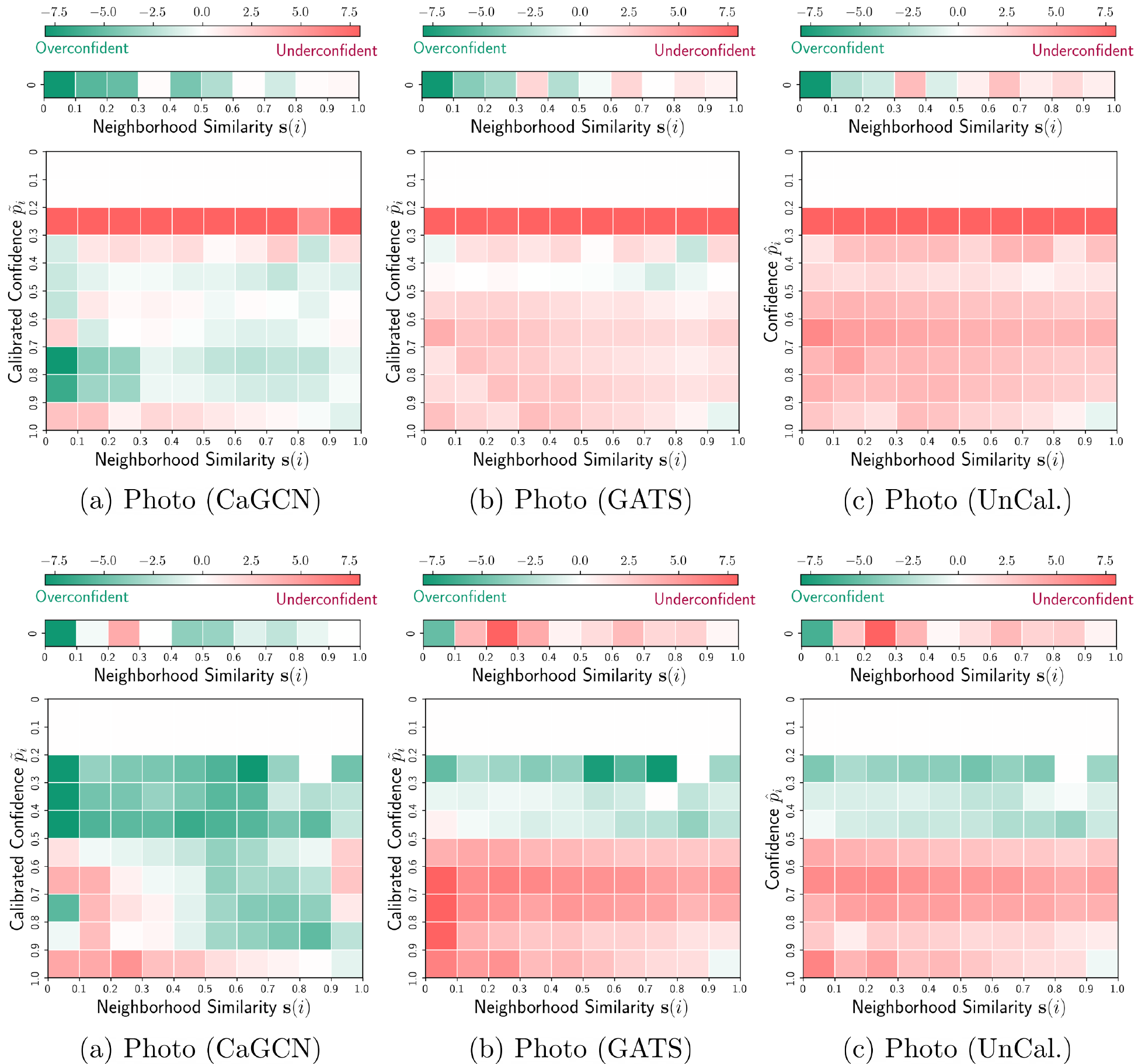}
    \caption{Investigation results comparing uncalibrated and calibrated logits on the Photo dataset using CaGCN and GATS, with GCN in the upper plot and GAT in the lower plot.
    }
    \label{Fig: computers_obs_plot}
    \vskip -10pt
\end{figure*}

\begin{figure*}[h]
    \centering
    \includegraphics[width=1\linewidth]{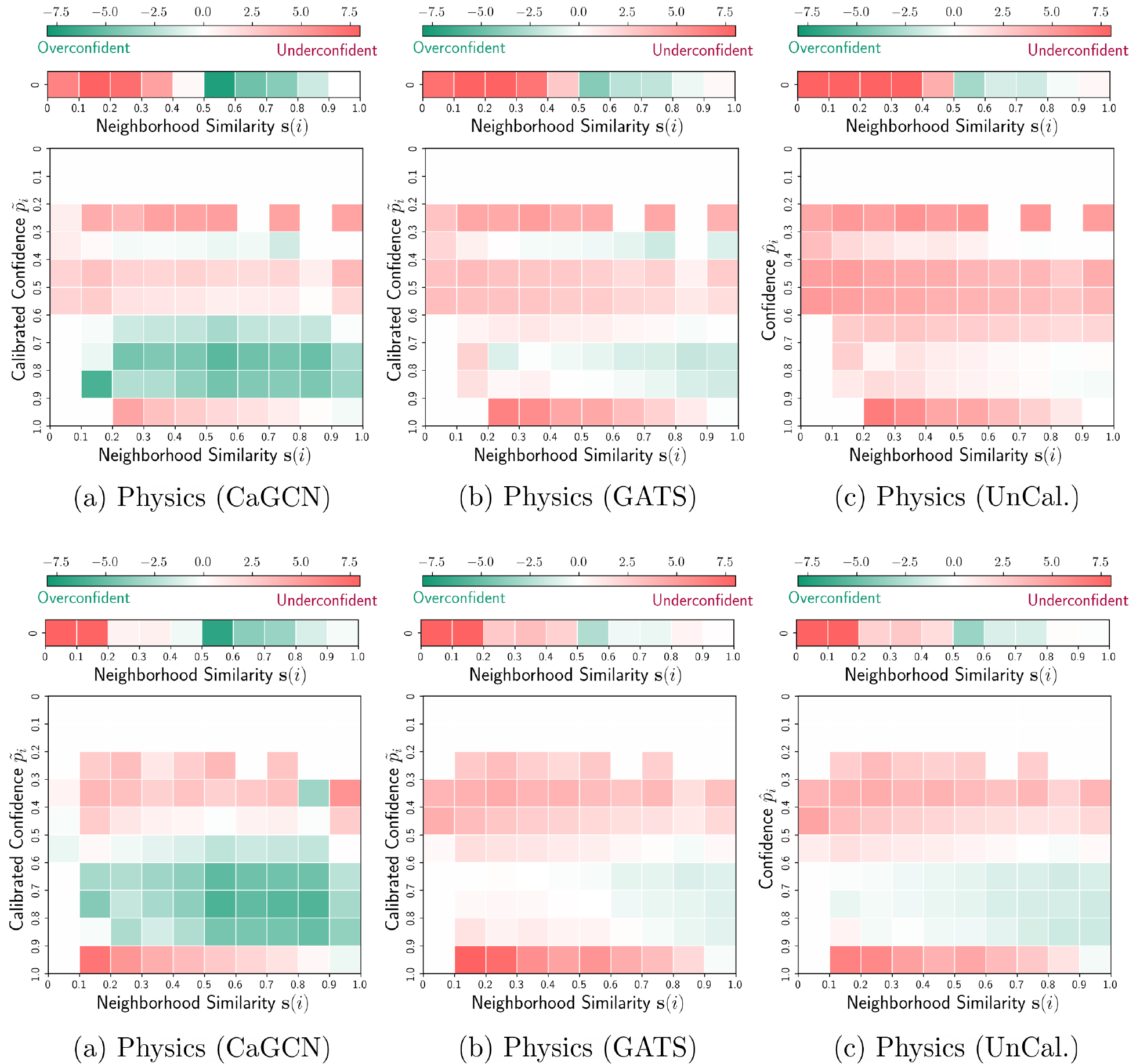}
    \caption{Investigation results comparing uncalibrated and calibrated logits on the Physics dataset using CaGCN and GATS, with GCN in the upper plot and GAT in the lower plot.
    }
    \label{Fig: physics_obs_plot}
    \vskip -10pt
\end{figure*}

\begin{figure*}[h]
    \centering
    \includegraphics[width=1\linewidth]{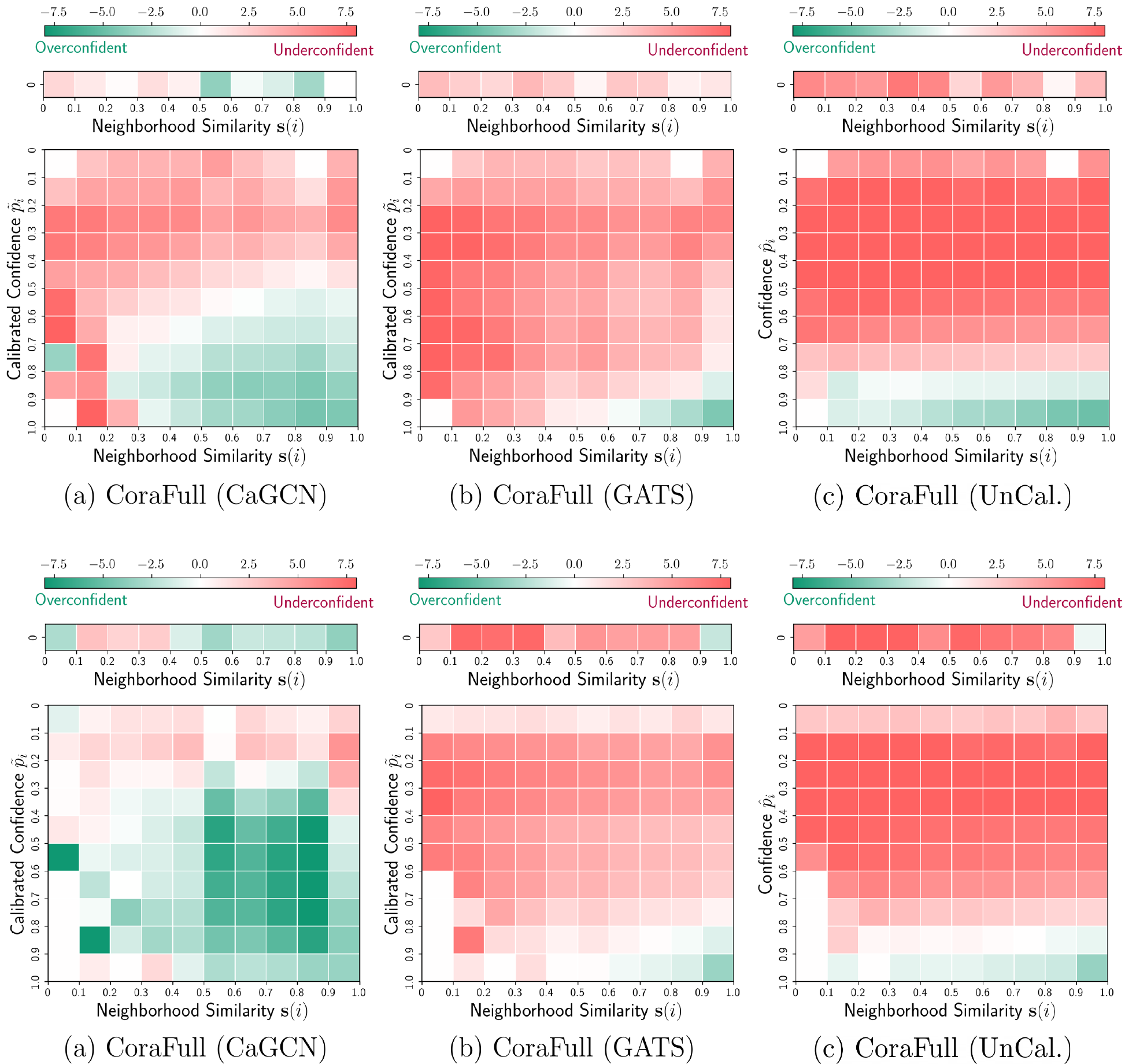}
    \caption{Investigation results comparing uncalibrated and calibrated logits on the CoraFull dataset using CaGCN and GATS, with GCN in the upper plot and GAT in the lower plot.
    }
    \label{Fig: corafull_obs_plot}
    \vskip -10pt
\end{figure*}

%% file: main.bbl
\begin{thebibliography}{76}
\providecommand{\natexlab}[1]{#1}

\bibitem[{Ahn and Kim(2021)}]{ahn2021variational}
Ahn, S.~J.; and Kim, M. 2021.
\newblock Variational graph normalized autoencoders.
\newblock In \emph{Proceedings of the 30th ACM international conference on information \& knowledge management}, 2827--2831.

\bibitem[{Angelopoulos et~al.(2020)Angelopoulos, Bates, Malik, and Jordan}]{angelopoulos2020uncertainty}
Angelopoulos, A.; Bates, S.; Malik, J.; and Jordan, M.~I. 2020.
\newblock Uncertainty sets for image classifiers using conformal prediction.
\newblock \emph{arXiv preprint arXiv:2009.14193}.

\bibitem[{Bober et~al.(2023)Bober, Monod, Saucan, and Webster}]{bober2023rewiring}
Bober, J.; Monod, A.; Saucan, E.; and Webster, K.~N. 2023.
\newblock Rewiring networks for graph neural network training using discrete geometry.
\newblock In \emph{International Conference on Complex Networks and Their Applications}, 225--236. Springer.

\bibitem[{Bojchevski and G{\"u}nnemann(2017)}]{bojchevski2017deep}
Bojchevski, A.; and G{\"u}nnemann, S. 2017.
\newblock Deep gaussian embedding of graphs: Unsupervised inductive learning via ranking.
\newblock \emph{arXiv preprint arXiv:1707.03815}.

\bibitem[{Brier et~al.(1950)}]{brier1950verification}
Brier, G.~W.; et~al. 1950.
\newblock Verification of forecasts expressed in terms of probability.
\newblock \emph{Monthly weather review}, 78(1): 1--3.

\bibitem[{Cauchois, Gupta, and Duchi(2021)}]{cauchois2021knowing}
Cauchois, M.; Gupta, S.; and Duchi, J.~C. 2021.
\newblock Knowing what you know: valid and validated confidence sets in multiclass and multilabel prediction.
\newblock \emph{Journal of machine learning research}, 22(81): 1--42.

\bibitem[{Chen et~al.(2024)Chen, Mao, Li, Jin, Wen, Wei, Wang, Yin, Fan, Liu et~al.}]{chen2024exploring}
Chen, Z.; Mao, H.; Li, H.; Jin, W.; Wen, H.; Wei, X.; Wang, S.; Yin, D.; Fan, W.; Liu, H.; et~al. 2024.
\newblock Exploring the potential of large language models (llms) in learning on graphs.
\newblock \emph{ACM SIGKDD Explorations Newsletter}, 25(2): 42--61.

\bibitem[{de~Haan(1999)}]{de1999use}
de~Haan, P. 1999.
\newblock On the use of density kernels for concentration estimations within particle and puff dispersion models.
\newblock \emph{Atmospheric Environment}, 33(13): 2007--2021.

\bibitem[{Depeweg et~al.(2018)Depeweg, Hernandez-Lobato, Doshi-Velez, and Udluft}]{depeweg2018decomposition}
Depeweg, S.; Hernandez-Lobato, J.-M.; Doshi-Velez, F.; and Udluft, S. 2018.
\newblock Decomposition of uncertainty in Bayesian deep learning for efficient and risk-sensitive learning.
\newblock In \emph{International conference on machine learning}, 1184--1193. PMLR.

\bibitem[{Dusenberry et~al.(2020)Dusenberry, Jerfel, Wen, Ma, Snoek, Heller, Lakshminarayanan, and Tran}]{dusenberry2020efficient}
Dusenberry, M.; Jerfel, G.; Wen, Y.; Ma, Y.; Snoek, J.; Heller, K.; Lakshminarayanan, B.; and Tran, D. 2020.
\newblock Efficient and scalable bayesian neural nets with rank-1 factors.
\newblock In \emph{International conference on machine learning}, 2782--2792. PMLR.

\bibitem[{Elinas, Bonilla, and Tiao(2020)}]{elinas2020variational}
Elinas, P.; Bonilla, E.~V.; and Tiao, L. 2020.
\newblock Variational inference for graph convolutional networks in the absence of graph data and adversarial settings.
\newblock \emph{Advances in neural information processing systems}, 33: 18648--18660.

\bibitem[{Fey and Lenssen(2019)}]{fey2019fast}
Fey, M.; and Lenssen, J.~E. 2019.
\newblock Fast graph representation learning with PyTorch Geometric.
\newblock \emph{arXiv preprint arXiv:1903.02428}.

\bibitem[{Gal and Ghahramani(2016)}]{gal2016dropout}
Gal, Y.; and Ghahramani, Z. 2016.
\newblock Dropout as a bayesian approximation: Representing model uncertainty in deep learning.
\newblock In \emph{international conference on machine learning}, 1050--1059. PMLR.

\bibitem[{Guo et~al.(2017)Guo, Pleiss, Sun, and Weinberger}]{guo2017calibration}
Guo, C.; Pleiss, G.; Sun, Y.; and Weinberger, K.~Q. 2017.
\newblock On calibration of modern neural networks.
\newblock In \emph{International conference on machine learning}, 1321--1330. PMLR.

\bibitem[{Hamilton, Ying, and Leskovec(2017)}]{hamilton2017inductive}
Hamilton, W.; Ying, Z.; and Leskovec, J. 2017.
\newblock Inductive representation learning on large graphs.
\newblock \emph{Advances in neural information processing systems}, 30.

\bibitem[{Hasanzadeh et~al.(2020)Hasanzadeh, Hajiramezanali, Boluki, Zhou, Duffield, Narayanan, and Qian}]{hasanzadeh2020bayesian}
Hasanzadeh, A.; Hajiramezanali, E.; Boluki, S.; Zhou, M.; Duffield, N.; Narayanan, K.; and Qian, X. 2020.
\newblock Bayesian graph neural networks with adaptive connection sampling.
\newblock In \emph{International conference on machine learning}, 4094--4104. PMLR.

\bibitem[{H{\'e}bert-Johnson et~al.(2018)H{\'e}bert-Johnson, Kim, Reingold, and Rothblum}]{hebert2018multicalibration}
H{\'e}bert-Johnson, U.; Kim, M.; Reingold, O.; and Rothblum, G. 2018.
\newblock Multicalibration: Calibration for the (computationally-identifiable) masses.
\newblock In \emph{International Conference on Machine Learning}, 1939--1948. PMLR.

\bibitem[{Hou et~al.(2022)Hou, Liu, Dong, Wang, Tang et~al.}]{hou2022graphmae}
Hou, Z.; Liu, X.; Dong, Y.; Wang, C.; Tang, J.; et~al. 2022.
\newblock Graphmae: Self-supervised masked graph autoencoders.
\newblock \emph{arXiv preprint arXiv:2205.10803}.

\bibitem[{Hsu, Shen, and Cremers(2022)}]{hsu2022graph}
Hsu, H. H.-H.; Shen, Y.; and Cremers, D. 2022.
\newblock A Graph Is More Than Its Nodes: Towards Structured Uncertainty-Aware Learning on Graphs.
\newblock In \emph{NeurIPS 2022 Workshop: New Frontiers in Graph Learning}.

\bibitem[{Hsu et~al.(2022)Hsu, Shen, Tomani, and Cremers}]{hsu2022makes}
Hsu, H. H.-H.; Shen, Y.; Tomani, C.; and Cremers, D. 2022.
\newblock What Makes Graph Neural Networks Miscalibrated?
\newblock \emph{Advances in Neural Information Processing Systems}, 35: 13775--13786.

\bibitem[{Hu et~al.(2020)Hu, Fey, Zitnik, Dong, Ren, Liu, Catasta, and Leskovec}]{hu2020open}
Hu, W.; Fey, M.; Zitnik, M.; Dong, Y.; Ren, H.; Liu, B.; Catasta, M.; and Leskovec, J. 2020.
\newblock Open graph benchmark: Datasets for machine learning on graphs.
\newblock \emph{Advances in neural information processing systems}, 33: 22118--22133.

\bibitem[{Huang et~al.(2024)Huang, Jin, Candes, and Leskovec}]{huang2024uncertainty}
Huang, K.; Jin, Y.; Candes, E.; and Leskovec, J. 2024.
\newblock Uncertainty quantification over graph with conformalized graph neural networks.
\newblock \emph{Advances in Neural Information Processing Systems}, 36.

\bibitem[{Jiang et~al.(2021)Jiang, Araki, Ding, and Neubig}]{jiang2021can}
Jiang, Z.; Araki, J.; Ding, H.; and Neubig, G. 2021.
\newblock How can we know when language models know? on the calibration of language models for question answering.
\newblock \emph{Transactions of the Association for Computational Linguistics}, 9: 962--977.

\bibitem[{Jung et~al.(2023)Jung, Seo, Jeong, and Choi}]{jung2023scaling}
Jung, S.; Seo, S.; Jeong, Y.; and Choi, J. 2023.
\newblock Scaling of class-wise training losses for post-hoc calibration.
\newblock In \emph{International Conference on Machine Learning}, 15421--15434. PMLR.

\bibitem[{Kipf and Welling(2016)}]{kipf2016semi}
Kipf, T.~N.; and Welling, M. 2016.
\newblock Semi-supervised classification with graph convolutional networks.
\newblock \emph{arXiv preprint arXiv:1609.02907}.

\bibitem[{Kull et~al.(2019)Kull, Perello~Nieto, K{\"a}ngsepp, Silva~Filho, Song, and Flach}]{kull2019beyond}
Kull, M.; Perello~Nieto, M.; K{\"a}ngsepp, M.; Silva~Filho, T.; Song, H.; and Flach, P. 2019.
\newblock Beyond temperature scaling: Obtaining well-calibrated multi-class probabilities with dirichlet calibration.
\newblock \emph{Advances in neural information processing systems}, 32.

\bibitem[{Lakshminarayanan, Pritzel, and Blundell(2017)}]{lakshminarayanan2017simple}
Lakshminarayanan, B.; Pritzel, A.; and Blundell, C. 2017.
\newblock Simple and scalable predictive uncertainty estimation using deep ensembles.
\newblock \emph{Advances in neural information processing systems}, 30.

\bibitem[{Lee, Rossi, and Kong(2018)}]{lee2018graph}
Lee, J.~B.; Rossi, R.; and Kong, X. 2018.
\newblock Graph classification using structural attention.
\newblock In \emph{Proceedings of the 24th ACM SIGKDD International Conference on Knowledge Discovery \& Data Mining}, 1666--1674.

\bibitem[{Liu et~al.(2022)Liu, Liu, Hildebrandt, Joblin, Li, and Tresp}]{liu2022calibration}
Liu, T.; Liu, Y.; Hildebrandt, M.; Joblin, M.; Li, H.; and Tresp, V. 2022.
\newblock On Calibration of Graph Neural Networks for Node Classification.
\newblock In \emph{2022 International Joint Conference on Neural Networks (IJCNN)}, 1--8. IEEE.

\bibitem[{Luan et~al.(2022)Luan, Hua, Lu, Zhu, Zhao, Zhang, Chang, and Precup}]{luan2022revisiting}
Luan, S.; Hua, C.; Lu, Q.; Zhu, J.; Zhao, M.; Zhang, S.; Chang, X.-W.; and Precup, D. 2022.
\newblock Revisiting heterophily for graph neural networks.
\newblock \emph{Advances in neural information processing systems}, 35: 1362--1375.

\bibitem[{Ma and Blaschko(2021)}]{ma2021meta}
Ma, X.; and Blaschko, M.~B. 2021.
\newblock Meta-cal: Well-controlled post-hoc calibration by ranking.
\newblock In \emph{International Conference on Machine Learning}, 7235--7245. PMLR.

\bibitem[{Ma et~al.(2021)Ma, Liu, Shah, and Tang}]{ma2021homophily}
Ma, Y.; Liu, X.; Shah, N.; and Tang, J. 2021.
\newblock Is homophily a necessity for graph neural networks?
\newblock \emph{arXiv preprint arXiv:2106.06134}.

\bibitem[{Maddox et~al.(2019)Maddox, Izmailov, Garipov, Vetrov, and Wilson}]{maddox2019simple}
Maddox, W.~J.; Izmailov, P.; Garipov, T.; Vetrov, D.~P.; and Wilson, A.~G. 2019.
\newblock A simple baseline for bayesian uncertainty in deep learning.
\newblock \emph{Advances in neural information processing systems}, 32.

\bibitem[{Mao et~al.(2024)Mao, Chen, Jin, Han, Ma, Zhao, Shah, and Tang}]{mao2024demystifying}
Mao, H.; Chen, Z.; Jin, W.; Han, H.; Ma, Y.; Zhao, T.; Shah, N.; and Tang, J. 2024.
\newblock Demystifying Structural Disparity in Graph Neural Networks: Can One Size Fit All?
\newblock \emph{Advances in Neural Information Processing Systems}, 36.

\bibitem[{Minderer et~al.(2021)Minderer, Djolonga, Romijnders, Hubis, Zhai, Houlsby, Tran, and Lucic}]{minderer2021revisiting}
Minderer, M.; Djolonga, J.; Romijnders, R.; Hubis, F.; Zhai, X.; Houlsby, N.; Tran, D.; and Lucic, M. 2021.
\newblock Revisiting the calibration of modern neural networks.
\newblock \emph{Advances in Neural Information Processing Systems}, 34: 15682--15694.

\bibitem[{Mukhoti et~al.(2020)Mukhoti, Kulharia, Sanyal, Golodetz, Torr, and Dokania}]{mukhoti2020calibrating}
Mukhoti, J.; Kulharia, V.; Sanyal, A.; Golodetz, S.; Torr, P.; and Dokania, P. 2020.
\newblock Calibrating deep neural networks using focal loss.
\newblock \emph{Advances in Neural Information Processing Systems}, 33: 15288--15299.

\bibitem[{Naeini, Cooper, and Hauskrecht(2015)}]{naeini2015obtaining}
Naeini, M.~P.; Cooper, G.; and Hauskrecht, M. 2015.
\newblock Obtaining well calibrated probabilities using bayesian binning.
\newblock In \emph{Proceedings of the AAAI conference on artificial intelligence}, volume~29.

\bibitem[{Nguyen et~al.(2023)Nguyen, Hieu, Nguyen, Ho, Osher, and Nguyen}]{nguyen2023revisiting}
Nguyen, K.; Hieu, N.~M.; Nguyen, V.~D.; Ho, N.; Osher, S.; and Nguyen, T.~M. 2023.
\newblock Revisiting over-smoothing and over-squashing using ollivier-ricci curvature.
\newblock In \emph{International Conference on Machine Learning}, 25956--25979. PMLR.

\bibitem[{Nixon et~al.(2019)Nixon, Dusenberry, Zhang, Jerfel, and Tran}]{nixon2019measuring}
Nixon, J.; Dusenberry, M.~W.; Zhang, L.; Jerfel, G.; and Tran, D. 2019.
\newblock Measuring Calibration in Deep Learning.
\newblock In \emph{CVPR workshops}, volume~2.

\bibitem[{Pal, Regol, and Coates(2019)}]{pal2019bayesian}
Pal, S.; Regol, F.; and Coates, M. 2019.
\newblock Bayesian graph convolutional neural networks using non-parametric graph learning.
\newblock \emph{arXiv preprint arXiv:1910.12132}.

\bibitem[{Park, Song, and Yang(2021)}]{park2021graphens}
Park, J.; Song, J.; and Yang, E. 2021.
\newblock Graphens: Neighbor-aware ego network synthesis for class-imbalanced node classification.
\newblock In \emph{International Conference on Learning Representations}.

\bibitem[{Paszke et~al.(2019)Paszke, Gross, Massa, Lerer, Bradbury, Chanan, Killeen, Lin, Gimelshein, Antiga, Desmaison, Kopf, Yang, DeVito, Raison, Tejani, Chilamkurthy, Steiner, Fang, Bai, and Chintala}]{NEURIPS2019_pytorch}
Paszke, A.; Gross, S.; Massa, F.; Lerer, A.; Bradbury, J.; Chanan, G.; Killeen, T.; Lin, Z.; Gimelshein, N.; Antiga, L.; Desmaison, A.; Kopf, A.; Yang, E.; DeVito, Z.; Raison, M.; Tejani, A.; Chilamkurthy, S.; Steiner, B.; Fang, L.; Bai, J.; and Chintala, S. 2019.
\newblock PyTorch: An Imperative Style, High-Performance Deep Learning Library.
\newblock In \emph{Advances in Neural Information Processing Systems 32}, 8024--8035.

\bibitem[{Pei et~al.(2020)Pei, Wei, Chang, Lei, and Yang}]{pei2020geom}
Pei, H.; Wei, B.; Chang, K. C.-C.; Lei, Y.; and Yang, B. 2020.
\newblock Geom-gcn: Geometric graph convolutional networks.
\newblock \emph{arXiv preprint arXiv:2002.05287}.

\bibitem[{Perez-Lebel, Morvan, and Varoquaux(2022)}]{perez2022beyond}
Perez-Lebel, A.; Morvan, M.~L.; and Varoquaux, G. 2022.
\newblock Beyond calibration: estimating the grouping loss of modern neural networks.
\newblock \emph{arXiv preprint arXiv:2210.16315}.

\bibitem[{Platt et~al.(1999)}]{platt1999probabilistic}
Platt, J.; et~al. 1999.
\newblock Probabilistic outputs for support vector machines and comparisons to regularized likelihood methods.
\newblock \emph{Advances in large margin classifiers}, 10(3): 61--74.

\bibitem[{Rizve et~al.(2021)Rizve, Duarte, Rawat, and Shah}]{rizve2021defense}
Rizve, M.~N.; Duarte, K.; Rawat, Y.~S.; and Shah, M. 2021.
\newblock In defense of pseudo-labeling: An uncertainty-aware pseudo-label selection framework for semi-supervised learning.
\newblock \emph{arXiv preprint arXiv:2101.06329}.

\bibitem[{Romano, Sesia, and Candes(2020)}]{romano2020classification}
Romano, Y.; Sesia, M.; and Candes, E. 2020.
\newblock Classification with valid and adaptive coverage.
\newblock \emph{Advances in Neural Information Processing Systems}, 33: 3581--3591.

\bibitem[{Rong et~al.(2019)Rong, Huang, Xu, and Huang}]{rong2019dropedge}
Rong, Y.; Huang, W.; Xu, T.; and Huang, J. 2019.
\newblock Dropedge: Towards deep graph convolutional networks on node classification.
\newblock \emph{arXiv preprint arXiv:1907.10903}.

\bibitem[{Scott(2015)}]{scott2015multivariate}
Scott, D.~W. 2015.
\newblock \emph{Multivariate density estimation: theory, practice, and visualization}.
\newblock John Wiley \& Sons.

\bibitem[{Sen et~al.(2008)Sen, Namata, Bilgic, Getoor, Galligher, and Eliassi-Rad}]{sen2008collective}
Sen, P.; Namata, G.; Bilgic, M.; Getoor, L.; Galligher, B.; and Eliassi-Rad, T. 2008.
\newblock Collective classification in network data.
\newblock \emph{AI magazine}, 29(3): 93--93.

\bibitem[{Shchur et~al.(2018)Shchur, Mumme, Bojchevski, and G{\"u}nnemann}]{shchur2018pitfalls}
Shchur, O.; Mumme, M.; Bojchevski, A.; and G{\"u}nnemann, S. 2018.
\newblock Pitfalls of graph neural network evaluation.
\newblock \emph{arXiv preprint arXiv:1811.05868}.

\bibitem[{Shi et~al.(2022)Shi, Chen, Qiao, Yang, Wang, and Yan}]{shi2022select}
Shi, S.; Chen, J.; Qiao, K.; Yang, S.; Wang, L.; and Yan, B. 2022.
\newblock Select and Calibrate the Low-confidence: Dual-Channel Consistency based Graph Convolutional Networks.
\newblock \emph{arXiv preprint arXiv:2205.03753}.

\bibitem[{Stadler et~al.(2021)Stadler, Charpentier, Geisler, Z{\"u}gner, and G{\"u}nnemann}]{stadler2021graph}
Stadler, M.; Charpentier, B.; Geisler, S.; Z{\"u}gner, D.; and G{\"u}nnemann, S. 2021.
\newblock Graph posterior network: Bayesian predictive uncertainty for node classification.
\newblock \emph{Advances in Neural Information Processing Systems}, 34: 18033--18048.

\bibitem[{Sui et~al.(2022)Sui, Wang, Wu, Lin, He, and Chua}]{sui2022causal}
Sui, Y.; Wang, X.; Wu, J.; Lin, M.; He, X.; and Chua, T.-S. 2022.
\newblock Causal attention for interpretable and generalizable graph classification.
\newblock In \emph{Proceedings of the 28th ACM SIGKDD Conference on Knowledge Discovery and Data Mining}, 1696--1705.

\bibitem[{Tang et~al.(2024)Tang, Wu, Wu, Huang, Chen, Lei, and Meng}]{simcalib}
Tang, B.; Wu, Z.; Wu, X.; Huang, Q.; Chen, J.; Lei, S.; and Meng, H. 2024.
\newblock SimCalib: Graph Neural Network Calibration Based on Similarity between Nodes.
\newblock In \emph{Proceedings of the AAAI Conference on Artificial Intelligence}, volume~38, 15267--15275.

\bibitem[{Tang et~al.(2020)Tang, Yao, Sun, Wang, Tang, Aggarwal, Mitra, and Wang}]{tang2020investigating}
Tang, X.; Yao, H.; Sun, Y.; Wang, Y.; Tang, J.; Aggarwal, C.; Mitra, P.; and Wang, S. 2020.
\newblock Investigating and mitigating degree-related biases in graph convoltuional networks.
\newblock In \emph{Proceedings of the 29th ACM International Conference on Information \& Knowledge Management}, 1435--1444.

\bibitem[{Topping et~al.(2021)Topping, Di~Giovanni, Chamberlain, Dong, and Bronstein}]{topping2021understanding}
Topping, J.; Di~Giovanni, F.; Chamberlain, B.~P.; Dong, X.; and Bronstein, M.~M. 2021.
\newblock Understanding over-squashing and bottlenecks on graphs via curvature.
\newblock \emph{arXiv preprint arXiv:2111.14522}.

\bibitem[{Veli{\v{c}}kovi{\'c} et~al.(2017)Veli{\v{c}}kovi{\'c}, Cucurull, Casanova, Romero, Lio, and Bengio}]{velivckovic2017graph}
Veli{\v{c}}kovi{\'c}, P.; Cucurull, G.; Casanova, A.; Romero, A.; Lio, P.; and Bengio, Y. 2017.
\newblock Graph attention networks.
\newblock \emph{arXiv preprint arXiv:1710.10903}.

\bibitem[{Vovk, Gammerman, and Shafer(2005)}]{vovk2005algorithmic}
Vovk, V.; Gammerman, A.; and Shafer, G. 2005.
\newblock \emph{Algorithmic learning in a random world}, volume~29.
\newblock Springer.

\bibitem[{Wang, Yang, and Cheng(2022)}]{wang2022gcl}
Wang, M.; Yang, H.; and Cheng, Q. 2022.
\newblock GCL: Graph Calibration Loss for Trustworthy Graph Neural Network.
\newblock In \emph{Proceedings of the 30th ACM International Conference on Multimedia}, 988--996.

\bibitem[{Wang et~al.(2021)Wang, Liu, Shi, and Yang}]{wang2021confident}
Wang, X.; Liu, H.; Shi, C.; and Yang, C. 2021.
\newblock Be confident! towards trustworthy graph neural networks via confidence calibration.
\newblock \emph{Advances in Neural Information Processing Systems}, 34: 23768--23779.

\bibitem[{Wen, Tran, and Ba(2020)}]{wen2020batchensemble}
Wen, Y.; Tran, D.; and Ba, J. 2020.
\newblock Batchensemble: an alternative approach to efficient ensemble and lifelong learning.
\newblock \emph{arXiv preprint arXiv:2002.06715}.

\bibitem[{Wu et~al.(2023)Wu, Chen, Yang, and Yan}]{wu2023energy}
Wu, Q.; Chen, Y.; Yang, C.; and Yan, J. 2023.
\newblock Energy-based out-of-distribution detection for graph neural networks.
\newblock \emph{arXiv preprint arXiv:2302.02914}.

\bibitem[{Xing et~al.(2019)Xing, Arik, Zhang, and Pfister}]{xing2019distance}
Xing, C.; Arik, S.; Zhang, Z.; and Pfister, T. 2019.
\newblock Distance-based learning from errors for confidence calibration.
\newblock \emph{arXiv preprint arXiv:1912.01730}.

\bibitem[{Xu et~al.(2018{\natexlab{a}})Xu, Hu, Leskovec, and Jegelka}]{xu2018powerful}
Xu, K.; Hu, W.; Leskovec, J.; and Jegelka, S. 2018{\natexlab{a}}.
\newblock How powerful are graph neural networks?
\newblock \emph{arXiv preprint arXiv:1810.00826}.

\bibitem[{Xu et~al.(2018{\natexlab{b}})Xu, Li, Tian, Sonobe, Kawarabayashi, and Jegelka}]{xu2018representation}
Xu, K.; Li, C.; Tian, Y.; Sonobe, T.; Kawarabayashi, K.-i.; and Jegelka, S. 2018{\natexlab{b}}.
\newblock Representation learning on graphs with jumping knowledge networks.
\newblock In \emph{International conference on machine learning}, 5453--5462. PMLR.

\bibitem[{Yang et~al.(2024)Yang, Yang, Shi, Li, Zhang, and Zhou}]{dcgc}
Yang, C.; Yang, C.; Shi, C.; Li, Y.; Zhang, Z.; and Zhou, J. 2024.
\newblock Calibrating Graph Neural Networks from a Data-centric Perspective.
\newblock In \emph{Proceedings of the ACM on Web Conference 2024}, 745--755.

\bibitem[{Yang, Zhan, and Gan(2023)}]{yang2023beyond}
Yang, J.-Q.; Zhan, D.-C.; and Gan, L. 2023.
\newblock Beyond Probability Partitions: Calibrating Neural Networks with Semantic Aware Grouping.
\newblock \emph{arXiv preprint arXiv:2306.04985}.

\bibitem[{Yun et~al.(2021)Yun, Kim, Lee, Kang, and Kim}]{yun2021neo}
Yun, S.; Kim, S.; Lee, J.; Kang, J.; and Kim, H.~J. 2021.
\newblock Neo-gnns: Neighborhood overlap-aware graph neural networks for link prediction.
\newblock \emph{Advances in Neural Information Processing Systems}, 34: 13683--13694.

\bibitem[{Zargarbashi, Antonelli, and Bojchevski(2023)}]{zargarbashi2023conformal}
Zargarbashi, S.~H.; Antonelli, S.; and Bojchevski, A. 2023.
\newblock Conformal prediction sets for graph neural networks.
\newblock In \emph{International Conference on Machine Learning}, 12292--12318. PMLR.

\bibitem[{Zeng et~al.(2019)Zeng, Zhou, Srivastava, Kannan, and Prasanna}]{zeng2019graphsaint}
Zeng, H.; Zhou, H.; Srivastava, A.; Kannan, R.; and Prasanna, V. 2019.
\newblock Graphsaint: Graph sampling based inductive learning method.
\newblock \emph{arXiv preprint arXiv:1907.04931}.

\bibitem[{Zhang, Kailkhura, and Han(2020)}]{zhang2020mix}
Zhang, J.; Kailkhura, B.; and Han, T. Y.-J. 2020.
\newblock Mix-n-match: Ensemble and compositional methods for uncertainty calibration in deep learning.
\newblock In \emph{International conference on machine learning}, 11117--11128. PMLR.

\bibitem[{Zhang and Chen(2018)}]{zhang2018link}
Zhang, M.; and Chen, Y. 2018.
\newblock Link prediction based on graph neural networks.
\newblock \emph{Advances in neural information processing systems}, 31.

\bibitem[{Zhao et~al.(2020)Zhao, Chen, Hu, and Cho}]{zhao2020uncertainty}
Zhao, X.; Chen, F.; Hu, S.; and Cho, J.-H. 2020.
\newblock Uncertainty aware semi-supervised learning on graph data.
\newblock \emph{Advances in Neural Information Processing Systems}, 33: 12827--12836.

\bibitem[{Zhu et~al.(2021)Zhu, Zhang, Xhonneux, and Tang}]{zhu2021neural}
Zhu, Z.; Zhang, Z.; Xhonneux, L.-P.; and Tang, J. 2021.
\newblock Neural bellman-ford networks: A general graph neural network framework for link prediction.
\newblock \emph{Advances in Neural Information Processing Systems}, 34: 29476--29490.

\bibitem[{Zhuang et~al.(2024)Zhuang, Jiang, Zheng, Wang, and Zhao}]{gets}
Zhuang, D.; Jiang, C.; Zheng, Y.; Wang, S.; and Zhao, J. 2024.
\newblock GETS: Ensemble Temperature Scaling for Calibration in Graph Neural Networks.
\newblock \emph{arXiv preprint arXiv:2410.09570}.

\end{thebibliography}
